\documentclass[letterpaper, twoside, 10pt]{article}

\usepackage{amsmath}
\usepackage{amssymb}
\usepackage{bm}

\usepackage[margin=1in]{geometry}

\usepackage{graphics}%
\usepackage{wrapfig}
\usepackage[tight]{subfigure}

\usepackage{tabularx}
\usepackage{booktabs}
\usepackage{multirow}
\usepackage{xspace} %

\usepackage[ruled]{algorithm2e}

\usepackage[utf8]{inputenc}
\usepackage{authblk}

\makeatletter
\renewcommand\AB@authnote[1]{\rlap{\textsuperscript{\normalfont#1}}}

\makeatother

\usepackage{tikz}
\usetikzlibrary{decorations.text,calc,shapes.geometric,shapes.callouts}

\usepackage{pgfplots}
\pgfmathdeclarefunction{gauss}{2}{%
  \pgfmathparse{1/(#2*sqrt(2*pi))*exp(-((x-#1)^2)/(2*#2^2))}%
}

\newcommand{\argmin}[1]{\ensuremath{\underset{#1}{\textrm{arg min}}\;}}
\newcommand{\argmax}[1]{\ensuremath{\underset{#1}{\textrm{arg max}}\;}}

\newcommand*{\parens}[1]{\left( #1 \right)}

\newcommand*{\brackets}[1]{ \left[ #1 \right] }

\newcommand*{\Wtranspose}{W^\top}

\newcommand*{\wktranspose}{w_k^\top}
\newcommand*{\astar}{{a^{*}_t}}
\newcommand*{\pistar}{\pi^*}

\newcommand*{\dldwfrac}{\frac{\partial \ell}{\partial W}}

\newcommand*{\Mapt}{S_t}
\newcommand*{\Mapti}{S_t^{\parens{i}}}
\newcommand*{\Momenta}{\hat{\Phi}_1}  %
\newcommand*{\Momentb}{\hat{\Phi}_2}

\newcommand*{\Momentk}{\hat{\Phi}_k}
\newcommand*{\MomentK}{\hat{\Phi}_K}

\newcommand*{\XAMapt}{\parens{x_t, a_t, \Mapt}}

\newcommand*{\phiXAMapi}{\phi \parens{x_t, a_t, \Mapti}}

\newcommand*{\costMapt}{c\parens{x_t, a_t, \Mapt}}
\newcommand*{\costMaptAstar}{c\parens{x_t, \astar \negmedspace, \Mapt}}

\newcommand*{\astop}{a_\textrm{stop}}

\newcommand{\DAgger}{\textsc{DAgger}\xspace}

\usepackage{hyperref}

\hypersetup{
    colorlinks=true,
    citecolor=blue,
    linkcolor=blue,
    urlcolor=blue,
    pdfauthor={Matthew R. Walter, Siddharth Patki, Andrea F. Daniele, Ethan Fahnestock, Felix Duvallet, Sachithra Hemachandra, Jean Oh, Anthony Stentz, Nicholas Roy, Thomas M. Howard},
    pdftitle={Language Understanding for Field and Service Robots\\ in a Priori Unknown Environments}
}

\usepackage{url}
\usepackage{ulem}
\usepackage{natbib}

\title{Language Understanding for Field and Service Robots\\ in a Priori Unknown Environments}
\date{}

\author[1\thanks{Corresponding author: mwalter@ttic.edu}]{Matthew R.\ Walter}
\author[2]{\,Siddharth Patki}
\author[1]{Andrea F.\ Daniele}
\author[2]{Ethan Fahnestock}
\author[3]{\\Felix Duvallet}
\author[4]{Sachithra Hemachandra}
\author[5]{Jean Oh}
\author[5]{Anthony Stentz}
\author[6]{\\Nicholas Roy}
\author[2]{Thomas M.\ Howard}
\affil[1]{Toyota Technological Institute at Chicago}
\affil[2]{University of Rochester}
\affil[3]{Kodiak Robotics, Inc.}
\affil[4]{Cruise}
\affil[5]{Carnegie Mellon University}
\affil[6]{Massachusetts Institute of Technology}

\begin{document}
\maketitle

\begin{abstract}
    Contemporary approaches to perception, planning, estimation, and control have allowed robots to operate robustly as our remote surrogates in uncertain, unstructured environments. This progress now creates an opportunity for robots to operate not only in isolation, but also with and alongside humans in our complex environments. Realizing this opportunity requires an efficient and flexible medium through which humans can communicate with collaborative robots. Natural language provides one such medium, and through  significant progress in statistical methods for natural-language understanding, robots are now able to interpret a diverse array of free-form navigation, manipulation, and mobile-manipulation commands. However, most contemporary approaches require a detailed, prior spatial-semantic map of the robot's environment that models the space of possible referents of an utterance. Consequently, these methods fail when robots are deployed in new, previously unknown, or partially-observed environments, particularly when mental models of the environment differ between the human operator and the robot. This paper provides a comprehensive description of a novel learning framework that allows field and service robots to interpret and correctly execute natural-language instructions in a priori unknown, unstructured environments. Integral to our approach is its use of language as a ``sensor''---inferring spatial, topological, and semantic information implicit in natural-language utterances and then exploiting this information to learn a distribution over a latent environment model. We incorporate this distribution in a probabilistic, language grounding model and infer a distribution over a symbolic representation of the robot's action space, consistent with the utterance. We use imitation learning to identify a belief-space policy that reasons over the environment and behavior distributions. We evaluate our framework through a variety of different navigation and mobile-manipulation experiments involving an unmanned ground vehicle, a robotic wheelchair, and a mobile manipulator, demonstrating that the algorithm can follow natural-language instructions without prior knowledge of the environment.
\end{abstract}
\section{Introduction} \label{sec:intro}
Advancements in perception, planning, and control have enabled robots to move
from controlled isolation in factories and laboratories to semi-structured and
unstructured environments. Notable domains where field and service robots have
proven successful at operating in the presence of uncertainty include material
handling~\citep{durrant1996autonomous,durrant-whyte07,walter15}, underground
mining~\citep{scheding1997experiments,scheding1999experiment,roberts2000autonomous,marshall2008autonomous,duff2003automation},
disaster mitigation~\citep{nagatani2008development,
nagatani2013emergency,keiji2011redesign}, space science and
exploration~\citep{maimone2007two,furgale2010visual,arvidson2010spirit},
underwater science and exploration~\citep{singh2004seabed, johnson2010generation,
williams2012monitoring, yoerger2007techniques, bowen2008nereus,
camilli2010tracking, german2008hydrothermal}, and autonomous
driving~\citep{thrun06a,urmson06a,urmson08a,bacha08a, miller08a, montemerlo08a,
bohren08a, leonard08a}. With few exceptions~\citep{walter15}, however, current
operational, field-robotic systems function with either full autonomy or under at least partially-supervised teleoperation. More common in field operations, including those
conducted by the
military~\citep{kang2003robhaz, ryu2004multi, yamauchi2004packbot}, teleoperation
places significant cognitive load on the operator, who must interpret the
robot's multiple, diverse sensor streams in order to establish situational
awareness, while simultaneously and continuously controlling the robot's
low-level degrees of freedom. Consequently, teleoperation limits the
effectiveness and efficiency of the tasks that can be performed.

\begin{wrapfigure}{r}{0.5\textwidth}
  \centering
  \includegraphics[width=0.48\textwidth]{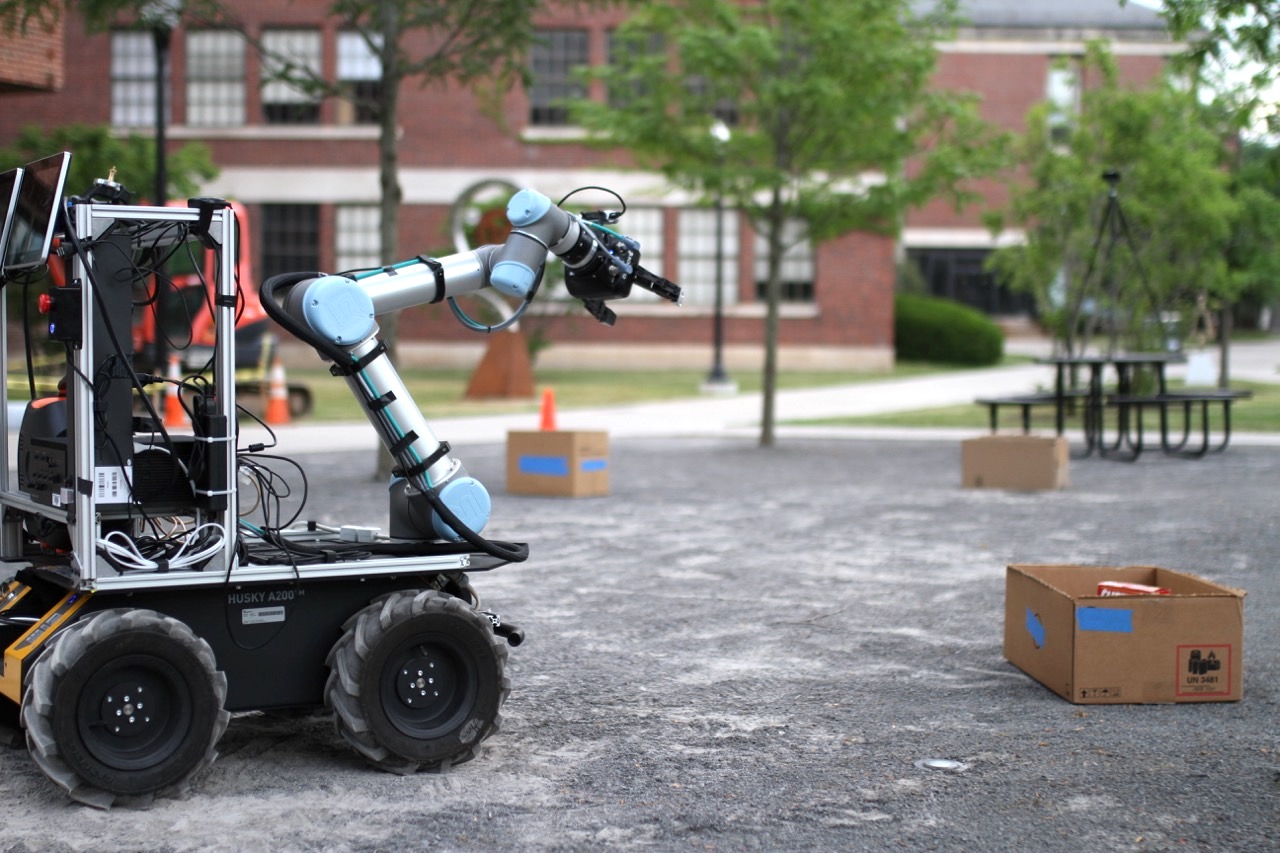}
  \caption{A robot is directed to ``retrieve the ball inside the box'' in an a priori unknown environment.}
  \label{fig:intro}
\end{wrapfigure}
A long-standing goal is to realize field and service robots that operate in the continuum between full autonomy and supervised teleoperation, not only as our surrogates, but also as our \emph{partners} working with and alongside people. Achieving this goal requires command and control mechanisms that are both intuitive and efficient, and spoken language offers a flexible medium
through which people can communicate with robots, without the need for specialized interfaces or significant training. For example, people can issue
verbal instructions that direct robotic forklifts~\citep{walter15} to load and unload cargo in semi-structured storage and distribution facilities. Similarly, a voice-commandable
wheelchair~\citep{hemachandra11} enables people with limited mobility to navigate their environments independently and safely simply by speaking to their wheelchair, rather than requiring the use of traditional sip-and-puff arrays or head-actuated switches.

The potential of natural language as an effective command and control mechanism has motivated significant advances in statistical approaches to language understanding. These models and algorithms enable robots operating in a variety of domains to interpret free-form instructions commanding tasks that include navigation~\citep{kollar10, matuszek10, chen11, matuszek12a, thomason15} and object manipulation~\citep{tellex11, howard14, thomason16, thomason18, shridhar18}, as well as to generate natural-language utterances~\citep{tellex14, daniele17, shridhar18}. Natural-language understanding is often formulated as a symbol-grounding problem, wherein the task is to associate linguistic phrases with their corresponding referents in the robot's symbolic model of its state and action spaces. Most contemporary approaches assume that this model is known a priori in the form of a ``world model'' that expresses relevant information about the robot's environment (e.g., the location, semantic class, and colloquial name of every object and spatial region that the user may refer to). Such a world model is typically constructed by manually adding semantic information to spatial maps produced using an off-the-shelf SLAM algorithm~\citep{kaess08, kaess08-software}. This practice inherently prevents natural-language understanding in environments that are unknown or partially known to the robot. Consider a mobile manipulator that is placed in a new environment consisting of multiple boxes that are all outside the field-of-view of the robot's sensors (Fig.~\ref{fig:intro}). Suppose that a person directs the robot to ``retrieve the ball inside the box''. Without knowledge of the environment, the robot is unable to associate the phrases ``the ball'' or ``the box'' to specific locations (i.e., symbol instantiations). In this case, most existing methods would either fail to ground the instruction or cause the robot to naively explore the environment until it happens upon a ball contained in a box.

\begin{figure}[!t]
  \centering
  \subfigure[Time $t = 0$\,sec]{\includegraphics[width=0.49\linewidth]{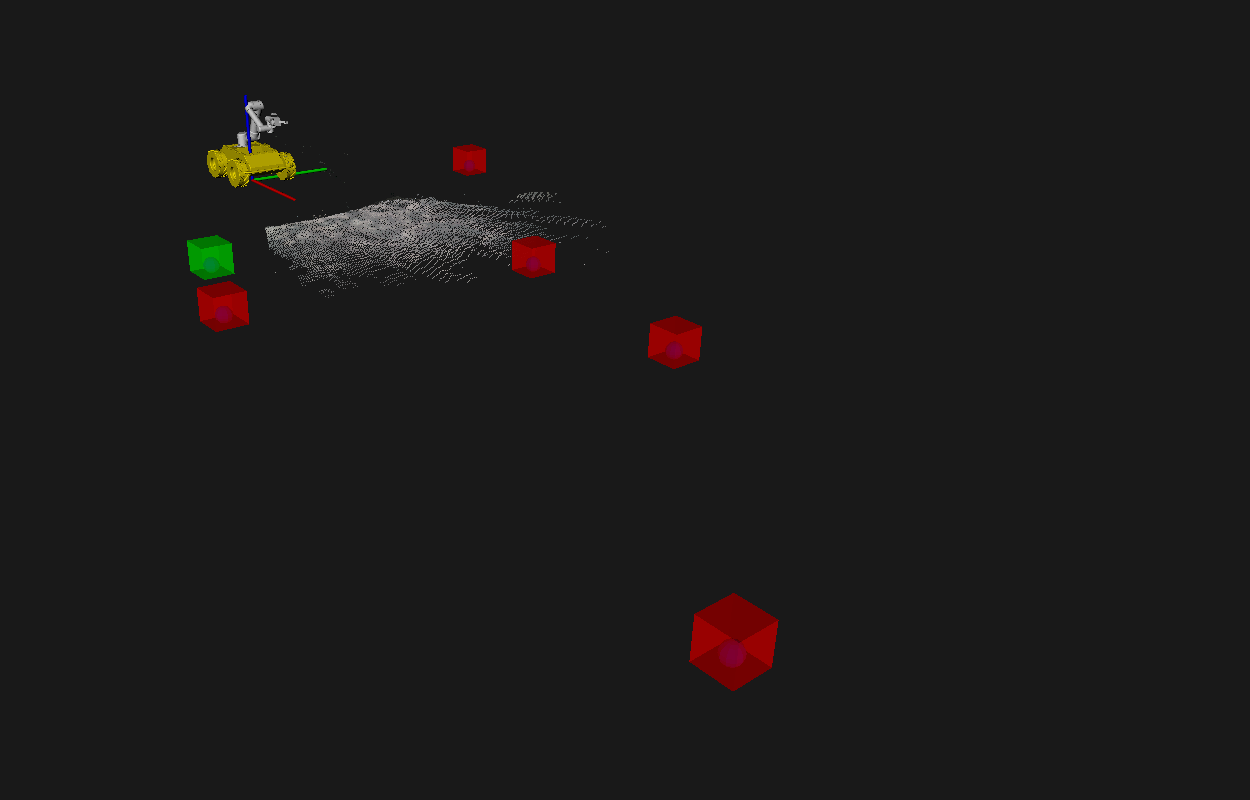}\label{fig:motivation-1}}\hfil%
  \subfigure[Time $t = 90$\,sec]{\includegraphics[width=0.49\linewidth]{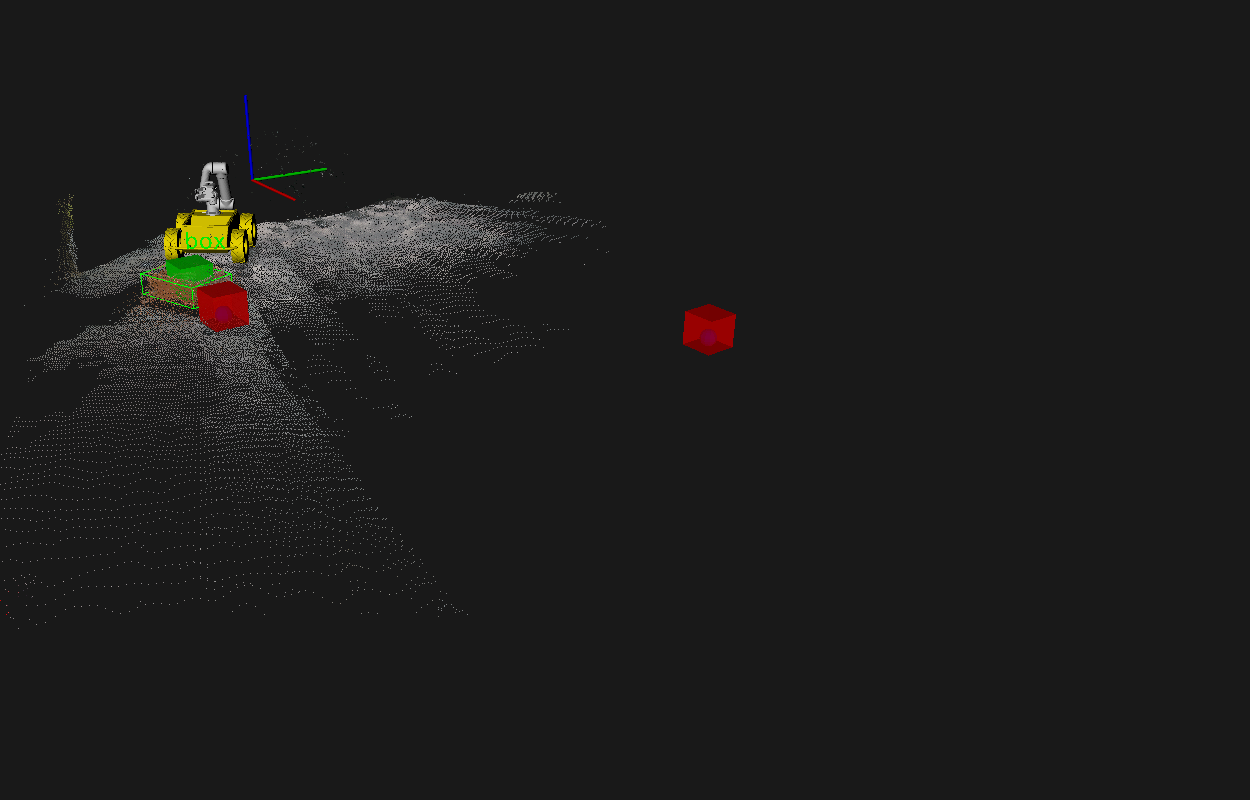}\label{fig:motivation-2}}\\%
  \subfigure[Time $t = 320$\,sec]{\includegraphics[width=0.49\linewidth]{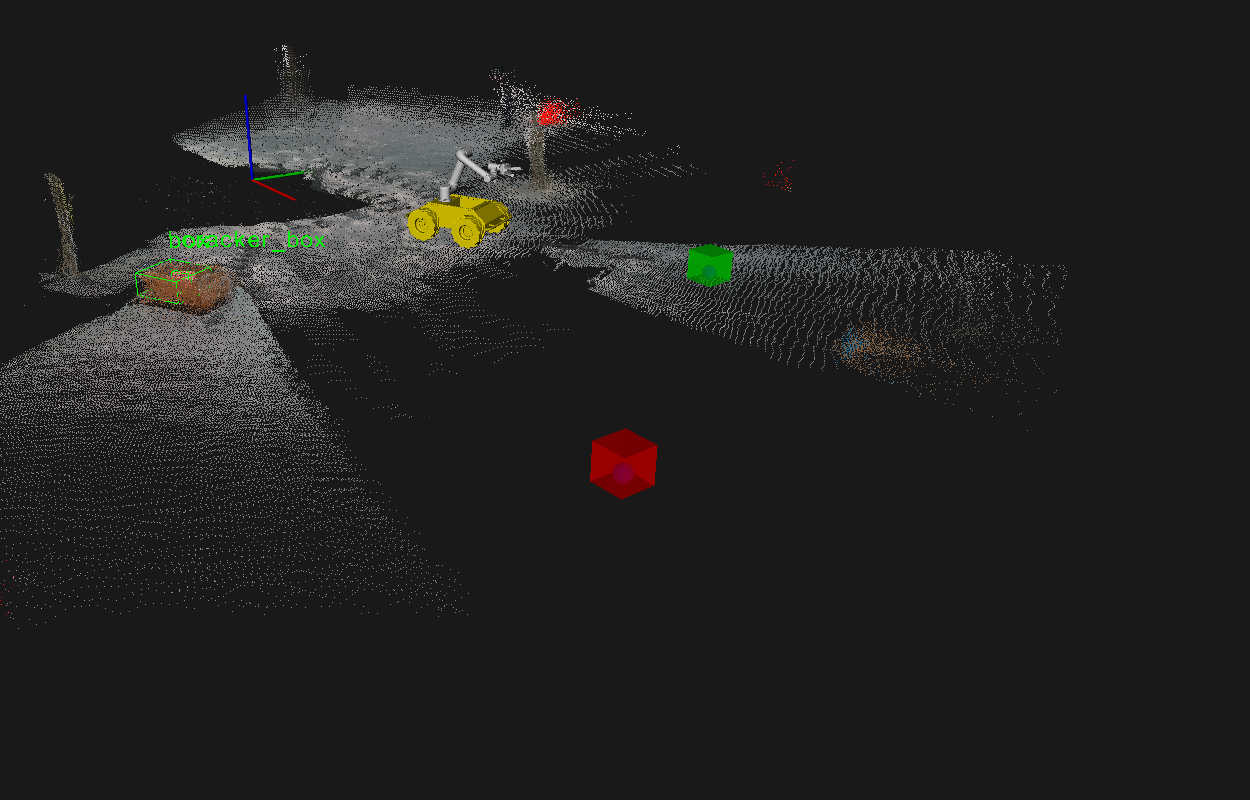}\label{fig:motivation-3}}\hfil%
  \subfigure[Time $t = 563$\,sec]{\includegraphics[width=0.49\linewidth]{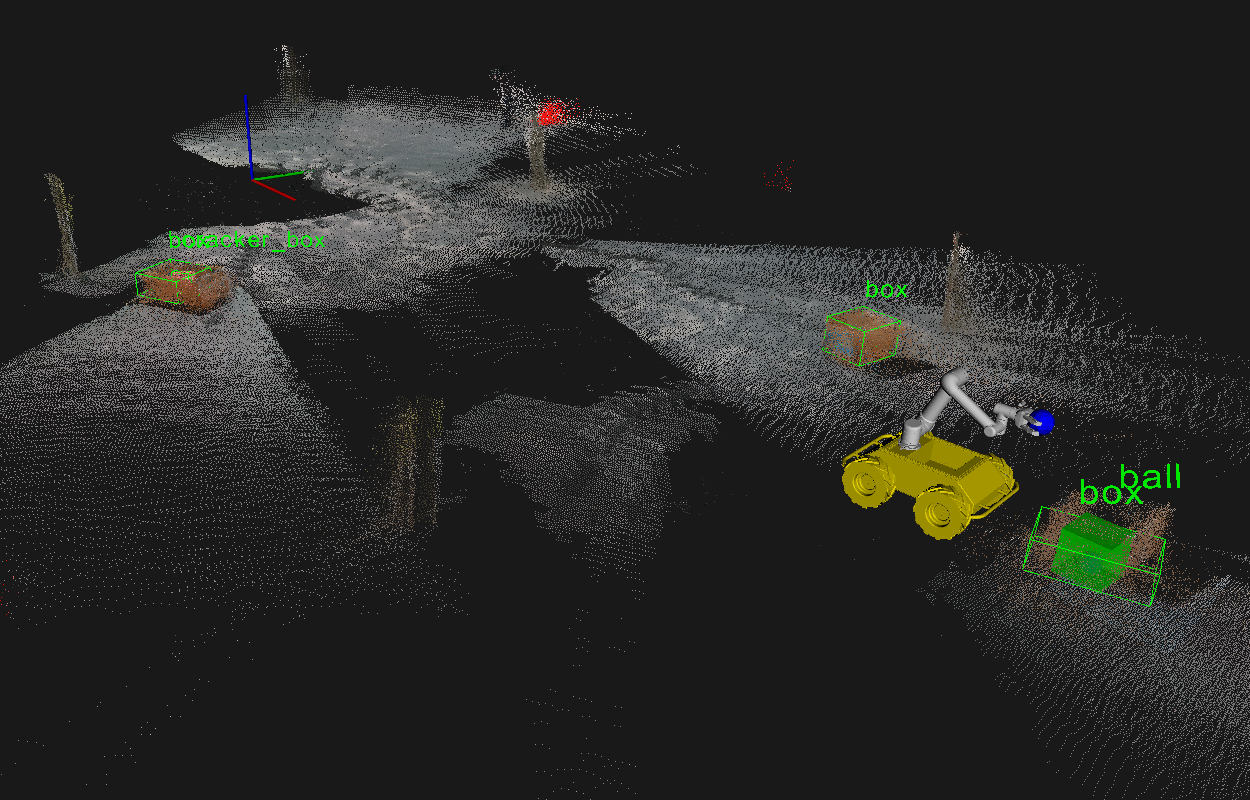}\label{fig:motivation-4}}\hfil%
  \caption{Our framework learns to exploit environment-related information implicit in a given utterance to hypothesize a distribution over possible maps in a priori unknown environments. Traditional approaches to language grounding involve reasoning over a detailed world model that is assumed to be known a priori. In order to allow grounding in a priori unknown or partially observed environments, our method maintains a distribution over hypothesized, spatial-semantic maps based upon environment information conveyed in the utterance. Consider a scenario in which the robot is instructed to ``pick up the ball inside the box'' in an unknown environment. Upon receiving the instruction,
  \subref{fig:motivation-1} the algorithm uses information extracted from the command to hypothesize the location of potential boxes, some of which contain a ball. Here, we visualize samples drawn from this distribution. The solid green cube denotes the hypothesized box that is the current goal of the planner. As the robot navigates, it \subref{fig:motivation-2} detects actual boxes (green wireframe) that are found to not contain a ball, while also failing to confirm the presence of hypothesized boxes sampled from the distribution. The algorithm updates the world model distribution accordingly, and the planner updates the goal. This continues until \subref{fig:motivation-4} the robot observes a box containing a ball and subsequently retrieves the ball, satisfying the instruction.}
  \label{fig:motivation}
\end{figure}

This paper provides a comprehensive presentation of a framework that enables robots to follow natural-language instructions that command navigation and mobile manipulation in a priori unknown environments (Fig.~\ref{fig:motivation}). Key to our approach is its exploitation of the spatial, topologic, and semantic information that the utterance conveys, effectively treating language as another ``sensor.'' Three algorithmic contributions are integral to this formulation.
\begin{itemize}
    \item First, a learned language-grounding model efficiently infers environment ``observations,'' implicit or explicit in the command, as well as the desired behaviors.
    \item Second, an estimation-theoretic algorithm hypothesizes the structure of the unobserved environment by using these language-based observations and those from the robot's traditional sensor streams to build and maintain a probability distribution over the world model (Fig.~\ref{fig:motivation-2}).
    \item Third, an imitation learning-based approach learns a belief-space policy that reasons over this world model distribution to identify navigation and manipulation actions that are optimal, given limited environment knowledge (Fig.~\ref{fig:motivation-4}).
\end{itemize}

In the following sections we first position our work in the context of contemporary approaches to language understanding. We then provide a thorough technical description of the proposed model and give the details of the experimental results on navigation and mobile manipulation tasks using three different robotic platforms. We then conclude with a discussion of the strengths and weaknesses of the proposed approach, along with directions for current and future work that seeks to address these limitations in order to facilitate more efficient human-robot collaboration.

\section{Related Work} \label{sec:related}

Natural-language understanding for human-robot interaction has been studied extensively over the past several decades. Symbol grounding~\citep{harnad90} is a common approach to language understanding, whereby words and phrases are interpreted in terms of their associated symbols in the robot's model of the  world. Early work in symbol grounding~\citep{winograd71,roy03,macmahon06} utilizes manually engineered correspondences and features that relate words to symbols comprised of the actions and a structured environment model. Consequently, these approaches employ a compact, predetermined grammar and a small set of symbols that limit the diversity of language that they can handle.

In contrast, contemporary approaches use statistical models that are trained in
a data-driven fashion to learn to express a large set of linguistic, spatial,
and semantic features. These approaches enable robots to successfully interpret
natural-language utterances that command navigation~\citep{kollar10, matuszek10,
chen11, matuszek12a, thomason15}, object manipulation~\citep{bollini10, howard14,
misra16, thomason16, thomason18, shridhar18, paul18}, and mobile
manipulation~\citep{tellex11, walter15}, as well as to generate natural-language
utterances~\citep{tellex14,daniele17,shridhar18} in a variety of complex domains.
Data-driven approaches to symbol
grounding~\citep{tellex11,howard14,howard14b,paul18} learn probabilistic models
that exploit the hierarchical structure of language in order to associate (i.e.,
``ground'') each phrase in an utterance to its corresponding referent in a
symbolic representation of the robot's state and action spaces. These methods
generally require that the robot has prior knowledge of the the environment, for
example in the form of a spatial-semantic map of the different objects and
regions (e.g., rooms). In practice, these maps are often generated by first
teleoperating the robot around the environment and using a state-of-the-art SLAM
algorithm~\citep{olson06, kaess08} to build a metric map of the environment.
These maps are then manually annotated to include semantic information, e.g., by
delineating each object and room, and then assigning them a label, to arrive at
an environment map  sufficient to provide a symbolic world model. An alternative
is to incorporate this information as part of the initial mapping using a
semantic SLAM framework~\citep{galindo05, mozos07, meger08, vasudevan08,
krieg05, zender08, pronobis10, hemachandra11, pronobis12}. These approaches
build on the effectiveness of SLAM by augmenting a low-level metric map with
layers that encode the topological and semantic properties of the environment
extracted from the robot's sensor data (e.g., LIDAR scans and camera images),
using scene classifiers~\citep{nuchter03, mozos07, pronobis10} and object
detectors~\citep{torralba03, meger08, vasudevan08, kollar09}. For a comprehensive discussion of the role of environment representations in language-based spatial reasoning, we refer the reader to the survey by \citet{landsiedel17}.

While most language grounding methods rely on access to such a prior map of the environment, there are notable exceptions that are capable of language understanding in unknown environments. Particularly relevant is the work of \citet{duvallet13}, which opportunistically builds a deterministic map of the a priori unknown environment as the robot navigates. This map serves as input to a policy that is trained via imitation learning to emulate the way in which humans follow instructions in unknown environments. Our approach similarly uses imitation learning to identify the robot's policy based on human demonstrations, but unlike the work of \citet{duvallet13}, our policy reasons over a probabilistic model of the environment that makes explicit information that the instruction conveys. Also relevant are recent neural network-based approaches to language-based navigation in novel (i.e., unknown) environments~\citep{mei16, anderson18}. Unlike our approach, these methods map language directly to actions, and do not (explicitly) infer a distribution over possible world models from language. Meanwhile, statistical parsing-based methods~\citep{matuszek10, chen11, matuszek12a, thomason15} associate natural-language utterances to a meaning representation that typically takes the form of a lambda calculus. Such an approach avoids the need for an explicit world model, typically at the expense of requiring a down-stream controller capable of executing inferred plans in unknown environments.

Also relevant is recent work that focuses on grounding unknown or ambiguous utterances. One approach to dealing with ambiguous utterances is to utilize inverse grounding~\citep{tellex14, gong18} to generate targeted questions for the user that are deemed to be most informative, e.g., in terms of the reduction in entropy for the grounding distribution~\citep{tellex12}. Meanwhile, several methods learn a priori unknown grounding models by exploring the relationship between novel linguistic predicates and the robot's world model and/or its percepts~\citep{thomason16, she17, tucker17, thomason18}. Our work differs in that we assume that the concepts are known, but that the instantiations of these concepts in the robot's environment are unknown.

Meanwhile, much attention of late has been applied to the problem of navigating a priori unknown environments towards a desired goal using only onboard sensing, typically in the form of monocular or RGBD images~\citep{kim15, zhu17, gupta17,rasouli20}, laser scans~\citep{chiang19,zeng19}, or a combination of the two~\citep{kollar09, aydemir11, aydemir13}. While language typically plays little-to-no role in these approaches (e.g., the goal may be specified by its named type), they are relevant to our work in that they learn a policy that is responsible for deciding where to navigate to next based on the robot's observation history. Similar to our approach, earlier work in this area reasons over a structured state space that is assumed to be partially observable. Search is then formulated as a decision-theoretic problem (e.g., in the context of a POMDP), whereby methods attempt to solve for the policy that is optimal based on the current state distribution. Similar to the way in which we use language to inform this distribution, these active search methods may exploit object-object and object-scene co-occurrence information~\citep{kollar09}, spatial relations~\citep{aydemir11}, or scene semantics~\citep{aydemir13}.

Most recent approaches to active visual search model the navigation policy as a neural network that maps low-level sensor data directly to actions. Trained in an end-to-end manner via reinforcement learning, the architectures learn to reason over scene semantics in an entirely data-driven fashion. In contrast, our framework maintains a distribution over the seen and unseen parts of the world, which serves as an explicit, intermediate representation suitable for language grounding and planning under uncertainty. Another notable difference compared to our work is that this family of methods typically assumes that the agent is aware of the distance and direction of the goal (i.e., the location of the goal relative to the robot) at each time step. In practice, this means that there is some way of localizing the robot in the environment (presumably, using an a priori map available to an oracle). We assume that the robot has no such information. Additionally, they assume that the agent's motion and observations are noise-free, which is not true for robots in practice. With few exceptions~\citep{sadeghi17,tai17,bansal19}, these methods have thus only been demonstrated in photorealistic simulators or on datasets~\citep{mirowski18,zeng19}.

A key aspect of our approach is its use of language as a sensor, whereby information conveyed in the instruction is used to build and maintain a distribution over the map of the environment. In this way, our approach is similar to recent methods that enable robots to learn spatial-semantic environment models from linguistic descriptions together with traditional sensor streams~\citep{zender08, pronobis12, walter13, walter14, hemachandra14}. Our method builds on this work in order to maintain a distribution over a model of the environment. However, this earlier work assumes access to free-form utterances that explicitly describe the robot's environment. In contrast, our proposed framework learns to infer environment knowledge that is implicit in natural-language descriptions. Additionally, these previous approaches focus on estimation as it pertains to building spatial-semantic environment models, whereas we consider mapping jointly with planning under uncertainty specifically to satisfy a user's natural-language instruction.

\section{Technical Approach} \label{sec:technicalapporach}

Contemporary approaches to natural-language understanding formulate the problem as probabilistic inference over a learned distribution that associates linguistic elements with their corresponding referents in a symbolic representation of the robot's state and action spaces. The space of symbols ${\Gamma} = \left\{\gamma_1, \gamma_2, \ldots, \gamma_n\right\}$ includes concepts derived from the robot's environment model, such as the location and category of objects and spatially extended regions (e.g., rooms, buildings, etc.), and a symbolic representation of viable robot behaviors, such as manipulating a specific object or navigating to a desired location. The distribution over symbols is conditioned on the parse of the free-form utterance ${\Lambda}_t = \left\{\lambda_1,\lambda_2 ... \lambda_n\right\}$,\footnote{In this way, we assume that the instructions are grammatically correct.} and a world model $\Mapt$ that represents environment knowledge extracted from the history of sensor measurements ${z}^t$ using a set of perceptual classifiers.
Framed as a symbol-grounding problem~\citep{harnad90}, natural-language understanding then typically follows as maximum a posteriori inference over the power set of referent symbols $\mathcal{P}({\Gamma})$
\begin{equation}
  {\Gamma^*_t} = \argmax{\mathcal{P}({\Gamma})} p( {\Gamma}  \vert {\Lambda}_t, \Mapt).
  \label{eqn:basic-1}
\end{equation}

This approach reasons over a model of the world $\Mapt$ that is assumed both to be known a priori and to express all information necessary to ground the given utterance. This precludes language understanding in unobserved (i.e., novel) or partially observed environments for which the world model is empty or incomplete, making accurate inference \eqref{eqn:basic-1} infeasible.

\subsection{Approach Overview}

We address this problem by treating symbol grounding as inference conditioned on a \emph{latent} model of the robot's environment ${{\Mapt}}$. In particular, we learn a probabilistic world model that exploits environmental information implicit in an utterance to build and maintain a distribution over the topological, metric, and semantic properties of the environment
\begin{equation} \label{eqn:world-distribution}
    p({{\Mapt}} \vert {\Lambda}^t, {z}^t, {u}^t),
\end{equation}
where ${\Lambda}^t$, ${z}^t$, and ${u}^t$ denote the history of utterances, sensor observations (e.g., laser scans, image streams, and object detections from the robot's perception pipeline), and odometry, respectively. In this way, we maintain a world model distribution that not only fuses information perceived with sensors onboard the robot, but also models the unperceived information about the environment that is expressed in the utterance. Treating the environment model as a latent random variable, we formulate symbol grounding as a problem of inferring a distribution over robot behaviors ${{\beta}}_t$. A behavior in $\beta_t$ is a symbolic representation of the intended robot actions expressed by the symbols in the inferred groundings $\Gamma^*_t$, and may include navigating to a specific location, grasping a particular object, etc.
\begin{figure}[!tb]
    \centering
    \begin{tikzpicture}[textnode/.style={anchor=mid,width=2cm},
        module/.style={rectangle,draw=black, align=center, minimum size=6mm,font=\footnotesize,align=center,text width=1.5cm,minimum width=1.75cm,minimum height=1cm}]
        \node[module] (annotation-inference) at (0,1.5) {annotation inference};
        \node[module] (semantic-mapping) at (0,-1.5) {semantic mapping};
        \node[module] (motion-planning) at (0,-3.5) {motion planning};
        \node[module] (behavior-inference) at (6.0,1.5) {behavior inference};
        \node[module] (policy-planner) at (6.0,-1.5) {policy planner};
        \node[] (robot) at (-6.0,-1.5) {\includegraphics[width=2.5cm]{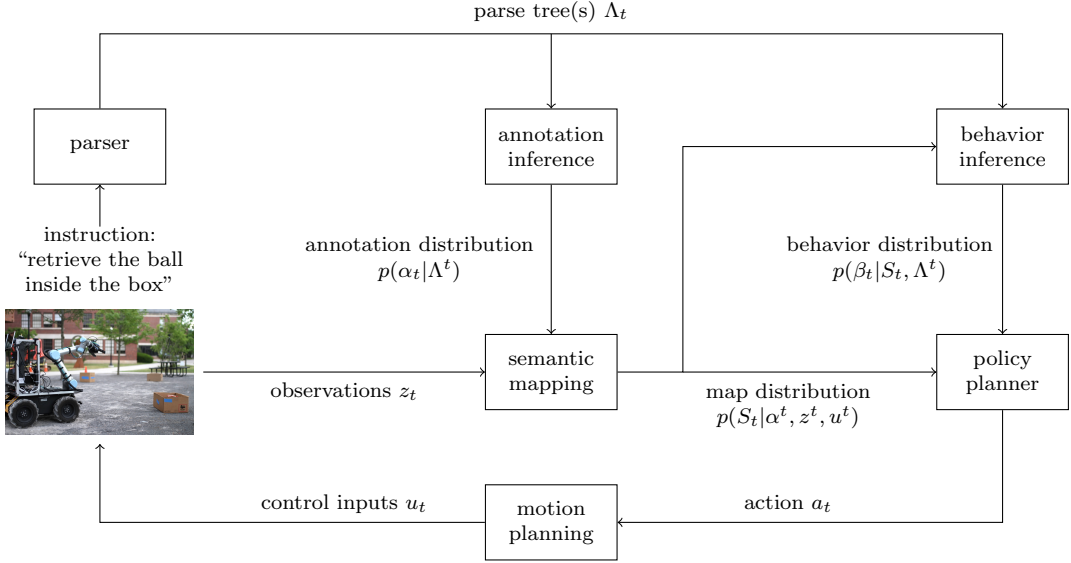}};
        \draw[->] (annotation-inference) -- (semantic-mapping);
        \draw[->] (semantic-mapping) --(1.75,-1.5) -- (1.75,1.5) -- (behavior-inference);
        \draw[->] (1.75,-1.5) -- (policy-planner);
        \draw[->] (behavior-inference) -- (policy-planner);
        \node[font=\footnotesize,align=center,text width=2.5cm] (observationlabel) at (3.15,-1.95) {map distribution\\ $p(S_t \vert \alpha^t, z^t, u^t)$};
        \node[font=\footnotesize,align=center,text width=4cm] (annotationdistributionlabel) at (-1.75,0) {annotation distribution\\$p(\alpha_t \vert \Lambda^t)$};
        \node[font=\footnotesize,align=center,text width=3.0cm] (behaviordistributionlabel) at (4.5,0) {behavior distribution\\$p(\beta_t \vert S_t, \Lambda^t)$};
        \node[module] (parserlabel) at (-6.0,1.5) {parser};
        \draw[->] (robot) -- (semantic-mapping);
        \draw[->] (0.0,3.0) --  (annotation-inference);
        \draw[->] (parserlabel) -- (-6.0,3.0) -- (6.0,3.0) -- (behavior-inference);
        \draw[->] (policy-planner) -- (6.0,-3.5) -- (motion-planning);
       	\draw[->] (motion-planning) -- (-6.0,-3.5) -- (robot);
        \node[anchor=center,font=\footnotesize,align=center,text width=2.4cm, inner sep=0pt] (instructionlabel) at (-6.0,0) {instruction:\\``retrieve the ball inside the box''};
        \draw[->] (instructionlabel) -- (parserlabel);
        \node[font=\footnotesize,align=center,text width=2cm] (observationslabel) at (-2.75,-1.75) {observations $z_t$};
        \node[font=\footnotesize,align=center,text width=3cm] (parsetreelabel) at (0, 3.3) {parse tree(s) $\Lambda_t$};
        \node[font=\footnotesize,align=center,text width=2cm] (actionlabel) at (3.15,-3.25) {action $a_t$};
        \node[font=\footnotesize,align=center,text width=3cm] (controllabel) at (-2.75,-3.25) {control inputs $u_t$};
    \end{tikzpicture}
    \caption{Our framework for language understanding in a priori unknown environments exploits environment descriptions available in a given instruction together with traditional sensing modalities to maintain a distribution over the environment model. A policy then reasons over this distribution together with inferred behaviors to identify an appropriate high-level action (e.g., an intermediate goal).  A motion planner converts these actions into control inputs that the robot executes. This process then repeats as the robot makes new observations, until the policy decides that the instruction has been satisfied. Details regarding state estimation, semantic perception, etc.\ are omitted for clarity.}
    \label{fig:framework}
\end{figure}
The optimal trajectory ${x}_{t}^*$ that the robot should take in the context of a distribution over behaviors then amounts to a planning under uncertainty problem formulated as inference over a model that marginalizes over the space of world models and robot behaviors
\begin{equation}
  {x}^*_t = \argmax{{x}_t \in {X}_t} \int_{\Mapt} \int_{\beta_t} \underbrace{p({x}_t \vert \beta_t, \Mapt)}_{\text{planning}} \times \underbrace{p( \beta_t \vert {\Lambda}_t, \Mapt)}_{\substack{\text{behavior} \\ \text{inference}}} \times \underbrace{p( \Mapt \vert  {\Lambda}^t, {z}^t, {u}^t)}_{\substack{\text{semantic} \\ \text{mapping}}} d \Mapt \, d\beta_t
  \label{eqn:nlu-marginalization}
\end{equation}

By structuring the problem in this way, we approach language understanding in a priori unknown environments as inference over three coupled learning problems. The framework (Fig.~\ref{fig:framework}) first converts
the parsed natural-language instruction into a set of environment annotations using a learned language grounding model. It then treats these annotations as observations of the environment (i.e., the existence, name,
and relative location of objects and rooms) that it uses together with data from the
robot's onboard sensors to learn a distribution over possible world models
(the third term in Eqn.~\ref{eqn:nlu-marginalization}). Following the example of executing the command to ``retrieve the ball inside the box'' (Fig.~\ref{fig:intro}), this may result in a distribution that hypothesizes the potential location of boxes in the environment, some of which contain hypothesized balls (Fig.~\ref{fig:motivation-1}). Our framework then infers
a distribution over behaviors conditioned on the world models and the
current utterance (the second term in Eqn.~\ref{eqn:nlu-marginalization}). In our example, this distribution would favor performing the ``pick'' action on an object of type ``ball'' whose location is consistent with being ``inside'' a box. Note that we assume that relevant information in any previous utterances, i.e., $\Lambda^{t-1}$, is captured in the map distribution. We then solve for the navigation and/or manipulation actions that are
consistent with this behavior distribution (the first term in Eqn.~\ref{eqn:nlu-marginalization}) using a learned
belief-space policy that commands a single, high-level action to the robot (e.g., navigating to a location where there is a high likelihood of there being a box).
As the robot executes this action, we update the world
model distribution based upon any new utterances and sensor observations, and
subsequently select an updated action according to the policy. This process
repeats until the policy concludes that the robot satisfied the instruction.

Reasoning over the entire space of behaviors and semantic maps would be intractable, particularly as the environment and behavior spaces grow. In order to make instruction-following tractable, we employ approximations to the individual distributions in Equation~\ref{eqn:nlu-marginalization} as well as use approximate inference methods. In particular, we represent the latent map and behaviors as discrete samples from their respective distributions. Each map sample represents one hypothesis of the robot's environment, and each behavior sample is a set of action constraints inferred from language in the context of a particular hypothesized map. We maintain the world model distribution using a Rao-Blackwellized particle filtering framework. The following details each component of our approach.

\subsection{Natural-Language Understanding} \label{sec:nlu} 
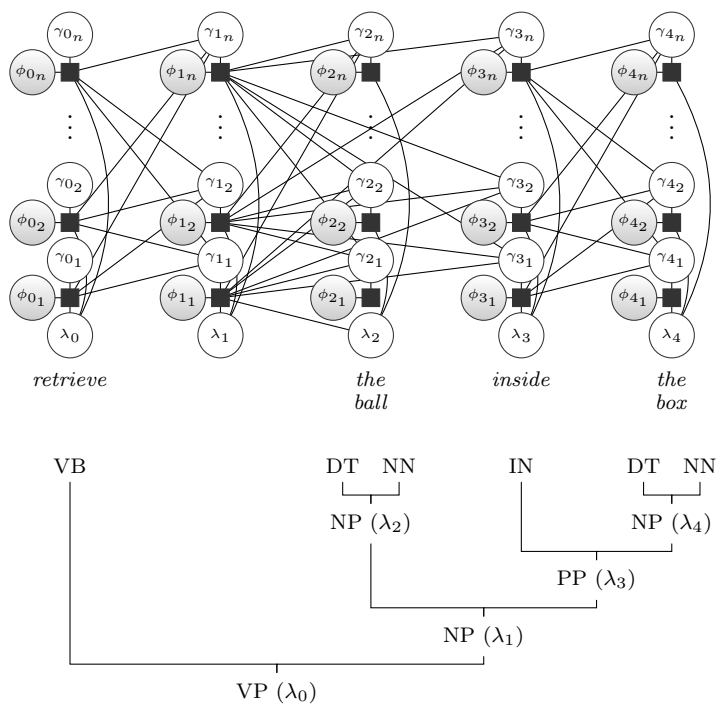
\begin{figure}[ht]
\centering
\begin{tikzpicture}[textnode/.style={anchor=mid,font=\tiny},nodeknown/.style={circle,draw=black!80,fill=white,minimum size=6mm,font=\tiny},nodeunknown/.style={circle,draw=black!80,fill=black!10,minimum size=6mm,font=\tiny,top color=white,bottom color=black!20},factor/.style={rectangle,draw=black!80,fill=black!80,minimum size=2mm,font=\tiny,text=white}]
\draw[-] (0,0.5) to (0,1);
\draw[-] (0,0.5) to [bend right=30] (0,2);
\draw[-] (0,0.5) to [bend right=30] (0,4);
\draw[-] (-0.5,1) to (0,1);
\draw[-] (-0.5,2) to (0,2);
\draw[-] (-0.5,4) to (0,4);
\draw[-] (0,1.5) to (0,1);
\draw[-] (0,2.5) to (0,2);
\draw[-] (0,4.5) to (0,4);
\draw[-] (2.0,1.5) to (0,1);
\draw[-] (2.0,2.5) to (0,1);
\draw[-] (2.0,4.5) to (0,1);
\draw[-] (2.0,1.5) to (0,2);
\draw[-] (2.0,2.5) to (0,2);
\draw[-] (2.0,4.5) to (0,2);
\draw[-] (2.0,1.5) to (0,4);
\draw[-] (2.0,2.5) to (0,4);
\draw[-] (2.0,4.5) to (0,4);
\draw[-] (4.0,1.5) to (2.0,1);
\draw[-] (4.0,2.5) to (2.0,1);
\draw[-] (4.0,4.5) to (2.0,1);
\draw[-] (4.0,1.5) to (2.0,2);
\draw[-] (4.0,2.5) to (2.0,2);
\draw[-] (4.0,4.5) to (2.0,2);
\draw[-] (4.0,1.5) to (2.0,4);
\draw[-] (4.0,2.5) to (2.0,4);
\draw[-] (4.0,4.5) to (2.0,4);
\draw[-] (2.0,0.5) to (2.0,1);
\draw[-] (2.0,0.5) to [bend right=30] (2.0,2);
\draw[-] (2.0,0.5) to [bend right=30] (2.0,4);
\draw[-] (1.5,1) to (2.0,1);
\draw[-] (1.5,2) to (2.0,2);
\draw[-] (1.5,4) to (2.0,4);
\draw[-] (2.0,1.5) to (2.0,1);
\draw[-] (2.0,2.5) to (2.0,2);
\draw[-] (2.0,4.5) to (2.0,4);
\draw[-] (6.0,1.5) to (2.0,1);
\draw[-] (6.0,2.5) to (2.0,1);
\draw[-] (6.0,4.5) to (2.0,1);
\draw[-] (6.0,1.5) to (2.0,2);
\draw[-] (6.0,2.5) to (2.0,2);
\draw[-] (6.0,4.5) to (2.0,2);
\draw[-] (6.0,1.5) to (2.0,4);
\draw[-] (6.0,2.5) to (2.0,4);
\draw[-] (6.0,4.5) to (2.0,4);
\draw[-] (4.0,0.5) to (2.0,1);
\draw[-] (4.0,0.5) to [bend right=30] (4.0,2);
\draw[-] (4.0,0.5) to [bend right=30] (4.0,4);
\draw[-] (3.5,1) to (4.0,1);
\draw[-] (3.5,2) to (4.0,2);
\draw[-] (3.5,4) to (4.0,4);
\draw[-] (4.0,1.5) to (4.0,1);
\draw[-] (4.0,2.5) to (4.0,2);
\draw[-] (4.0,4.5) to (4.0,4);
\draw[-] (6.0,0.5) to (6.0,1);
\draw[-] (6.0,0.5) to [bend right=30] (6.0,2);
\draw[-] (6.0,0.5) to [bend right=30] (6.0,4);
\draw[-] (5.5,1) to (6.0,1);
\draw[-] (5.5,2) to (6.0,2);
\draw[-] (5.5,4) to (6.0,4);
\draw[-] (6.0,1.5) to (6.0,1);
\draw[-] (6.0,2.5) to (6.0,2);
\draw[-] (6.0,4.5) to (6.0,4);
\draw[-] (8.0,1.5) to (6.0,1);
\draw[-] (8.0,2.5) to (6.0,1);
\draw[-] (8.0,4.5) to (6.0,1);
\draw[-] (8.0,1.5) to (6.0,2);
\draw[-] (8.0,2.5) to (6.0,2);
\draw[-] (8.0,4.5) to (6.0,2);
\draw[-] (8.0,1.5) to (6.0,4);
\draw[-] (8.0,2.5) to (6.0,4);
\draw[-] (8.0,4.5) to (6.0,4);
\draw[-] (8.0,0.5) to (8.0,1);
\draw[-] (8.0,0.5) to [bend right=30] (8.0,2);
\draw[-] (8.0,0.5) to [bend right=30] (8.0,4);
\draw[-] (7.5,1) to (8.0,1);
\draw[-] (7.5,2) to (8.0,2);
\draw[-] (7.5,4) to (8.0,4);
\draw[-] (8.0,1.5) to (8.0,1);
\draw[-] (8.0,2.5) to (8.0,2);
\draw[-] (8.0,4.5) to (8.0,4);
\node[nodeknown] (p0) at (0,0.5) {};
\node[textnode] (l0) at (0,-0.1) {\footnotesize{\textit{retrieve}}};
\node[font=\tiny] (p0label) at (0,0.5) {$\lambda_{0}$};
\node[nodeunknown] (c01) at (-0.5,1) {};
\node[font=\tiny] (c01label) at (-0.5,1) {$\phi_{0_{1}}$};
\node[nodeunknown] (c02) at (-0.5,2) {};
\node[font=\tiny] (c02label) at (-0.5,2) {$\phi_{0_{2}}$};
\node[nodeunknown] (c0n) at (-0.5,4) {};
\node[font=\tiny] (c0nlabel) at (-0.5,4) {$\phi_{0_{n}}$};
\node[nodeknown] (g01) at (0,1.5) {};
\node[font=\tiny] (g01label) at (0,1.5) {$\gamma_{0_{1}}$};
\node[nodeknown] (g02) at (0,2.5) {};
\node[font=\tiny] (g02label) at (0,2.5) {$\gamma_{0_{2}}$};
\node[] (g0dots) at (0,3.375) {$\vdots$};
\node[nodeknown] (g0n) at (0,4.5) {};
\node[font=\tiny] (g0nlabel) at (0,4.5) {$\gamma_{0_{n}}$};
\node[factor] (f01) at (0,1) {};
\node[factor] (f02) at (0,2) {};
\node[factor] (f0n) at (0,4) {};
\node[nodeknown] (p1) at (2.0,0.5) {};
\node[font=\tiny] (p1label) at (2.0,0.5) {$\lambda_{1}$};
\node[nodeunknown] (c11) at (1.5,1) {};
\node[font=\tiny] (c11label) at (1.5,1) {$\phi_{1_{1}}$};
\node[nodeunknown] (c12) at (1.5,2) {};
\node[font=\tiny] (c12label) at (1.5,2) {$\phi_{1_{2}}$};
\node[nodeunknown] (c1n) at (1.5,4) {};
\node[font=\tiny] (c1nlabel) at (1.5,4) {$\phi_{1_{n}}$};
\node[nodeknown] (g11) at (2.0,1.5) {};
\node[font=\tiny] (g11label) at (2.0,1.5) {$\gamma_{1_{1}}$};
\node[nodeknown] (g12) at (2.0,2.5) {};
\node[font=\tiny] (g12label) at (2.0,2.5) {$\gamma_{1_{2}}$};
\node[] (g1dots) at (2.0,3.375) {$\vdots$};
\node[nodeknown] (g1n) at (2.0,4.5) {};
\node[font=\tiny] (g1nlabel) at (2.0,4.5) {$\gamma_{1_{n}}$};
\node[factor] (f11) at (2.0,1) {};
\node[factor] (f12) at (2.0,2) {};
\node[factor] (f1n) at (2.0,4) {};
\node[textnode] (l2) at (4.0,-0.4) {\footnotesize{\textit{ball}}};
\node[textnode] (l2) at (4.0,-0.1) {\footnotesize{\textit{the}}};
\node[nodeknown] (p2) at (4.0,0.5) {};
\node[font=\tiny] (p2label) at (4.0,0.5) {$\lambda_{2}$};
\node[nodeunknown] (c21) at (3.5,1) {};
\node[font=\tiny] (c21label) at (3.5,1) {$\phi_{2_{1}}$};
\node[nodeunknown] (c22) at (3.5,2) {};
\node[font=\tiny] (c22label) at (3.5,2) {$\phi_{2_{2}}$};
\node[nodeunknown] (c2n) at (3.5,4) {};
\node[font=\tiny] (c2nlabel) at (3.5,4) {$\phi_{2_{n}}$};
\node[nodeknown] (g21) at (4.0,1.5) {};
\node[font=\tiny] (g21label) at (4.0,1.5) {$\gamma_{2_{1}}$};
\node[nodeknown] (g22) at (4.0,2.5) {};
\node[font=\tiny] (g22label) at (4.0,2.5) {$\gamma_{2_{2}}$};
\node[] (g2dots) at (4.0,3.375) {$\vdots$};
\node[nodeknown] (g2n) at (4.0,4.5) {};
\node[font=\tiny] (g2nlabel) at (4.0,4.5) {$\gamma_{2_{n}}$};
\node[factor] (f21) at (4.0,1) {};
\node[factor] (f22) at (4.0,2) {};
\node[factor] (f2n) at (4.0,4) {};
\node[textnode] (l3) at (6.0,-0.1) {\footnotesize{\textit{inside}}};
\node[nodeknown] (p3) at (6.0,0.5) {};
\node[font=\tiny] (p3label) at (6.0,0.5) {$\lambda_{3}$};
\node[nodeunknown] (c31) at (5.5,1) {};
\node[font=\tiny] (c31label) at (5.5,1) {$\phi_{3_{1}}$};
\node[nodeunknown] (c32) at (5.5,2) {};
\node[font=\tiny] (c32label) at (5.5,2) {$\phi_{3_{2}}$};
\node[nodeunknown] (c3n) at (5.5,4) {};
\node[font=\tiny] (c3nlabel) at (5.5,4) {$\phi_{3_{n}}$};
\node[nodeknown] (g31) at (6.0,1.5) {};
\node[font=\tiny] (g31label) at (6.0,1.5) {$\gamma_{3_{1}}$};
\node[nodeknown] (g32) at (6.0,2.5) {};
\node[font=\tiny] (g32label) at (6.0,2.5) {$\gamma_{3_{2}}$};
\node[] (g3dots) at (6.0,3.375) {$\vdots$};
\node[nodeknown] (g3n) at (6.0,4.5) {};
\node[font=\tiny] (g3nlabel) at (6.0,4.5) {$\gamma_{3_{n}}$};
\node[factor] (f31) at (6.0,1) {};
\node[factor] (f32) at (6.0,2) {};
\node[factor] (f3n) at (6.0,4) {};
\node[textnode] (l4) at (8.0,-0.4) {\footnotesize{\textit{box}}};
\node[textnode] (l4) at (8.0,-0.1) {\footnotesize{\textit{the}}};
\node[nodeknown] (p4) at (8.0,0.5) {};
\node[font=\tiny] (p4label) at (8.0,0.5) {$\lambda_{4}$};
\node[nodeunknown] (c41) at (7.5,1) {};
\node[font=\tiny] (c41label) at (7.5,1) {$\phi_{4_{1}}$};
\node[nodeunknown] (c42) at (7.5,2) {};
\node[font=\tiny] (c42label) at (7.5,2) {$\phi_{4_{2}}$};
\node[nodeunknown] (c4n) at (7.5,4) {};
\node[font=\tiny] (c4nlabel) at (7.5,4) {$\phi_{4_{n}}$};
\node[nodeknown] (g41) at (8.0,1.5) {};
\node[font=\tiny] (g41label) at (8.0,1.5) {$\gamma_{4_{1}}$};
\node[nodeknown] (g42) at (8.0,2.5) {};
\node[font=\tiny] (g42label) at (8.0,2.5) {$\gamma_{4_{2}}$};
\node[] (g4dots) at (8.0,3.375) {$\vdots$};
\node[nodeknown] (g4n) at (8.0,4.5) {};
\node[font=\tiny] (g4nlabel) at (8.0,4.5) {$\gamma_{4_{n}}$};
\node[factor] (f41) at (8.0,1) {};
\node[factor] (f42) at (8.0,2) {};
\node[factor] (f4n) at (8.0,4) {};
\node[textnode] (pt1) at (7.625,-1.25) {\footnotesize{DT}};
\node[textnode] (pt2) at (8.375,-1.25) {\footnotesize{NN}};
\node[textnode] (pt3) at (6.0,-1.25) {\footnotesize{IN}};
\node[textnode] (pt4) at (3.625,-1.25) {\footnotesize{DT}};
\node[textnode] (pt5) at (4.375,-1.25) {\footnotesize{NN}};
\node[textnode] (pt6) at (0,-1.25) {\footnotesize{VB}};
\node[textnode] (pt7) at (8.0,-2.0) {\footnotesize{NP $\left(\lambda_{4}\right)$}};
\draw[] (pt1) to (7.625,-1.625) to (8.0,-1.625) to (pt7);
\draw[] (pt2) to (8.375,-1.625) to (8.0,-1.625) to (pt7);
\node[textnode] (pt8) at (4.0,-2.0) {\footnotesize{NP $\left(\lambda_{2}\right)$}};
\draw[] (pt4) to (3.625,-1.625) to (4.0,-1.625) to (pt8);
\draw[] (pt5) to (4.375,-1.625) to (4.0,-1.625) to (pt8);
\node[textnode] (pt9) at (7.0,-2.75) {\footnotesize{PP $\left(\lambda_{3}\right)$}};
\draw[] (pt3) to (6.0,-2.375) to (7.0,-2.375) to (pt9);
\draw[] (pt7) to (8.0,-2.375) to (7.0,-2.375) to (pt9);
\node[textnode] (pt10) at (5.5,-3.5) {\footnotesize{NP $\left(\lambda_{1}\right)$}};
\draw[] (pt8) to (4.0,-3.125) to (5.5,-3.125) to (pt10);
\draw[] (pt9) to (7.0,-3.125) to (5.5,-3.125) to (pt10);
\node[textnode] (pt11) at (2.75,-4.25) {\footnotesize{VP $\left(\lambda_{0}\right)$}};
\draw[] (pt6) to (0,-3.875) to (2.75,-3.875) to (pt11);
\draw[] (pt10) to (5.5,-3.875) to (2.75,-3.875) to (pt11);
\end{tikzpicture}
\caption{The DCG for the expression ``retrieve the ball inside the box" aligned with the corresponding parse tree.  The observed and unknown variable nodes in the factor graph are shown in white and gray, respectively.}
\label{fig:nlu-example}
\end{figure}

The approach to
natural-language understanding of robot instructions in this paper relies on
variations of the Distributed Correspondence Graph (DCG)~\citep{howard14,
hemachandra15, boteanu16a, barber16a, broad17a, oh17a, patki18a, arkin2018iser,
patki2019a, arkin20a}.  The DCG and the Hierarchical Distributed Correspondence
Graph (HDCG)~\citep{chung15}, Adaptive Distributed Correspondence Graph
(ADCG)~\citep{paul16a}, and Hierarchical Adaptive Distributed Correspondence
Graph (HADCG)~\citep{paul18} variations of the model formulate the problem of
natural-language understanding as probabilistic inference in a factor graph
using models to approximate the conditional probabilities of a correspondence
variable $\phi$ in the context of language $\Lambda$, symbols $\Gamma$, and the
environment model $S$ from corpora of annotated examples.  Such models have the ability to generalize to instructions that are not explicitly represented in the corpora by independently learning the conditional probabilities of concepts for linguistic structures like nouns (e.g., ``the box'' and `` briefcase''), prepositions (e.g., ``near'' and ``inside''), and verbs (e.g., ``drive'' and ``pick up'') and leveraging the structure of the parse tree for probabilistic inference.
Natural-language instructions are converted to parse tree representations to construct DCGs in this framework as studied and more explicitly demonstrated in \citet{paul18}.  Inference is
made efficient by assuming conditional independence across linguistic and
symbolic constituents.  The HDCG, ADCG, and HADCG employ approximations of the
symbolic representation $\Gamma$ informed by expressed symbols to further
improve the efficiency of probabilistic inference.  Consider the example
illustrated in Figure~\ref{fig:nlu-example}.  This figure shows the DCG and the
corresponding parse tree for the instruction ``retrieve the ball inside the
box''.  Each of the symbols $\gamma_{i_{j}}$ in this graph represent objects,
spatial relations, actions, and other concepts needed to interpret the meaning
of the instruction.  The connection to the environment model $S$ is
implicit in this illustration of the DCG.

Assuming that the symbol-grounding model is accurately trained from corpora of annotated examples, there are
three potential outcomes that are observed to be conditioned on the information contained in the robot's model of the environment.  To illustrate such outcomes, we consider the example utterance ``the ball inside the box'' from Figure~\ref{fig:intro}.  First, if the environment model is known or partially
known and there is only one ``ball'' object that is uniquely identified as being
inside of a ``box'' object, then the inference procedure returns a behavior that
describes the navigation and manipulation action with respect to those
physically grounded objects.  Second, if the environment model is known or
partially known and there is more than one ``ball'' object that satisfies the
relationship of being inside of a ``box'' object, then the model would express
this ambiguity to the user through a different set of symbols.  The third
case, which is the focus of this paper, is that the environment model is
partially known and there are no ``ball'' objects that satisfy the relationship
of being inside of a ``box'' object.  The results of the inference procedure are
no symbols that express a relationship to this unknown object because the object
and the corresponding actions and spatial relations are not part of the symbolic
representation $\Gamma$.  However, the user implies that there is an
object that satisfies a specific relationship, indicating that their
representation of the world is richer in this dimension than the robot's
environment model.  We therefore approach the problem of natural-language
understanding as a two-part process that first exploits the environment-related
information explicit or implicit in the utterance to build a distribution over possible
world models that now contains hypothesized objects and spatial relationships
that were missing in the incomplete world model. This is followed by a step that
infers the behaviors in the context of this distribution over environment models
that is informed by sensor observations and the information gleaned from the
first step.  The first and second steps, described as annotation inference and
behavior inference, respectively, use specially adapted symbolic representations
suitable for each problem.  We describe the space of symbols in the annotation and behavior inference models as in
\citet{paul18}, which expresses each space as the union of different symbols and
their constituents.

For annotation inference, we assume a symbolic representation that is
independent of an environment model because it is meant to inform the state of
the world, whereas behavior inference exploits the synthesized environment model
from the semantic mapping process using the outputs of annotation inference and
past sensor observations.  The symbolic representation for annotation inference
is defined by four different types of symbols.  First, a space of object classes
$\Gamma^{\mathcal{O_{C}}}$ is defined by an object type
$\mathcal{O_{C}}$.  These symbols represent the meaning of noun phrases like
``box'' or ``ball'' in the aforementioned example.  Note that object class
symbols do not correspond to specific instances of these objects in the
environment model. Object class symbols bridge the gap between the diversity of
language and the space of semantic object classes interpretative by the robot.
Second, a space of locations $\Gamma^{\mathcal{L}}$ corresponds to physical
locations in the environment model defined by a location type
$\mathcal{L}$.  These are similar to objects in that they occupy some bounded
region in the environment model, but they are not observed using traditional
object detectors to infer their location.  Third, a space of spatial
relation classes $\Gamma^{\mathcal{S}}$ is defined by a spatial relation label $\mathcal{S}$.  Spatial relation classes can be used to
represent noun phrases like ``the left'' or ``front''.  Next, we define a space of region
types $\Gamma^{\mathcal{R_{S}}}$ by a spatial relationship
class $\mathcal{S}$ for every object class in
$\mathcal{O_{C}}$.  Region classes are used to represent prepositional phrases
like ``inside the box'' or ``to the left of the ball'.   Lastly, a space of
relationships $\Gamma^{\mathcal{R_{C}}}$ is defined by a spatial relationship
 $\mathcal{S}$ between a pair of
object classes in $\mathcal{O_{C}}$.  Relationship
classes are used to represent spatial relationships between pairs of objects
that occur in noun phrases like ``the ball inside the box'' or ``the car behind
the garage''.  When environments are partially observed, the extraction of these
relationships from language can inform the distribution of objects and locations in
the environment model constructed from visual perception.  These relationships
can be extracted implicitly from instructions like ``pick up the ball from
inside the box,'' or the last of part of a dialogue that begins with an ambiguous
instruction (e.g., ``pick up the ball'' in an environment without a ball), a question
that the robot poses to the human (``where is the ball''), and a response
describing one or more of these relationship (``the ball is inside the box
underneath the table on the far side of the room'').  The space of symbols for
annotation inference $\Gamma^{\mathcal{A}}$ is defined as the union of object, spatial relation, region, and relation symbols
\begin{subequations}
    \begin{align}
        \Gamma^{A} &= \left\{ \Gamma^{\mathcal{O_{C}}} \cup \Gamma^{\mathcal{L}} \cup\Gamma^{\mathcal{S}} \cup \Gamma^{\mathcal{R}_{\mathcal{S}}} \cup \Gamma^{\mathcal{R}_{\mathcal{C}}} \right\}\label{eqn:annotation-symbol-space},\\
        \intertext{where the constituent symbol spaces are defined as}
        \Gamma^{\mathcal{O_{C}}} &= \left\{ \gamma_{o_{c_{i}}} \vert o_{c_{i}} \in \mathcal{O_{C}} \right\} \label{eqn:annotation-object-classes}\\
        \Gamma^{\mathcal{L}} &= \left\{ \gamma_{l_{i}} \vert l_{i} \in \mathcal{L} \right\} \label{eqn:annotation-locations}\\
        \Gamma^{\mathcal{S}} &= \left\{ s_{i} \vert s_{i} \in \mathcal{S} \right\}\label{eqn:annotation-spatial-relations}\\
        \Gamma^{\mathcal{R}_{\mathcal{S}}} &= \left\{ \gamma_{o_{c_{{j}}}}^{s_{i}} \vert s_{i} \in \mathcal{S}, o_{c_{j}} \in \mathcal{O_{C}} \right\} \label{eqn:annotation-region-classes}\\
        \Gamma^{\mathcal{R}_{\mathcal{C}}} &= \left\{
        \gamma_{o_{c_{{j}}},o_{c_{{k}}}}^{s_{i}} \vert s_{i} \in \mathcal{S}, o_{c_{j}} \in \mathcal{O_{C}}, o_{c_{k}} \in \mathcal{O_{C}} \right\}.
        \label{eqn:annotation-relation-classes}
    \end{align}
\end{subequations}

For behavior inference, we assume a symbolic representation formed from objects,
spatial relations, and regions.  The space of objects $\Gamma^{\mathcal{O}}$ is
described by the object $\mathcal{O}$ from the environment model
$S$.  These symbols represent actual objects that the robot has a model
of and enables unique reference of objects by their semantic class (``the box''
in an environment model with only one box) or by their spatial, temporal, or
other relationship to other objects in an environment not uniquely defined by
their semantic class (``the box underneath the table'' in an environment with
several boxes but only one table or bench).  The space of regions
$\Gamma^{\mathcal{R}_{\mathcal{O}}}$ is defined as the composition of a spatial relation with an object.  Notice that the space of regions is defined by the uniquely defined objects in the environment model.  The spaces of spatial relation and object classes are also
applied for behavior inference.  The space of modes $\Gamma^{\mathcal{M}}$ defines variations of actions that influence the cost functions used to determine the optimal plan.  Mode types such as ``safe'' and ``quickly'' given the same set of goals may, for example, produce different actions in the same environment. Lastly, the space of actions
$\Gamma^{\mathcal{A}_{\mathcal{O}}}$ is defined by a unique
object, spatial relation, reference object, and action type $\mathcal{A_{O}}$.  The space of spatial relation classes is additionally
used in behavior inference.  The space of symbols for behavior inference
$\Gamma^{\mathcal{B}}$ is defined by an object, spatial relation
class, region, and action symbols.
\begin{subequations}
    \begin{align}
    \Gamma^{B} &= \left\{ \Gamma^{\mathcal{O}} \cup \Gamma^{\mathcal{S}} \cup \Gamma^{\mathcal{O_{C}}}
        \cup \Gamma^{\mathcal{R}_{\mathcal{O}}} \cup \Gamma^{\mathcal{M}} \cup \Gamma^{\mathcal{A}_{\mathcal{O}}} \right\},\\
        \intertext{where the new constituent symbol spaces are defined as}
        \Gamma^{\mathcal{M}} &= \left\{ m_{i} \vert m_{i} \in \mathcal{M} \right\}\label{eqn:experiments-modes}\\
        \Gamma^{\mathcal{O}} &= \left\{ \gamma_{o_{i}} \vert o_{i} \in \mathcal{O} \right\}\label{eqn:experiments-objects}\\
        \Gamma^{\mathcal{R}_{\mathcal{O}}} &= \left\{ \gamma_{o_{j}}^{s_{i}} \vert s_{i} \in \mathcal{S}, o_{j} \in \mathcal{O} \right\} \label{eqn:experiments-regions}\\
        \Gamma^{\mathcal{A}_{\mathcal{O}}} &=  \left\{ \gamma_{o_{j},o_{k}}^{s_{i},a_{o_{m},m_{n}}} \vert s_{i} \in \mathcal{S}, o_{j} \in \mathcal{O}, o_{k} \in \mathcal{O}, a_{o_{m}} \in \mathcal{A_{O}}, m_{n} \in \mathcal{M},  \right\} \label{eqn:experiments-actions}\\
        \label{eqn:experiments-behavior-inference-space}
    \end{align}
\end{subequations}

These symbols generally describe a uniform representation of the search space for the experiments described in Section~\ref{sec:experiments}.  Note that the complexity of these symbolic representations can be increased or decreased depending on the diversity of tasks explored in the application of this framework.  Existing works~\citep{paul16a, paul18} consider grounding natural-language expressions in relation to inferred object sets, however this is not studied here.

\subsection{Semantic Mapping}

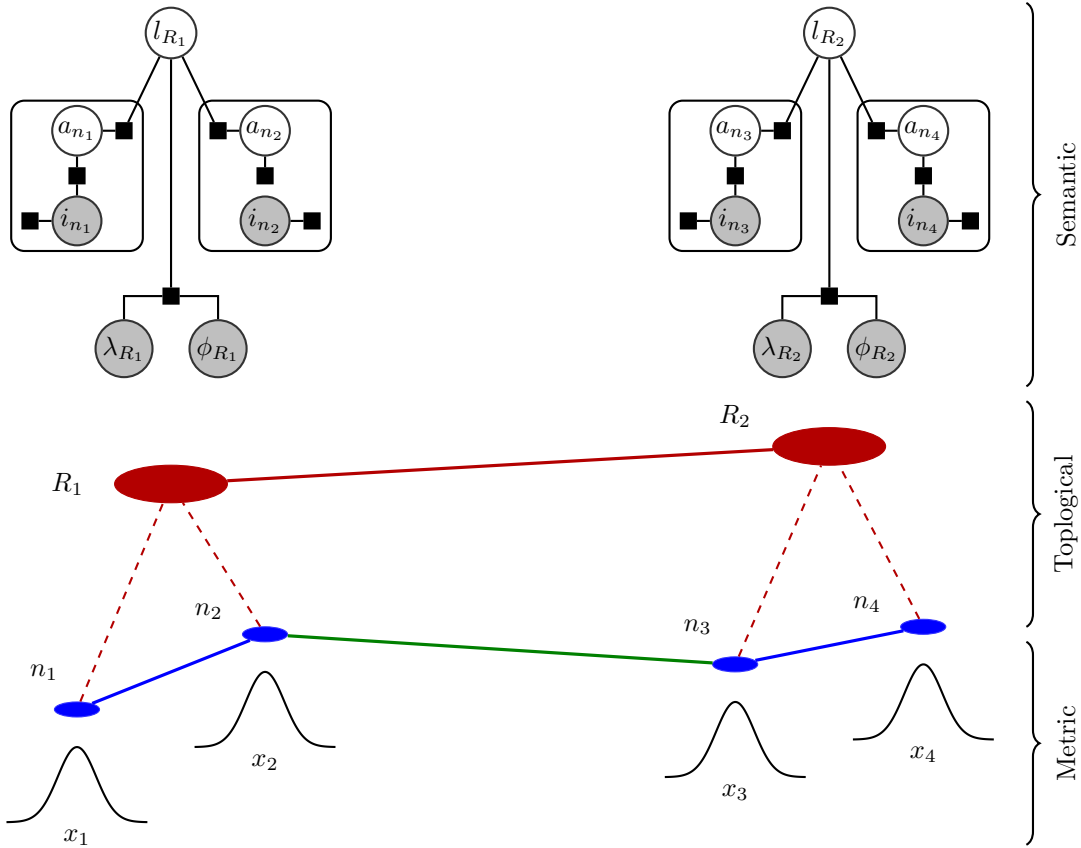
\begin{figure}[!t]
    \centering
    \begin{tikzpicture}[xscale=2.5,yscale=2.0]
        \tikzstyle{latent}=[circle, minimum size = 6mm, inner sep=1pt, thick, draw = black!80, node distance = 10mm]
        \tikzstyle{observed}=[circle, minimum size = 6mm, inner sep=1pt, thick, draw = black!80, node distance = 10mm,fill=gray!50]
        \tikzstyle{factor}=[rectangle, minimum size = 2mm, inner sep=0pt, thick, draw = black, node distance = 10mm,fill=black]
        \tikzstyle{connect}=[thick]

        \tikzstyle{node}=[ellipse, minimum width = 6mm, minimum height = 2mm, inner sep=0mm, draw = blue!80, fill=blue]
        \tikzstyle{region}=[ellipse, minimum width = 15mm, minimum height = 5mm, inner sep=0mm, draw = black!30!red, fill=black!30!red]

        \tikzstyle{intraregion-connect}=[very thick,blue]
        \tikzstyle{interregion-connect}=[very thick,black!50!green]
        \tikzstyle{region-connect}=[very thick,black!30!red]
        \tikzstyle{region-node-connect}=[thick, black!30!red,dashed]

        \tikzstyle{notice}=[draw, rectangle callout, callout relative pointer={#1}]
        {
        \node[node] (n1) at (0.5,0.5) {};
        \node[node] (n2) at (1.5,1) {};
        \node[node] (n3) at (4.0,0.8) {};
        \node[node] (n4) at (5.0,1.05) {};

        \node[region] (r1) at (1.0,2.0) {};
        \node[region] (r2) at (4.5,2.25) {};

        \node at (0.45,2.0) {$R_1$};
        \node at (4.0,2.45) {$R_2$};

        \path[-] (n1) edge [intraregion-connect] (n2);
        \path[-] (n2) edge [interregion-connect] (n3);
        \path[-] (n3) edge [intraregion-connect] (n4);

        \path[-] (r1) edge [region-connect] (r2);

        \path[-] (n1) edge [region-node-connect] (r1);
        \path[-] (n2) edge [region-node-connect] (r1);
        \path[-] (n3) edge [region-node-connect] (r2);
        \path[-] (n4) edge [region-node-connect] (r2);

        \begin{scope}[scale=0.5,shift={($(n1) -(0,1.5)$)}]
            \draw[thick] plot[samples=100,domain=-0.75:0.75] ({\x},{exp(-2*\x*\x*6.0)});
            \node at (0,-0.2) {$x_1$};
            \node at (-0.35,2.0) {$n_1$};
        \end{scope}

        \begin{scope}[scale=0.5,shift={($(n2) -(0,1.5)$)}]
            \draw[thick] plot[samples=100,domain=-0.75:0.75] ({\x},{exp(-2*\x*\x*6.0)});
            \node at (0,-0.2) {$x_2$};
            \node at (-0.6,1.8) {$n_2$};
        \end{scope}

        \begin{scope}[scale=0.5,shift={($(n3) -(0,1.5)$)}]
            \draw[thick] plot[samples=100,domain=-0.75:0.75] ({\x},{exp(-2*\x*\x*6.0)});
            \node at (0,-0.2) {$x_3$};
            \node at (-0.4,2.0) {$n_3$};
        \end{scope}

        \begin{scope}[scale=0.5,shift={($(n4) -(0,1.5)$)}]
            \draw[thick] plot[samples=100,domain=-0.75:0.75] ({\x},{exp(-2*\x*\x*6.0)});
            \node at (0,-0.2) {$x_4$};
            \node at (-0.6,1.8) {$n_4$};
        \end{scope}

        \begin{scope}[shift={($(r1|-r2) +(0,1.5)$)}]
            \node[latent] (lr1) at (0,1.25){$l_{R_1}$};
            \node[factor] (flr1) at ($(lr1) - (0.0,1.75)$) {};

            \begin{scope}[shift={($(lr1) - (0.5, 1.25)$)}]
                \node[latent] (a1) at (0,0.6){$a_{n_1}$};
                \node[observed] (l1) at (0,0){$i_{n_1}$};

                \node[factor] (fl1a1) at ($(a1)!0.5!(l1)$) {};
                \node[factor] (fa1) at ($(a1) + (0.25,0)$) {};
                \node[factor] (fl1) at ($(l1) - (0.25,0)$) {};

                \path[-] (l1) edge [connect] (fl1a1);
                \path[-] (a1) edge [connect] (fl1a1);
                \path[-] (a1) edge [connect] (fa1);
                \path[-] (l1) edge [connect] (fl1);

                \draw[connect, rounded corners=5pt] (-0.35,-0.2) rectangle (0.35,0.8);
            \end{scope}

            \begin{scope}[shift={($(lr1) + (0.5, -1.25)$)}]
                \node[latent] (a2) at (0,0.6){$a_{n_2}$};
                \node[observed] (l2) at (0,0){$i_{n_2}$};

                \node[factor] (fl2a2) at ($(a2)!0.5!(l2)$) {};
                \node[factor] (fa2) at ($(a2) - (0.25,0)$) {};
                \node[factor] (fl2) at ($(l2) + (0.25,0)$) {};

                \path[-] (a2) edge [connect] (fl2a2);
                \path[-] (a2) edge [connect] (fa2);
                \path[-] (l2) edge [connect] (fl2);

                \draw[connect, rounded corners=5pt] (-0.35,-0.2) rectangle (0.35,0.8);
            \end{scope}

            \node[observed] (lambda2) at ($(flr1) - (0.25,0.35)$){$\lambda_{R_1}$};
            \node[observed] (phi2) at ($(flr1) + (0.25,-0.35)$){$\phi_{R_1}$};

            \path[-] (lr1) edge [connect] (flr1);

            \path[-] (lr1) edge [connect] (fa1);
            \path[-] (lr1) edge [connect] (fa2);

            \draw[connect] (lambda2)|-(flr1);
            \draw[connect] (phi2)|-(flr1);

        \end{scope}

        \begin{scope}[shift={($(r2) +(0,1.5)$)}]
            \node[latent] (lr2) at (0,1.25){$l_{R_2}$};
            \node[factor] (flr2) at ($(lr2) - (0.0,1.75)$) {};

            \begin{scope}[shift={($(lr2) - (0.5, 1.25)$)}]
                \node[latent] (a3) at (0,0.6){$a_{n_3}$};
                \node[observed] (l3) at (0,0){$i_{n_3}$};

                \node[factor] (fl3a3) at ($(a3)!0.5!(l3)$) {};
                \node[factor] (fa3) at ($(a3) + (0.25,0)$) {};
                \node[factor] (fl3) at ($(l3) - (0.25,0)$) {};

                \path[-] (l3) edge [connect] (fl3a3);
                \path[-] (a3) edge [connect] (fl3a3);
                \path[-] (a3) edge [connect] (fa3);
                \path[-] (l3) edge [connect] (fl3);

                \draw[connect, rounded corners=5pt] (-0.35,-0.2) rectangle (0.35,0.8);
            \end{scope}

            \begin{scope}[shift={($(lr2) + (0.5, -1.25)$)}]
                \node[latent] (a4) at (0,0.6){$a_{n_4}$};
                \node[observed] (l4) at (0,0){$i_{n_4}$};

                \node[factor] (fl4a4) at ($(a4)!0.5!(l4)$) {};
                \node[factor] (fa4) at ($(a4) - (0.25,0)$) {};
                \node[factor] (fl4) at ($(l4) + (0.25,0)$) {};

                \path[-] (l4) edge [connect] (fl4a4);
                \path[-] (a4) edge [connect] (fl4a4);
                \path[-] (a4) edge [connect] (fa4);
                \path[-] (l4) edge [connect] (fl4);

                \draw[connect, rounded corners=5pt] (-0.35,-0.2) rectangle (0.35,0.8);
            \end{scope}

            \node[observed] (lambda4) at ($(flr2) - (0.25,0.35)$){$\lambda_{R_2}$};
            \node[observed] (phi4) at ($(flr2) + (0.25,-0.35)$){$\phi_{R_2}$};

            \path[-] (lr2) edge [connect] (flr2);

            \path[-] (lr2) edge [connect] (fa3);
            \path[-] (lr2) edge [connect] (fa4);

            \draw[connect] (lambda4)|-(flr2);
            \draw[connect] (phi4)|-(flr2);
        \end{scope}

        \draw [thick,decorate,decoration={brace,amplitude=5pt,mirror},xshift=-10pt,yshift=0pt] (5.9,-0.4) -- (5.9,0.95)node [black,midway,xshift=15pt,rotate=90] {Metric};
        \draw [thick,decorate,decoration={brace,amplitude=5pt,mirror},xshift=-10pt,yshift=0pt] (5.9,1.05) -- (5.9,2.55)node [black,midway,xshift=15pt,rotate=90] {Toplogical};
        \draw [thick,decorate,decoration={brace,amplitude=5pt,mirror},xshift=-10pt,yshift=0pt] (5.9,2.65) -- (5.9,5.2)node [black,midway,xshift=15pt,rotate=90] {Semantic};
        }
    \end{tikzpicture}
    \caption{The semantic graph $S_t$ models the metric (spatial),
    topological, and semantic properties of the environment. The topology
    $G_t$ consists of a set of nodes that represent objects and locations in
    the environment that have either been observed or visited by the robot,
    or are hypothesized based upon language. Each node is associated with a
    region $R_i$ that represents spatially coherent areas in the
    environment. Edges in the topology, where intra- and inter-region edges
    are rendered in blue and green, respectively, and model spatial
    relationships between nodes, reflect the robot's motion, observations,
    or constraints inferred from language. The edge rendered in red between the two regions is for visualization purposes only. The topology induces a pose graph, which corresponds to the metric map $X_t$.  The semantic layer
    $L_t$ models the category $l_{R_j}$ of each region $R_j$, which we infer from language $\lambda_{R_j}$ and node-level scene and object classifiers $i_{n_k}$.}
    \label{fig:semantic-graph}
\end{figure}

Integral to our approach is the ability to exploit environment-related
information implicit in the natural-language instruction to maintain a
distribution over the world model. Here, we detail a Bayesian filtering-based
approach to maintaining this distribution. We represent the world model as a
modified \emph{semantic graph}~\citep{walter13} $S_t = \{G_t, X_t, L_t\}$, a
hybrid metric, topological, and semantic representation of the robot's
environment, which we visualize in Figure~\ref{fig:semantic-graph}.

The topology $G_t$ consists of a set of nodes that represent objects (e.g., boxes, balls, cones, etc.) and places (e.g., offices, lobbies, etc.) in the environment that have either been observed or visited by the robot, or are hypothesized based upon language. Nodes are partitioned into regions $R_i = \{n_1, n_2, \ldots, n_m\}$ that represent spatially
coherent areas (e.g., rooms and hallways), which we refer to as \emph{spatial regions}, as well as individual objects (e.g., boxes), which we refer to as \emph{object regions}. Object regions typically consist of a single node. Edges in the topology model spatial
relationships between nodes, and reflect the robot's motion, or constraints inferred from language. More specifically, similar to pose graph formulations of SLAM~\citep{eustice05, olson06, kaess08}, an edge connects two nodes (locations) that the robot has
transitioned between or nodes (objects or places) that it has observed, as well as nodes for which language indicates the
existence of a specific spatial relation (e.g., that the kitchen is
``down'' the hallway or that there is a ball ``inside'' a box).  We associate a pose
$x_i$ with each node $n_i$. The concatenation of these poses constitutes the metric map
$X_t$. The semantic layer $L_t$ expresses the semantic attributes of each spatial and object region, which include its colloquial name and type $l_{R_i}$.

Annotations that are inferred from a given instruction under our language model provide information regarding
the existence, type, and location (relative to the robot or another landmark) of entities in the environment. We learn a
distribution over possible world models consistent with these annotations by
treating them as observations $\alpha_t$ in a filtering framework, effectively treating language as another sensing modality. We
combine these observations with those from other sensors onboard the robot
(e.g., LIDAR and camera-based object and spatial region appearance observations) $z_t$ to maintain a distribution
over the semantic map:
\begin{subequations} \label{eqn:semantic-map-distribution}
  \begin{align}
    p(S_t \vert \Lambda^t, z^t, u^t) &\approx p(S_t \vert
    \alpha^t, z^t, u^t)\\
    &= p(G_t, X_t, L_t \vert \alpha^t, z^t, u^t)\\
    &= p(L_t \vert X_t, G_t, \alpha^t, z^t, u^t) \, p(X_t \vert G_t, \alpha^t, z^t, u^t) \, p(G_t \vert \alpha^t, z^t, u^t), \label{eqn:semantic-map-distribution-factored}
  \end{align}
\end{subequations}
where we replace the utterance $\Lambda^t$ with the set of inferred annotations
$\alpha_t$ (Sec.~\ref{sec:nlu}). The factorization in the last line models the metric map as being induced by the topology, much like is done with pose graph approaches to SLAM~\citep{kaess08}.

In theory, the set of possible graphs for any given environment is combinatorial, because the number of edges can be exponential in the number of nodes. This suggests that it would be intractable to maintain a full distribution over the set of graphs for all but trivially small environments.
To simplify this complexity, we adopt an assumption made by others~\citep{ranganathan11}) that the distribution is dominated by only a few topologies. More specifically, unless the environment is perceptually aliased, a large number of particles will be inconsistent with the robot's observations and will thus be assigned low likelihood. Conditioning on the environment annotations further supports this assumption as it decreases the probability of edges that are inconsistent with the command. Any structure in the environment will also bound the inter-connectivity of nodes (e.g., walls occluding the robot's LIDAR prevent observations of an adjacent room that would otherwise result in an edge to the node that denotes the robot's pose) as does the limited field-of-view of the robot's sensors, further increasing the number of topologies having near-zero likelihood. Consequently, only a few topologies that are consistent with the observations and annotations will be associated with non-negligible likelihoods. In environments for which this assumption does not hold (e.g., open areas with objects distributed sparsely throughout), our framework can still be used, but the computational cost will increase due to the need to maintain a distribution over the larger set of topologies.

Leveraging this assumption, we use a sample-based representation of the
posterior distribution over graphs $p(G_t \vert \alpha^t, z^t, u^t)$ in Equation~\ref{eqn:semantic-map-distribution-factored}.
Specifically, we maintain the factored semantic map distribution
using a Rao-Blackwellized
particle filter (RBPF)~\citep{doucet00}, where we employ a sample-based
representation of the distribution over the environment topology $p(G_t \vert \alpha^t, z^t, u^t)$ and a Gaussian
representation of the metric (pose) distribution $p(X_t \vert G_t, \alpha^t,
z^t, u^t)$ that is induced by the topology. We parameterize the Gaussian pose
distribution in the canonical (information) form, i.e., in terms of the
information (inverse covariance) matrix and information vector, as opposed to
the standard covariance form. The structure of the information matrix follows
that of the topology~\citep{eustice05, walter07, kaess08}, which is sparse for typical environments,
thereby allowing us to exploit this sparsity to improve the efficiency of
inference~\citep{thrun04, eustice05, kaess08, walter07}.

We model the semantic distribution $p(L_t \vert
X_t, G_t, \alpha^t, z^t, u^t)$ using a factor graph, which we visualize in plate notation in
Figure~\ref{fig:semantic-graph} (top). The factor graph includes a random variable for each region that
expresses its type (i.e., category) $l_{R_i}$. The factor graph includes variables that express language-based annotations, where the node $\lambda_{R_i}$ denotes a potential reference to region $R_i$ in the
natural-language utterance, while $\phi_i$ is a Boolean variable that specifies
whether or not the reference corresponds to the region. For each node $n$ in the
region, the robot makes indirect observations of the region's appearance $a_n$,
which is coupled with the region's category, via image-based scene
classifiers $i_n$.

Together, this gives rise to the following
parameterization of the posterior
\begin{equation}
    \mathcal{P}_t = \left\{P_t^{(1)}, P_t^{(2)}, \ldots, P_t^{(n)} \right\},
\end{equation}
where each particle $P_t^{(i)}$ includes a candidate topology $G_t^{(i)}$, Gaussian pose graph $X_t^{(i)}$, semantic map $L_t^{(i)}$, and particle weight $w_t^{(i)}$
\begin{equation}
    P_t^{(i)} = \left\{G_t^{(i)}, X_t^{(i)}, L_t^{(i)}, w_t^{(i)} \right\}
\end{equation}
\begin{algorithm}[t]
    \DontPrintSemicolon
    \KwIn{$P_{t-1} =\left\{P_{t-1}^{(i)}\right\}$, and $\left(u_t, z_t, a_t, \alpha_t\right)$,
        where $P_{t-1}^{(i)} = \left\{ G_{t-1}^{(i)}, X_{t-1}^{(i)}, L_{t-1}^{(i)}, w_{t-1}^{(i)} \right\}$}
    \BlankLine
    \KwOut{$P_t = \left\{P_t^{(i)}\right\}$}
    \BlankLine
    \For{ $i = 1$ to $n$} {
      \begin{enumerate}
       \item Employ proposal distribution $p(G_t \lvert G_{t-1}^{(i)}, z^{t-1}, u^t, \alpha^t)$ to propagate the graph sample according to odometry on $u_t$, the inferred region labels $l_t$, and scene classifications $z_t$.
	\begin{enumerate}
	 \item Sample region allocation%
	 \item Sample region edges
	 \item Merge newly connected regions
	\end{enumerate}
      \item Update the Gaussian distribution over the node poses
        $X_t^{(i)}$  according to the constraints induced by the newly-added graph edges.\;
      \item Update the factor graph representing semantic properties for the
	 topology based on appearance observations $z_t$ and language-based annotations
         $\alpha_t$.\;
      \item Compute the new particle weight $w_t^{(i)}$ based upon the
        previous weight $w_{t-1}^{(i)}$ and the metric data $z_t$.\;

      \end{enumerate}

    }
    \BlankLine
     Normalize weights and resample if needed, i.e., if $N_\text{eff} < n/2$, where $N_\text{eff} = \frac{1}{\sum_{i=0}^n{w_i^2}}$.
\caption{Semantic Mapping Algorithm}
\label{alg:semantic-mapping-algorithm}
\end{algorithm}
The robot detects and labels objects in the environment based on camera observations, using a neural object detector built on top of the YOLO~V3 architecture~\citep{yolov3}. A neural image-based scene classifier also provides noisy observations of the semantic class of spatial regions. In addition to assigning labels, we use the inferred scene classes to segment spatial regions, using the presence of conflicting appearance labels to suggest a region segmentation. As we describe next, we couple this with a spectral clustering-based measure of the spatial coherence between laser scans in order to refine the boundaries of spatial regions. Since this clustering reasons over the similarity of laser scans associated with different nodes, it typically segments a region after the robot last observes it. However, using inferred scene classes allows the method to be aware of new regions when they are first observed, enabling us to
immediately evaluate each particle's likelihood based on the observation of
region appearance. In turn, we can down-weight particles that are
inconsistent with the actual layout of the world sooner, reducing the
number of actions the robot must take to satisfy the command.

We maintain each particle through the three steps of the RPBF as detailed in Algorithm~\ref{alg:semantic-mapping-algorithm}. First, we
propagate the topology by sampling modifications to the graph when the
robot receives new sensor observations or annotations. Second, we perform a
Bayesian update to the pose distribution based upon the sampled
modifications to the underlying graph. Third, we update the weight of each
particle based on the likelihood of generating the given observations, and
resample as needed to avoid particle depletion. We now outline this process
in more detail.

\subsubsection{The Proposal Distribution}

We compute the predictive posterior over the graph $G_t$ using a proposal that is the distribution over the current graph given the previous graph $G_{t-1}$, sensor data (excluding the current time step),
appearance data, odometry, and language
\begin{equation} \label{eqn:proposal2}
    p(G_t \vert G_{t-1}^{(i)}, z^{t-1}, u^t, \alpha^t).
\end{equation}
For each particle $P_{t-1}^{(i)}$, we update the topology $G_{t-1}^{(i)}$
according to the robot's motion, annotations inferred from language, and
environment observations. In particular, we first add a node $n_t$ that denotes
the robot's current pose and an edge between it and the previous node that
encodes the temporal (odometry) constraint between the two poses.\footnote{We add a node corresponding to the robot's current pose when it observes a new object or spatial region, or has traveled more than a specified distance. In practice, we typically use $1.0$\,m as the distance threshold.} %
We
initially assign the new node to the nearest region, which most often is that of
the previous node, resulting in an intermediate graph
$G_t^{(i)-}$.\footnote{This and the remainder of this discussion apply to
individual particles, however we remove the particle label for readability.} We
then propose modifications to the graph $\Delta_t^{(i)} =
\{\Delta_{t, \alpha}^{(i)}, \Delta_{t, z}^{(i)}\}$ based upon
appearance observations $\Delta_{t, z}^{(i)}$ and
natural-language annotations $\Delta_{t, \alpha}^{(i)}$:
\begin{equation} \label{eqn:proposal2-factored}
 p(G_t^{(i)} \vert G_{t-1}^{(i)}, z^{t-1}, u^t, \alpha^t) = p(\Delta_{t, \alpha}^{(i)} \vert G_t^{(i)-}, \alpha^t)\, p(\Delta_{t, z}^{(i)} \vert G_t^{(i)-}, z^{t-1})\, p(G^{(i)-} \vert G_{t-1}^{(i)}, u_t),
\end{equation}
where we define the three distributions on the right-hand side below.
This formulation updates the graph $G_t^{(i)-}$ with modifications $\Delta_t^{(i)}$ that can include the addition and deletion of nodes and regions, as well as the addition of edges that model spatial relations inferred from environment observations and natural language-based annotations. We sample graph modifications from two independent proposal distributions, one for those that reflect annotations $\alpha_t$ and the other that reflects traditional observations $z_{t-1}$.

\paragraph{Updating the current spatial region}

The term $p(G^{(i)-} \vert G_{t-1}^{(i)}, u_t)$ in Equation~\ref{eqn:proposal2-factored} is the distribution that follows from adding the node and edge that account for the robot's motion. which results in the intermediate graph $G_t^{(i)-}$. We then probabilistically bisect the robot's current spatial region $R_c$ using the spectral
clustering method proposed by~\citet{blanco06}.
We generate the similarity matrix based on the overlap between the laser scans associated with each pair
of nodes in the spatial region. Equation~\ref{eqn:segmentation_likelihood} defines the likelihood of
bisecting the region, which is based on the normalized cut value $N_c$ of the
graph involving the proposed segments. The likelihood of accepting a
proposed segmentation increases as the value of $N_c$ decreases, i.e., as the separation
of the two segments increases (minimizing the inter-region similarity).
\begin{equation} \label{eqn:segmentation_likelihood}
  P(s/N_{cut})=\frac{1}{(1+\alpha{N_{c}^3})}
\end{equation}
The result is that areas that are more spatially distinct have a greater probability of being represented as separate spatial regions, i.e., more particles will model these regions as distinct. If a particle chooses to segment the current region, it creates a new spatial region $R_i$ that does not include the newly added node.

\paragraph{Modifying the graph based on observations} In the event that the algorithm creates a new region $R_i$ via segmentation, it then considers connecting the new region to existing regions in the topology, including those that are hypothesized based on language.  The method samples edges to observed regions using a spatial distribution that is a function of the regions' constituent nodes. We refer to these modifications to each particle as $\Delta_{t, z}^{(i)}$, giving rise to the corresponding distribution from Equation~\ref{eqn:proposal2-factored} %
\begin{equation} \label{eqn:edge_proposal}
        p(\Delta_{t, z}^{(i)} \vert G_t^{(i)-}, z^{t-1}, u^t, \alpha^{t-1})
        = \prod_{j:e_{ij} \notin E_t^-}  p(G_t^{ij} \vert G_t^{(i)-}, z^{t-1}, u^t, \alpha^{t-1})%
\end{equation}
Here, we assume that additional edges expressing
constraints that involve the current node $e_{ij} \notin E_t^-$ are
conditionally independent.

We model the spatial distribution prior in terms of
the distance $d_{ij}$ between the nodes in each of the two regions that are nearest to the center of the region
\begin{subequations} \label{eqn:spatial_distance_proposal}
    \begin{align}
        p(G_t^{ij} \vert G_t^{(i)-}, z^{t-1}, u^t, \alpha^{t-1}) &=
        \int_{X^-_t} p(G_t^{ij}  \vert X_t^-,
        G_t^{(i)-}, z^{t-1}, u^t, \alpha^{t-1}) \, p(X_t^- \vert G_t^{(i)-}, z^{t-1}, u^t, \alpha^{t-1})\\
        &\approx \int_{d_{ij}} p(G_t^{ij} \vert
        d_{ij}, G_t^{(i)-}, z^{t-1}, u^t, \alpha^{t-1})\, p(d_{ij} \vert G_t^{(i)-}, z^{t-1}, u^t, \alpha^{t-1}).
    \end{align}
\end{subequations}
The conditional distribution $p(G_t^{ij} \vert
d_{ij}, G_t^{(i)-}, z^{t-1}, u^t, \alpha^{t-1})$ expresses the likelihood of adding an edge between spatial regions $R_i$ and
$R_j$ based upon the location of their mean nodes. We represent the distribution
for a particular edge between regions $R_i$ and $R_j$ with
distance $d_{ij} = \lvert \bar{X}_{R_i} - \bar{X}_{R_j}\rvert_2$  as
\begin{equation}
    p(G_t^{ij} \vert
    d_{ij}, G_t^{(i)-}, z^{t-1}, u^t, \alpha^{t-1}) = p(G_t^{ij} \vert d_{ij})
    \propto \frac{1}{1+\gamma d_{ij}^2},
\end{equation}
where $\gamma$ expresses a distance bias.\footnote{In practice, we have found $\gamma = 0.3$ to work well empirically.} We approximate the
distance prior $p(d_{ij} \vert G_t^{(i)-}, z^{t-1}, u^t, \alpha^{t-1})$
with a folded Gaussian distribution.

\paragraph{Merging with observed or hypothesized regions}
After adding a new spatial region $R_i$ and any inter-region edges, we then evaluate whether to merge the region with any of the regions to which it is connected. We merge the region with an existing (connected) region if the modes of their distributions over region type (i.e., category) are the same.
This
results in regions being merged when the robot revisits locations
already represented in the graph. This merge
process is designed such that the complexity of the
topology increases only when the robot explores new areas, leading to more
efficient region edge proposals as well as more compact language groundings.

If the newly added spatial region was not merged with one that was previously visited, or when an object region was added based upon camera observations, we check whether it
matches a region that was previously hypothesized based on an annotation
(for example, a toolbox that the robot detected may be the same one that was hypothesized earlier based on the instruction).
We do so by sampling a grounding to any unobserved regions in the topology
using a Dirichlet process prior.
If this results in a grounding to an existing hypothesized region,
we remove the hypothesized region and adjust the topology accordingly,
resampling any edges to yet-unobserved regions.  For example, if an
annotation suggested the existence of a ``toolbox inside the garage,'' and
we grounded the robot's current region to the hypothesized garage, we
would reevaluate the ``inside'' relation for the hypothesized toolbox with
respect to this detected garage.

\paragraph{Modifying the graph based on natural language}

In the third part of the proposal step (Eqn.~\ref{eqn:proposal2-factored}), we sample modifications to the graph for each particle based on (the possibly empty) set of annotations \mbox{$\alpha_t=\{\alpha_{t,j}\}$} using a factored form of the distribution:%
\begin{equation}
  p(\Delta_{t, \alpha}^{(i)} \vert S_{t}^{(i)-}, \alpha_t)  = \prod_j
  p(\Delta_{\alpha_{t,j}}^{(i)} \vert S_{t}^{(i)-}, \alpha_{t,j}). \label{eq:proposal_annotation}
\end{equation}
An annotation $\alpha_{t,j}$ contains a spatial relation and figure when
the language describes one region (e.g., ``go to the elevator lobby''), and
an additional landmark when the language describes the relation between two
regions (e.g., ``retrieve the wrench from inside the toolbox'' or ``get the mug from the kitchen''), which may be spatial or object regions. We use a
likelihood model over the spatial relation to sample landmark and figure
pairs for the grounding. This model employs a Dirichlet process prior that
accounts for the fact that the annotation may refer to regions that exist
in the map or to regions that are currently unknown. For each landmark and/or region that is sampled as being new, we add a new node to the graph and assign the node to its own region. We then add an edge between the figure and landmark and define the metric constraint associated with this
edge based on the spatial relation. We represent the distribution associated with this constraint as a Gaussian with a mean expressed as a linear function of features that describe the locations of the regions, their bounding boxes,
and the robot's location at the time of the utterance. We learn this function along with the covariance of the Gaussian using a natural-language corpus of spatial relations~\citep{tellex11}.

\subsubsection{Updating the metric map based on new edges}

The proposal step results in an update to the graph $G_t^{(i)}$ associated with each particle that includes the addition of a node representing the robot's current pose, the addition of edges, and the possible creation and merging of regions. These modifications need to then be reflected in the distribution over poses. To that end, we first augment the pose vector $X_t^{-}$ to include the  robot's current pose. We then incorporate the relative pose constraints expressed by the new edges, including the temporal (odometry) edge between the current and previous poses, into the Gaussian representation for the marginal distribution over the pose history
\begin{equation}
    p(X_t \vert G_t, \alpha^t, z^t, u^t) = \mathcal{N}^{-1}(X_t; \Sigma_t^{-1}, \eta_t),
\end{equation}
where $\Sigma_t^{-1}$ is the information (inverse covariance) matrix and $\eta_t$ is the information vector that together parameterize the canonical form of the Gaussian.\footnote{We maintain separate parameters for each particle, but omit the superscript indicating the particle ID for readability.} Key to this parameterization is that it corresponds to a Gaussian Markov random field, whereby the structure of the information matrix is given by the topology of the underlying graph. Specifically, the off-diagonal blocks of the information matrix relating pose $x_i$ and pose $x_j$ are non-zero if and only if there is an edge in the graph between the corresponding nodes $n_i$ and $n_j$. For any new edge added to the graph during the proposal step, we update the corresponding entries of the information matrix following the standard information filtering procedure~\citep{eustice05,walter07,kaess08}. We refer the reader to \citet{walter07} for a description of this process. Critically, the sparsity of the information matrix is determined by the underlying graph $G_t$, which is generally sparse for typical environments. We exploit this sparsity to improve the computational cost of inference~\citep{paskin03a,thrun04,eustice05,walter07,kaess08}. In particular, we utilize the iSAM algorithm~\citep{kaess08} to update the canonical form by iteratively solving for the QR factorization of the information matrix. We refer the reader \citet{kaess08} for the details of this process.

\subsubsection{Re-weighting particles and resampling}

After modifying each particle's topology, we perform a Bayesian
update to its Gaussian distribution. We then re-weight each particle
according to the likelihood of generating language annotations and
region appearance observations:
\begin{equation}\label{eq:weight_update_all}
    w_t^{(i)}= p(z_t, \alpha_t \vert S_{t-1}^{(i)}) w_{t-1}^{(i)} = p(\alpha_t\vert S_{t-1}^{(i)}) p(z_t\vert S_{t-1}^{(i)}) w_{t-1}^{(i)}.
\end{equation}
When calculating the likelihood of each region appearance observation, we
consider the current node's region type and calculate the likelihood of
generating this observation given the topology. %
In effect, this down-weights any particle with a sampled
region of a particular type existing on top of a known traversed region of
a different type.  We use a likelihood model that describes the observation
of a region's type, with a latent binary variable $v$ that denotes whether or not
the observation is valid.
We marginalize over $v$ to arrive at the likelihood of generating the given
observation, where $R_u$ is the set of unobserved regions in particle
$S_{t-1}^{(i)}$:
\begin{equation}\label{eq:likelihood_appearance}
  p(z^t\vert{S_{t-1}^{(i)}}) = \prod_{R_i\in {R_u}}\left(\sum\limits_{v\in {1,0}}{p(z^t\vert{v,R_i}) \; p(v\vert{R_i})}\right).
\end{equation}
We define the probability of annotations $p(\alpha_t \vert S_{t-1}^{(i)})$ as the language grounding likelihood under the
map at the previous time step (Sec.~\ref{sec:nlu}). As such, a particle with an existing pair of
regions that conform to a specified language constraint will be weighted
higher than one without. In an effort to avoid particle depletion, we resample the particles when the variance of the weights exceeds a threshold, as measured by the number of effective particles~\citep{doucet98}
\begin{equation}
    N_\text{eff} = \frac{1}{\sum_{i=0}^n w_i^2}.
\end{equation}
In the experiments conducted in this paper, we set the threshold to half of the number of particles.

\subsection{Planning Under Uncertainty} \label{sec:planning}

It is intractable to search over the complete trajectory that is optimal in the distribution over maps. Instead, we formulate instruction-following in unknown environments as a planning under uncertainty problem, whereby we seek a policy $\pi$ that minimizes the one-step expected cost $c$.

In earlier work~\citep{duvallet13}, the cost was a function of a single topological representation of the environment, which included nodes that were previously visited as well as those that represent frontiers in the environment. In this work, however,  the planner must reason over a \emph{distribution} over semantic maps as opposed to a single world model. Thus, we express the cost $c$ as a function of the robot's current pose $x_t$, the available actions $a_t \in A_t$, and the map distribution $p(\Mapt)$
\begin{equation} \label{eqn:policy}
  \pi \parens{x_t, p(\Mapt)} = \argmin{a \in A_t} c\parens{x_t, a_t, p(\Mapt)},
\end{equation}
where the belief-space actions include paths from the robot's current pose to each node (landmark) in the graph $v_t \in G_t$, paired with the distribution over the nodes. The action space also includes an explicit stop action $\astop$ that declares that the planner has satisfied the instruction.
\begin{equation}
    A_t = \left\{\text{path}(x_t,v_t) \times p(v_t) \; \forall v_t \in G_t\right\}  \cup \{\astop\}
\end{equation}
In this manner, the planner differentiates between landmarks that the robot has detected and thus have low uncertainty, and those that are hypothesized from language and correspondingly more uncertain. Represented in this way, the robot can choose to explore unknown locations (e.g., continuing to search the hallway or navigating to a box that is hypothesized as containing the ball), backtrack when a previous action hasn't been fruitful (e.g., traveling to the wrong room or to an empty box), and stop when the planner concludes that the instruction has been satisfied.

The following sections explain how the policy reasons in belief space, and the novel imitation learning formulation to train the policy from demonstrations of correct behavior.  %

\subsubsection{Belief-Space Reasoning using Distribution Embedding}

A standard choice for the cost function is to represent it as a linear combination of features over the current state and actions~\citep{ratliff06,abbeel04,syed08a}, where the state nominally includes the robot's pose, the semantic map, and the current time. However, rather than a single world model, the semantic map distribution $p(\Mapt)$ provides a distribution over the observed and hypothesized location of landmarks (i.e., objects and regions) relevant to the given natural-language instruction.
As such, the policy $\pi$ must represent and compute distances between \emph{distributions} over action features when computing the cost of any action $a_t$.

Hilbert Space embeddings~\citep{smola2007,gretton07} provide a convenient representation of distributions with which one can efficiently measure the distance between distributions via a pseudometric. We embed the distribution over action features in a Reproducing Kernel Hilbert Space (RKHS), using the mean feature map~\citep{smola2007,gretton07} that consists of the first $K$ moments of the features computed with respect to each map sample $\Mapti$ and its likelihood as a sample-based representation of the distribution
\begin{subequations} \label{eqn:policy-decomposition}
    \begin{align}
        \Momenta \parens{x_t, a_t, \Mapt} &= \sum_{\Mapti} p(\Mapti) \: \phiXAMapi \\
        \Momentb \parens{x_t, a_t, \Mapt} &= \sum_{\Mapti} p(\Mapti) \: \parens{\phiXAMapi - \Momenta }^2\\
        &\;\;\vdots \nonumber\\
        \MomentK \parens{x_t, a_t, \Mapt} & = \sum_{\Mapti} p(\Mapti) \: \parens{\phiXAMapi - \Momenta }^K.
    \end{align}
\end{subequations}

\usetikzlibrary{positioning}

\newcommand{\robotX}{-3.5}
\newcommand{\robotY}{+.2}
\newcommand{\wallLengthBeyondRobot}{2.0}
\newcommand{\lowerWall}{-0.7}

\newcommand{\boxAX}{-8.0}
\newcommand{\boxAY}{1.5}
\newcommand{\boxBX}{-8.75}
\newcommand{\boxBY}{0.0}

\tikzstyle{invisibleLandmark}=[
  draw=black, fill=black!05, thick, dashed]

\tikzstyle{boxStyle}=[
  rectangle,
  draw,
  dashed,
  fill=black!15,
  minimum width=4em,
  minimum height=1.5em,
  label={center:Box}
]

\newcommand{\drawBeliefSpace}{

  \node[] (startPosition) at (0, 0.2) {};
  \node[] (robotPosition) at (\robotX, \robotY) {};
  \node[] (actionStraight) at (\robotX-\wallLengthBeyondRobot, .2) {};
  \node[] (midBox) at (\robotX, .2) {};
  \node[] (actionRight) at (\robotX, 2) {};

  \node[] (centroid) at (-8.5,0.9){};
  \draw[rotate around={-55:(centroid.center)},
    fill=black!05,
    dashed,
  ] (centroid.center) ellipse (1.7cm and 2cm);

  \draw[visibleWallStyle]  %
  (\robotX+1, 2) -- (\robotX+1, 1)  -- (1, 1) --  (1, \lowerWall) --
  (\robotX-\wallLengthBeyondRobot, \lowerWall);

  \draw[visibleWallStyle]  %
  (\robotX-1, 2) -- (\robotX-1, 1)  -- (\robotX-\wallLengthBeyondRobot, 1) ;

  \def\shift{0.5mm}  %
  \draw[policyActionStyle, transform canvas={yshift=\shift}, shorten >=\shift]
  (startPosition.west) -- (midBox.center) -- (actionRight.center);

  \def\shift{-0.5mm}  %
  \draw[policyActionStyle, transform canvas={yshift=\shift}, shorten >=\shift]
  (startPosition.west) -- (actionStraight.center);

  \node[frontierNode, label=right:$a_1$] (frontierBox) at (actionRight) {};
  \node[frontierNode, label=below:$a_2$] (frontierStair) at (actionStraight) {};

  \node (boxA) at (\boxAX,\boxAY) [boxStyle] {};
  \node (boxB) at (\boxBX,\boxBY) [boxStyle] {};

  \drawRobot{robotPosition}
  \drawStart
}

\tikzstyle{features1Style}=[
  thick,
  dotted,
  rounded corners,
  color=blue,
]

\tikzstyle{features2Style}=[
  thick,
  dotted,
  rounded corners,
  color=red,
]

\newcommand{\drawFeatureArrows}{
  \draw[features1Style] (actionRight) -- (boxA.east);
  \draw[features1Style] (actionRight) -- (boxB.east);
  \node[features1Style] at (\robotX-1.5, 2.75) {$\phi(a_1, S^1), \phi(a_1, S^2)$};

  \draw[features2Style] (actionStraight) -- (boxA.east);
  \draw[features2Style] (actionStraight) -- (boxB.east);
  \node[features2Style] at (\robotX-3, -1.2) {$\phi(a_2, S^1), \phi(a_2, S^2)$};
}

\tikzstyle{visibleLandmark}=[
  draw=black, fill=black!20, thick,
  font = \footnotesize,
]

\tikzstyle{landmarkText} = [
  text centered,
  font = \footnotesize]

\newcommand{\nodeSize}{7}
\tikzstyle{visibleNode}=[
  circle,
  draw=black!80,
  fill=blue!10,
  thick,
  inner sep = 0,
  minimum size = \nodeSize]
\tikzstyle{frontierNode}=[
  circle, double,
  draw=black!80,
  fill=orange!40,
  thick,
  inner sep = 0,
  minimum size = \nodeSize]
\tikzstyle{unknownNode}=[
  circle,
  draw=blue!50,
  fill=blue!20,
  thick, dashed,
  inner sep = 0,
  minimum size = \nodeSize]

\tikzstyle{edgeStyle} = [
  very thick,
  dashed
]
\tikzstyle{pathStyle} = [
  ultra thick,
  rounded corners,
  cap = round,
  color = red!90,
  join = round,
]

\tikzset{dashdot/.style={dash pattern=on .4pt off 4pt on 4pt off 4pt}}

\tikzstyle{policyActionStyle} = [
  ultra thick,
  rounded corners,
  cap = round,
  color = gray!80,
  join = round,
  dashed,
]

\tikzstyle{policyChosenActionStyle} = [
  ultra thick,
  rounded corners,
  cap = round,
  color = blue!90,
  join = round,
  dashdot,
]

\tikzstyle{legendStyle} = [
  matrix,fill=blue!10,draw=blue,very thick
]

\tikzstyle{visibleWallStyle} = [
  very thick,
  rounded corners,
  join = round
]

\newcommand{\robotSize}{.42}
\newcommand{\drawRobot}[1] {
  \fill (#1) circle (\robotSize);
  \node[] (robotInternal) at (#1) {};
}

\newcommand{\labelRobot}{
  \node[font=\footnotesize, below = .05 of robotInternal] {Robot};
}

\newcommand{\drawStart}{
  \node[] at (0, -0.1) {Start};
}

\newcommand{\drawGrid}{
  \draw[help lines,xstep=1,ystep=1] (-9,-2) grid (0,5);
  \foreach \x in {-9,...,0} { \node [anchor=south] at (\x,5) {\x}; }
  \foreach \y in {-2,...,5} { \node [anchor=west] at (0,\y) {\y}; }

}

\newcommand{\drawLandmarksLegend}{
  \node [legendStyle] (my matrix) at (-7,-2.5){
    \node[] {\textbf{Legend}}; \pgfmatrixnextcell  \\
    \node[left] {Known Object:}; \pgfmatrixnextcell
    \node[visibleLandmark, rectangle] {Landmark}; \\
    \node[left] {Unknown Object:}; \pgfmatrixnextcell
    \node[invisibleLandmark, rectangle] {Landmark}; \\
  };
}

\newcommand{\drawGraphLegend}{
  \node [legendStyle] (my matrix) at (-7,-2.5){
    \node[] {\textbf{Legend}}; \pgfmatrixnextcell  \\
    \node[left] {Visited Vertex:}; \pgfmatrixnextcell
    \node[visibleNode] {}; \\
    \node[left] {Frontier Vertex:}; \pgfmatrixnextcell
    \node[frontierNode] {}; \\
  };
}

\newcommand{\tikzpicturescale}{.6}
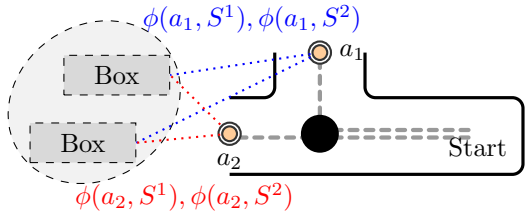
\begin{figure}[t]
  \centering
  \begin{tikzpicture}[scale=\tikzpicturescale]
    \drawBeliefSpace
    \drawFeatureArrows
    \end{tikzpicture}
  \caption{A simplified illustration of the computation of feature moments in the space of hypothesized landmarks. In this example, there are two hypothesized boxes. In order to determine the features over a landmark distribution, we compute the features for each action across all hypothesized landmark samples and aggregate by means of moment statistics.}
  \label{fig:feature-computation}
\end{figure}

The vector $\phi (x_t, a_t, \Mapti)$ is a concatenation of features that are a function of action $a_t$ and a \emph{single} landmark in $\Mapti$. These include geometric features that express the shape of the path associated with the action, such as the cumulative change in angle, which may be correlated with actions that go straight or that turn. They also include features that express the geometry of the landmark, such as the area of the landmark region, and relationships between the action and landmark, such as the difference between the distance from the landmark at the start and the distance at the end of the path, which may be correlated with going towards vs.\ away from the landmark. Please see our earlier work~\citet{duvallet13} for a thorough description of these features.

This RKHS-based representation of the features (Eqn.~\ref{eqn:policy-decomposition}) computes features individually for the action and all hypothesized landmarks in the sample-based representation of the map distribution. We aggregate these feature vectors, and then compute moments of the feature vector distribution (i.e., the mean, variance, and higher order statistics). Figure~\ref{fig:feature-computation} provides a simple illustration of how we compute belief-space features for two actions with a hypothesized box that has two possible locations.

We express the cost function in Equation~\ref{eqn:policy} as a weighted sum of the first $K$ moments of the feature distribution (Eqn.~\ref{eqn:policy-decomposition}):
\begin{equation} \label{eqn:belief-policy-sum}
  \costMapt = \sum_{k=1}^K \wktranspose \; \Momentk \XAMapt.
\end{equation}
where $w_k$ is the weight vector associated with moment $k$.
Concatenating the weights and moments into column vectors $W = \begin{bmatrix} w_1 & \cdots & w_K\end{bmatrix}^\top$ and $F = \begin{bmatrix} \Momenta & \cdots & \MomentK \end{bmatrix}^\top$, we can express the policy in Equation~\ref{eqn:policy} as minimizing a weighted sum of the feature moments $F_{a_t}$ for action $a_t$ :
\begin{equation} \label{eqn:belief-policy-vector}
  \pi \parens{x_t, \Mapt} = \argmin{a_t \in A_t} W^\top F_{a_t}.
\end{equation}

\subsubsection{Imitation Learning Formulation}

We train the policy using imitation learning whereby we treat the problem of predicting the action as a multi-class classification problem. Specifically, given an expert demonstration, we seek a policy that correctly predicts their chosen action among all possible actions for the same state.
Our earlier work proposed the use of imitation learning for training a direction-following policy, however it assumes that the environment is at least partially known a priori~\citep{duvallet13}. We make no such assumption here, and instead train a belief-space policy that reasons in a \emph{distribution} of hypothesized maps, thereby supporting instances in which the agent has no a priori knowledge of the environment.

We assume that the expert's policy $\pistar$ minimizes the unknown immediate
cost $c(x_t, \astar, \Mapt)$ of performing the demonstrated action $\astar$ from
state $x_t$, under the map distribution~$p(\Mapt)$.
However, since we cannot directly observe the true costs of the expert's policy, we must instead minimize a surrogate loss that penalizes disagreements between the expert's action~$\astar$ and the action $a_t$ selected by the policy using the multi-class hinge loss~\citep{crammer02}, computed over all demonstrations:
\begin{equation} \label{eqn:svm-loss}
    \ell \parens{x_t, \astar , c, \Mapt}  =   \max  \parens{  0, 1  + \costMaptAstar  -  \min_{a_t \ne \astar} \brackets{\costMapt} }.
\end{equation}
Formulated in this manner, the policy selects an action that differs from that of the expert (i.e., $a_t \neq \astar$) if and only if the cost associated with that action ($c(x_t, a_t, \Mapt)$) is less than the cost of the expert's action ($\costMaptAstar$) by a margin of one. The loss can be re-written and combined with Equation~\ref{eqn:belief-policy-vector} to yield:
\begin{equation} \label{eqn:loss-augmentation}
  \ell \parens{x_t, \astar, W, \Mapt} = \Wtranspose F_\astar - \min_{a_t} \brackets{ \Wtranspose F_{a_t} - l_{xa} },
\end{equation}
where the margin $l_{xa} = 0$ if $a_t=\astar$ and $l_{xa} = 1$ otherwise.
This ensures that the expert's action is better than all other actions by a margin~\citep{ratliff06}. We further add a regularization term $\lambda$ to Equation~\ref{eqn:loss-augmentation}, yielding the complete optimization loss:
\begin{equation} \label{eqn:optimization-loss}
\ell \parens{x_t, \astar, W, \Mapt}  =  \frac{\lambda}{2} \Vert W \Vert^2 + \Wtranspose F_\astar - \min_{a_t} \left[ \Wtranspose F_{a_t} - l_{xa} \right].
\end{equation}

Although this loss function is convex, it is not differentiable.
However, we can optimize it efficiently by taking the subgradient of Equation~\ref{eqn:optimization-loss} and computing action predictions for the loss-augmented policy~\citep{ratliff06}:
\begin{equation}
    \dldwfrac = \lambda W + F_\astar - F_{a_t^\prime} \\
\end{equation}
where $a_t^\prime$ is the best loss-augmented action associated with state $x_t$ (i.e., the solution to our policy using the loss-augmented cost):
\begin{equation}
    a_t^\prime = \argmin{a_t} \left[ \Wtranspose F_{a_t} - l_{xa} \right].
\end{equation}
The subgradient leads to the update rule for the weights $W_t$:
\begin{equation} \label{eqn:update-rule}
  W_{t+1} \gets W_t - \alpha \; \dldwfrac
\end{equation}
with a learning rate $\alpha \propto 1/t^\gamma$.
Intuitively, if the current policy disagrees with the expert's demonstration, Equation~\ref{eqn:update-rule} decreases the weight (and thus the cost) for the features of the demonstrated action $F_\astar$, and increases the weight for the features of the planned action $F_{a_t^\prime}$.
If the policy produces actions that are consistent with the expert's demonstration, only the regularization term is updated.
As in our prior work, we train the policy using the \DAgger algorithm~\citep{ross11}, which learns a policy by iterating between collecting data (using the current policy) and applying expert corrections on all states visited by the policy (using the expert's demonstrated policy).

Treating direction following in the space of possible semantic maps as a problem of sequential decision making under uncertainty provides an efficient approximate solution to the belief-space planning problem.
By using a kernel embedding of the distribution of features for a given action, our approach can learn a policy that reasons about the distribution of semantic maps.

\section{Experiments}\label{sec:experiments}
\begin{figure*}[!th]
  \centering
  \subfigure[Clearpath Robotics Husky A200 UGV]{\includegraphics[height=0.35\linewidth]{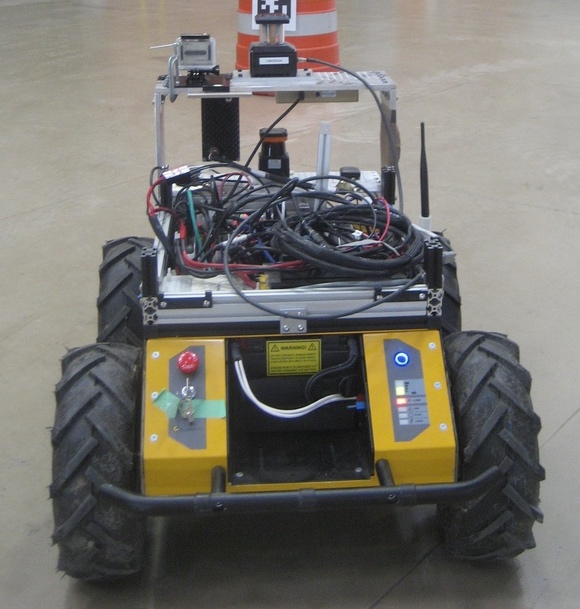}\label{fig:husky-nrec}}\hfil%
  \subfigure[Robotic Wheelchair]{\includegraphics[height=0.35\linewidth]{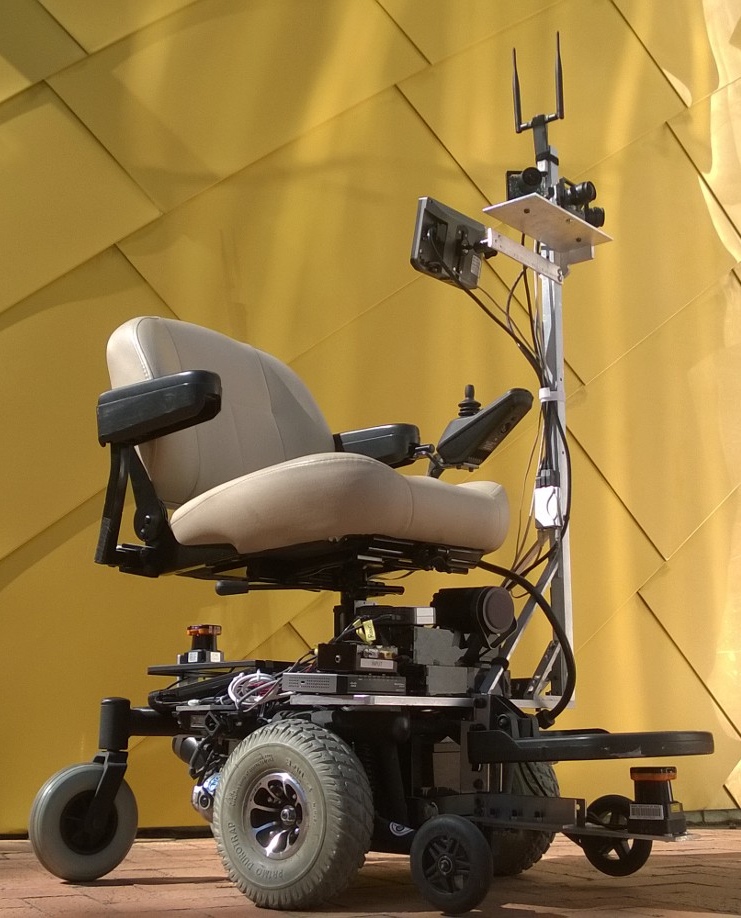}\label{fig:wheelchair-stata-outdoors}}\hfil%
  \subfigure[Clearpath Robotics Husky A200 UGV with a Universal Robotics UR5 Manipulator]{\includegraphics[height=0.35\linewidth]{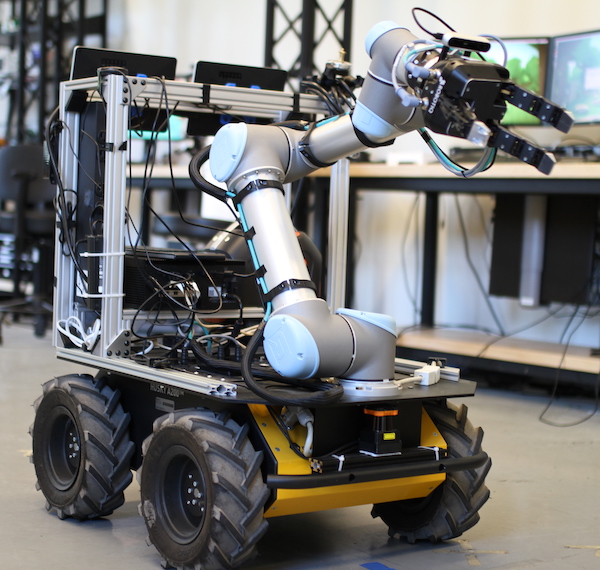}\label{fig:husky-rochester}}
  \caption{The three robots used in physical experiments.  The Clearpath Robotics Husky A200 UGV in (a) and the Robotic Wheelchair in (b) were used for experiments involving following of route instructions in Section \ref{sec:results-following-natural-language-route-instructions} while the Clearpath Robotics Husky A200 UGV outfitted with a Universal Robotics UR5 manipulator in (c) was used for natural language understanding for mobile manipulation in Section \ref{sec:results-natural-language-understanding-for-mobile-manipulation}}
  \label{fig:robots}
\end{figure*}

To evaluate the effectiveness of the proposed approach to natural-language instruction-following in a priori unknown environments, we performed experiments on the three different robotic systems illustrated in Figure \ref{fig:robots}.  For each of these, we used a form of a symbolic representation for annotation inference and behavior inference that involved several different types of symbols.

The first system evaluated was an unmanned ground vehicle.  For annotation and behavior inference for these experiments we assumed seven different spatial relation types $\mathcal{S}$ (``unknown'', ``near'', ``away'', ``front'', ``back'', ``left'', ``right''),  eight object types $\mathcal{O_{C}}$ ('' unknown'', ``robot'', ``cone'', ``tree'', ``car'', ``building'', ``hydrant'', and ``wall'') and no location types.  For the space of behaviors, we considered two possible action types $\mathcal{A_{O}}$  (``unknown'' and ``navigation'') and three possible modes $\mathcal{M}$ (``safely'', ``quickly'', and ``unknown'').  For symbol grounding, both annotation and behavior inference used the DCG model~\cite{howard14} with a corpus of 39 fully labeled examples.  Additional information on the symbolic representation and corpus used in the experiments can be found in \citet{duvallet14}.

The second system evaluated was a robotic wheelchair. This system was evaluated in an office building environment.  For both annotation and behavior inference, we assumed 12 different spatial relation types $\mathcal{S}$ (``unknown'', ``near'', ``away'', ``front'', ``back'', ``left'', ``right'', ``down'', ``through'', ``towards'', ``past'', and ``around'') and 17 object and location types. Specifically, we used eight object types $\mathcal{O_{C}}$ (``unkown'', ``robot'', ``cone'', ``tree'', ``car'', ``building'', ``hydrant'', and ``wall'') and nine location types  $\mathcal{L}$ (``generic'', ``kitchen'', ``office'', ``hallway'', ``lab'',``lounge'', ``elevator'', ``conference room'', and ``cafeteria'').  These experiments also included twelve spatial relation types.  For the space of behaviors, we considered four possible actions (``unknown'', ``navigation'', ``right'', and ``left'') and three possible modes $\mathcal{M}$ (``safely'', ``quickly'', and ``unknown'').  For annotation inference and behavior inference, we used the HDCG~\cite{chung15} model with a corpus of 54 fully labeled examples.  More information about the symbolic representation used in these experiments can be found in \citet{hemachandra15}.

The third system that we evaluated was a mobile manipulator consisting of a Clearpath Robotics Husky A200 Unmanned Ground Vehicle outfitted with a Universal Robotic UR5 manipulator and Robotiq Adaptive Robotic Gripper. We conducted experiments in both indoor and outdoor environments with a series of instructions that required the robot to both navigate and manipulate objects.  A subset of these objects were selected from the YCB dataset~\cite{calli2015benchmarking}.  For annotation inference, the object types $\mathcal{O_{C}}$ included ``drill'', ``suitcase'', ``banana'', ``pitcher'', ``cracker box'', ``mustard bottle'', ``ball'', ``box'', and ``cone'', the location types $\mathcal{L}$ included ``lab'', ``hallway'', and ``office'', and the spatial relation types $\mathcal{S}$  included ``inside'', ``behind'', ``left'', ``right'', ``front'', and ``back''.  For behavior inference, the action types $\mathcal{A_{O}}$ included ``pick'', ``retrieve'', and ``navigate'', the object types $\mathcal{O_{C}}$ included ``drill'', ``suitcase'', ``banana'', ``pitcher'', ``cracker box'', ``mustard bottle'', ``ball'', ``box'', and ``cone'', and the spatial relation types $\mathcal{S}$  included ``inside'' and ``behind''.  No model types were used for the mobile manipulation experiments.  For symbol grounding, both annotation and behavior inference used the DCG model~\cite{howard14} with a corpus of 115 fully labeled examples.  The corpus contained instructions such as ``Pick up the drill behind the cone'', ``Pick up the pitcher'', ``Go to the mustard bottle'', ``Retrieve the crackers box inside the box'' etc. Additional information on the symbolic representation and corpus used in the experiments can be found in \citet{patki2020a}.

\section{Experimental Results} \label{sec:experimentaldesign}

In the following, we discuss the results of the various experiments intended to evaluate the performance of our framework. We first consider the experiments focused on route instruction-following with the different robot platforms. We then analyze the results of the experiments that task a robot with carrying out natural language commands that involve mobile manipulation.

\subsection{Instruction Following for Robot Navigation}

We evaluated our algorithm's ability to follow route instructions in a priori unknown environments through experiments conducted in simulation as well as those involving the Husky and wheelchair robots. For comparison, we include a ``Known Map'' ground-truth baseline that performs language understanding with the environment being completely known. In this manner, the baseline provides an upper-bound on the performance of our framework.

\subsubsection{Monte Carlo Simulations}

\begin{figure}[!t]
  \centering
  \subfigure[$t=0$]{\includegraphics[width=0.4\linewidth]{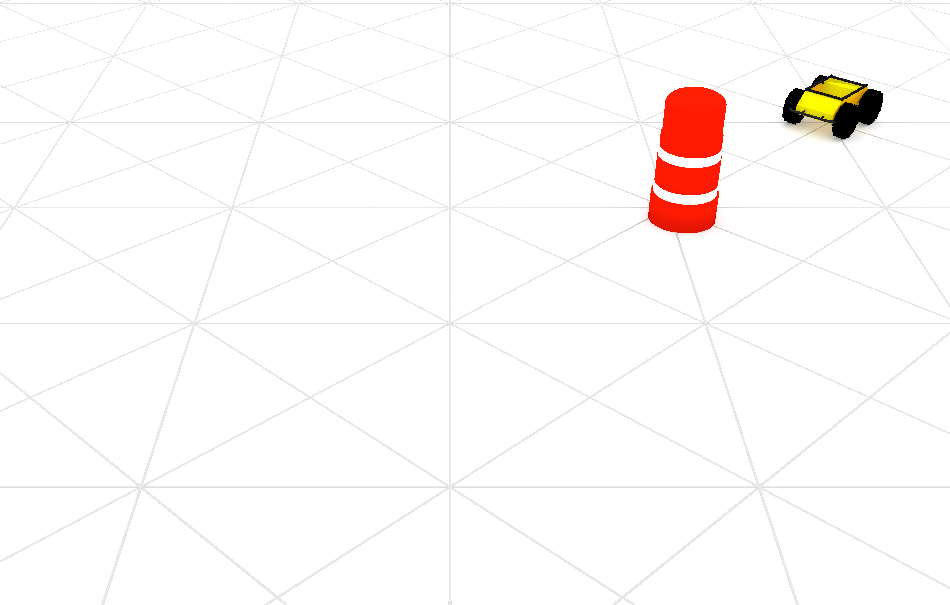}\label{fig:husky-evolution-simulation-a}}\hfil%
  \subfigure[$t=1$]{\includegraphics[width=0.4\linewidth]{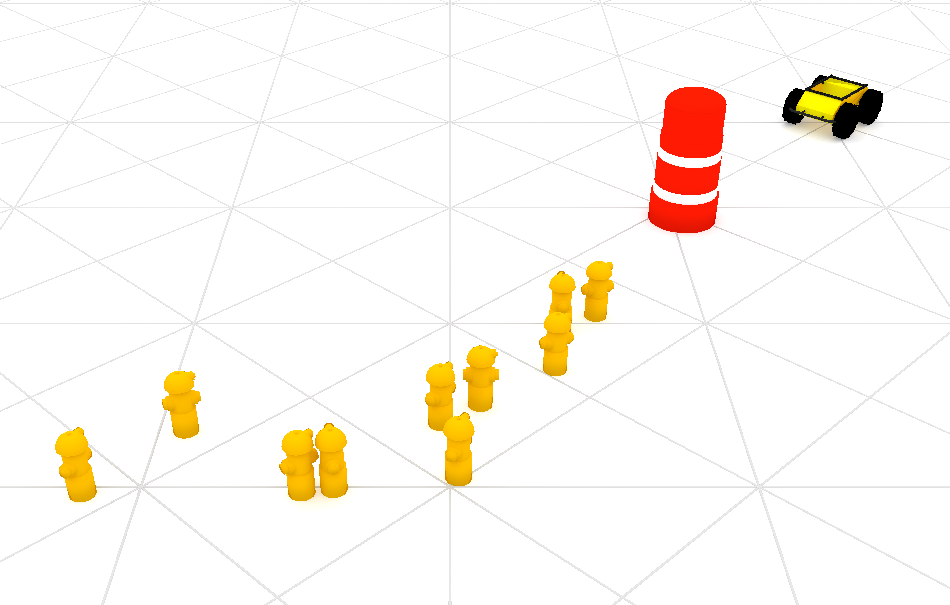}\label{fig:husky-evolution-simulation-b}}\\
  \subfigure[$t=2$]{\includegraphics[width=0.4\linewidth]{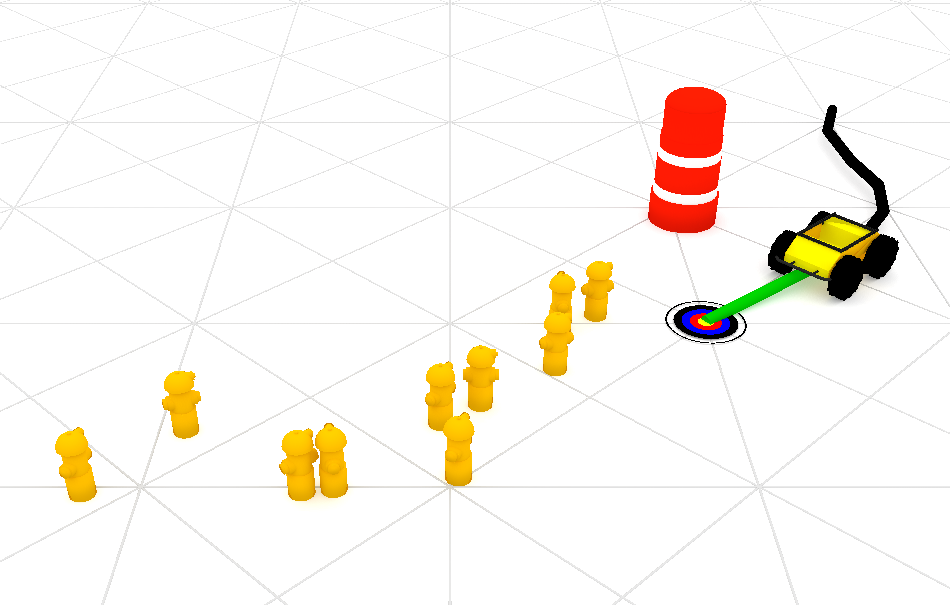}\label{fig:husky-evolution-simulation-c}}\hfil%
  \subfigure[$t=3$]{\includegraphics[width=0.4\linewidth]{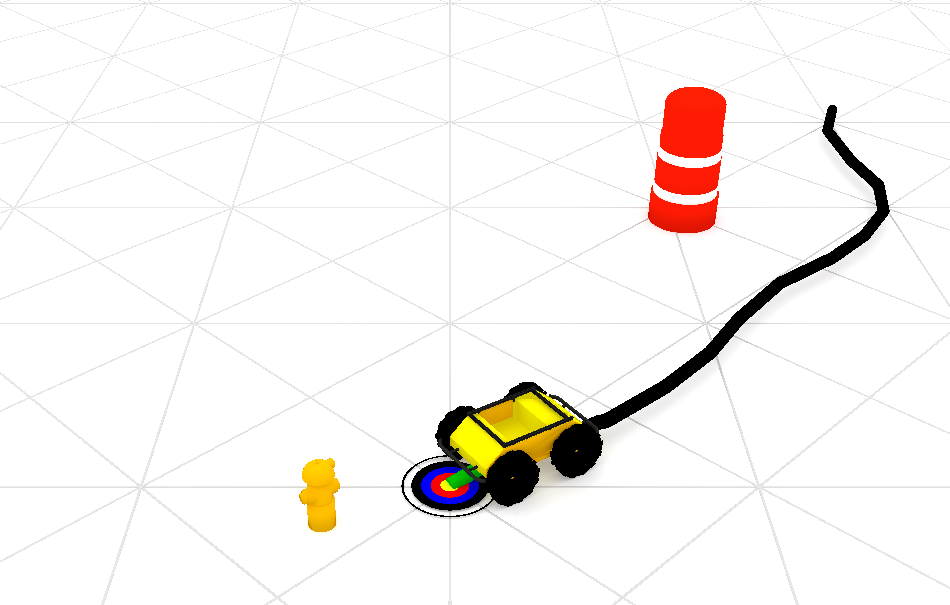}\label{fig:husky-evolution-simulation-d}}
  \caption{The evolution of our framework for a simulation-based experiment that tasks the Husky with following the instruction ``go to the hydrant behind the cone''. When the robot receives the command, \subref{fig:husky-evolution-simulation-a} the traffic cone is visible, but not the hydrant, which is occluded by the cone. Our algorithm \subref{fig:husky-evolution-simulation-b} proposes a distribution over the environment that effectively hypothesizes the location of the hydrant, which we visualize as the mean location for each of the particles. Based on this distribution, \subref{fig:husky-evolution-simulation-c} the imitation learning-based policy decides to navigate behind the cone. At this point, \subref{fig:husky-evolution-simulation-d} the robot observes the hydrant upon which the environment distribution effectively converges to the location of the detected hydrant (in practice, the distribution will still assign small, but non-zero likelihood for the hydrant being located elsewhere), which the policy then identifies as the goal.}
\end{figure}
\paragraph{Object-relative navigation} We begin with a series of Monte Carlo simulation-based evaluations of our framework for the task of following natural language route directions. The first set of experiments considers a simple setup in which a Husky A200 Unmanned Ground Vehicle navigates an open environment consisting of different combinations of objects. We consider four environment
templates, with different numbers of hydrants and cones. For each configuration, we sample ten environments, each with
different object poses. For these environments, we issued three natural
language instructions ``go to the hydrant,'' ``go to the hydrant behind the
cone,'' and ``go to the hydrant nearest to the cone.'' We note that these
commands were not part of the corpus that we used to train the DCG
model. Additionally, we considered six different settings for the robot's
field-of-view, 2\,m, 3\,m, 5\,m, 10\,m, 15\,m, and 20\,m, and performed
approximately 100 simulations for each combination of the environment, command,
and field-of-view. As a ground-truth baseline, we performed ten runs of
each configuration with a completely known world model.
\begin{table}[!h]
    \newcolumntype{S}{c@{\hskip 0.15in}}
    \newcolumntype{B}{c@{\hskip 0.17in}}
    \newcommand{\hp}[1]{\leavevmode\hphantom{#1}}
    \centering
    \caption{Monte Carlo simulation results with $1\sigma$ confidence intervals for the Husky experiments.}
    \begin{tabularx}{0.95\textwidth}{llSSSSBBB}
        \toprule
        &&&& \multicolumn{2}{c}{Success Rate (\%)} && \multicolumn{2}{c}{Distance
          (m)}\\
        \cmidrule{5-6} \cmidrule{8-9}
        \multicolumn{2}{c}{Environment} & FOV (m) & Relation & Known & Ours && Known & Ours\\
        \midrule
        1 hydrant & 1 cone & 3.0 & null        & 100.0 & \hp{0}93.9 && \hp{0}8.75 (1.69) & 16.78 \hp{0}(7.90)\\
        1 hydrant & 1 cone & 3.0 & ``behind''  & 100.0 & \hp{0}98.3 && \hp{0}8.75 (1.69) & 13.43 \hp{0}(7.02)\\
        1 hydrant & 2 cones & 3.0 & null        & 100.0 & 100.0      &&      11.18 (1.38) & 32.54 (18.50)\\
        1 hydrant & 2 cones & 3.0 & ``behind''  & 100.0 & \hp{0}99.5 &&      11.18 (1.38) & 40.02 (29.66)\\
        2 hydrants & 1 cone & 3.0 & null        & 100.0 & \hp{0}54.4 &&      10.49 (1.81) & 21.56 (10.32)\\
        2 hydrants & 1 cone & 3.0 & ``behind''  & 100.0 & \hp{0}67.4 &&      10.38 (1.86) & 18.72 (10.23)\\
        2 hydrants & 1 cone & 5.0 & ``nearest'' & 100.0 & \hp{0}46.2 && \hp{0}9.19 (1.54) & 12.05 \hp{0}(5.76)\\
        \bottomrule
    \end{tabularx}
    \label{tab:husky-results-sim}
\end{table}
\begin{figure}[!t]
    \centering
	\subfigure[Distance traveled for ``behind'' relation]{%
    \includegraphics[width=0.49\linewidth]{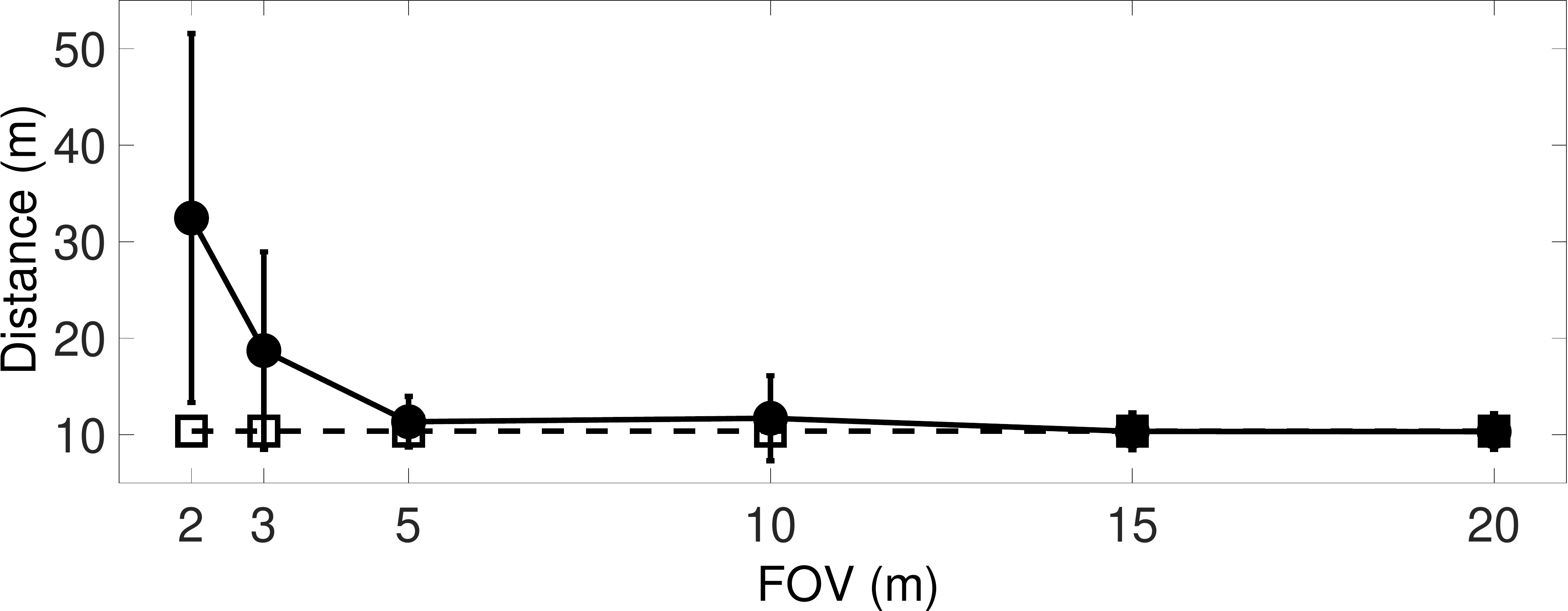}}%
    \hfil%
    \subfigure[Distance traveled for ``nearest'' relation]{%
	\includegraphics[width=0.49\linewidth]{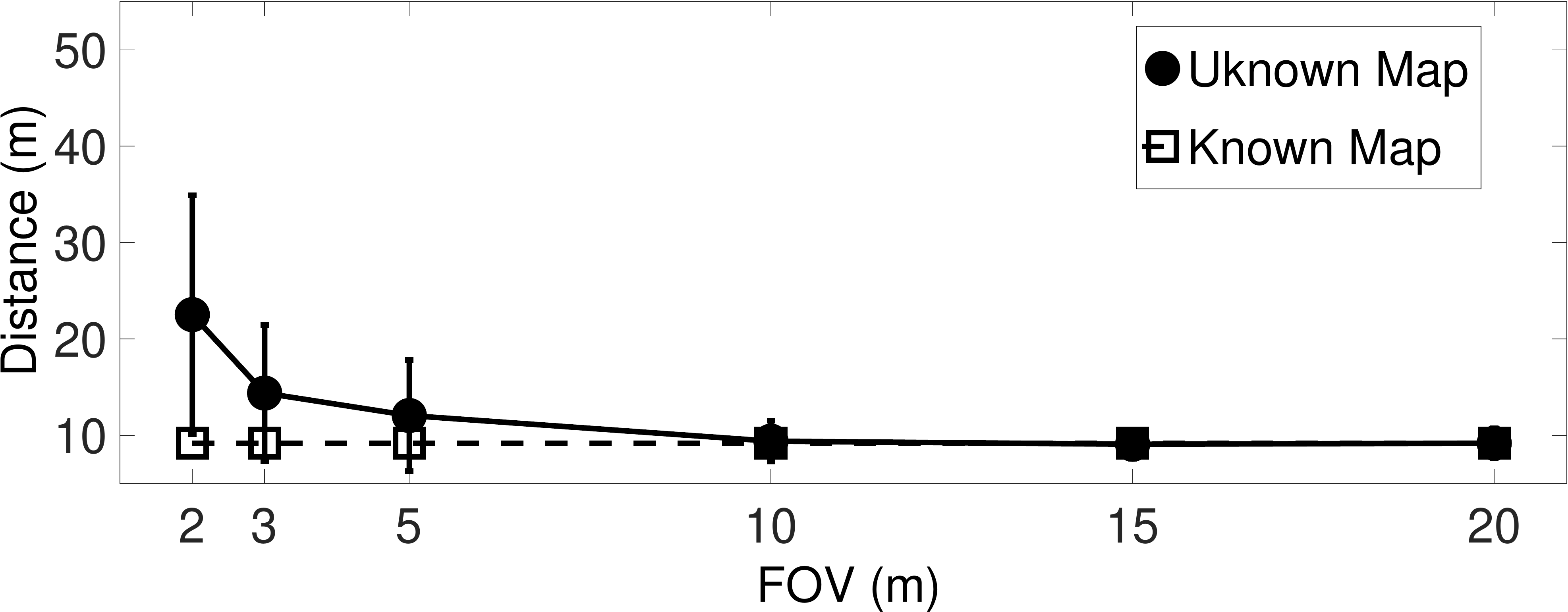}}\\
    \subfigure[Success rate for ``behind'' relation]{%
	\includegraphics[width=0.49\linewidth]{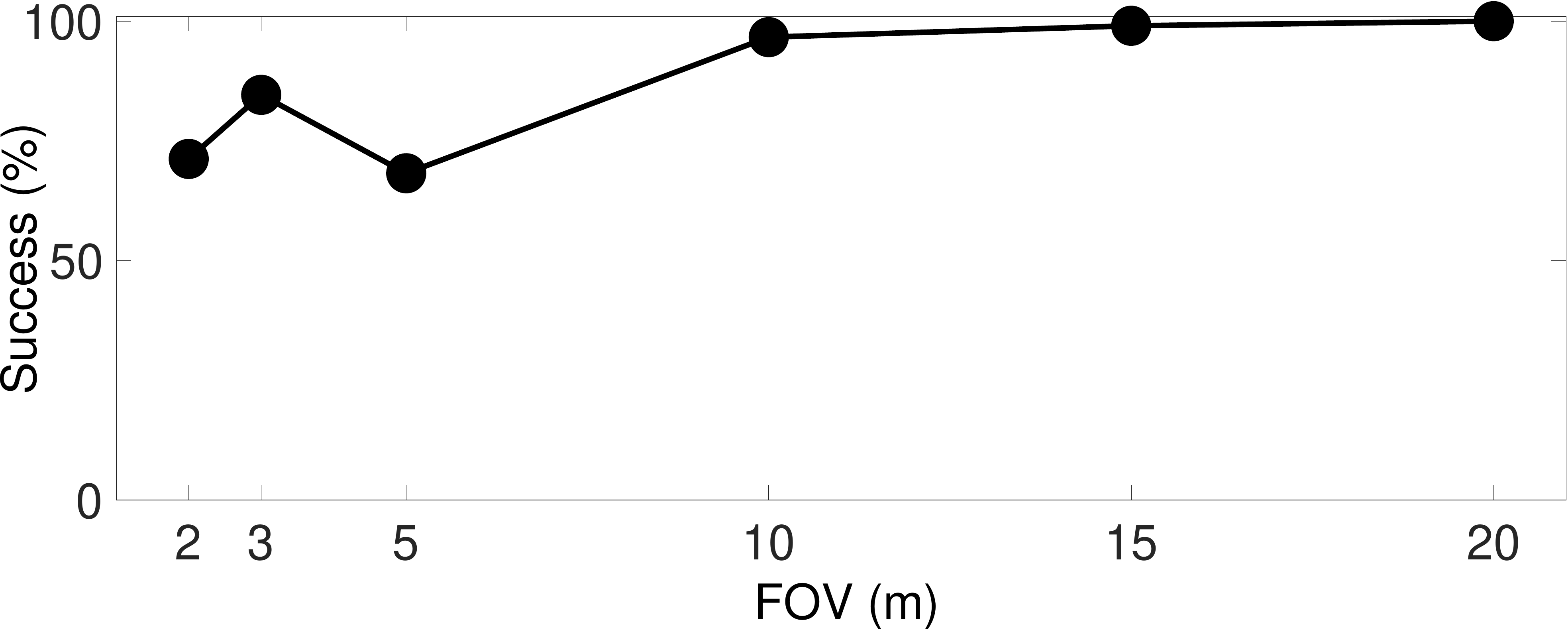}}%
    \hfil%
    \subfigure[Success rate for ``nearest'' relation]{%
	\includegraphics[width=0.49\linewidth]{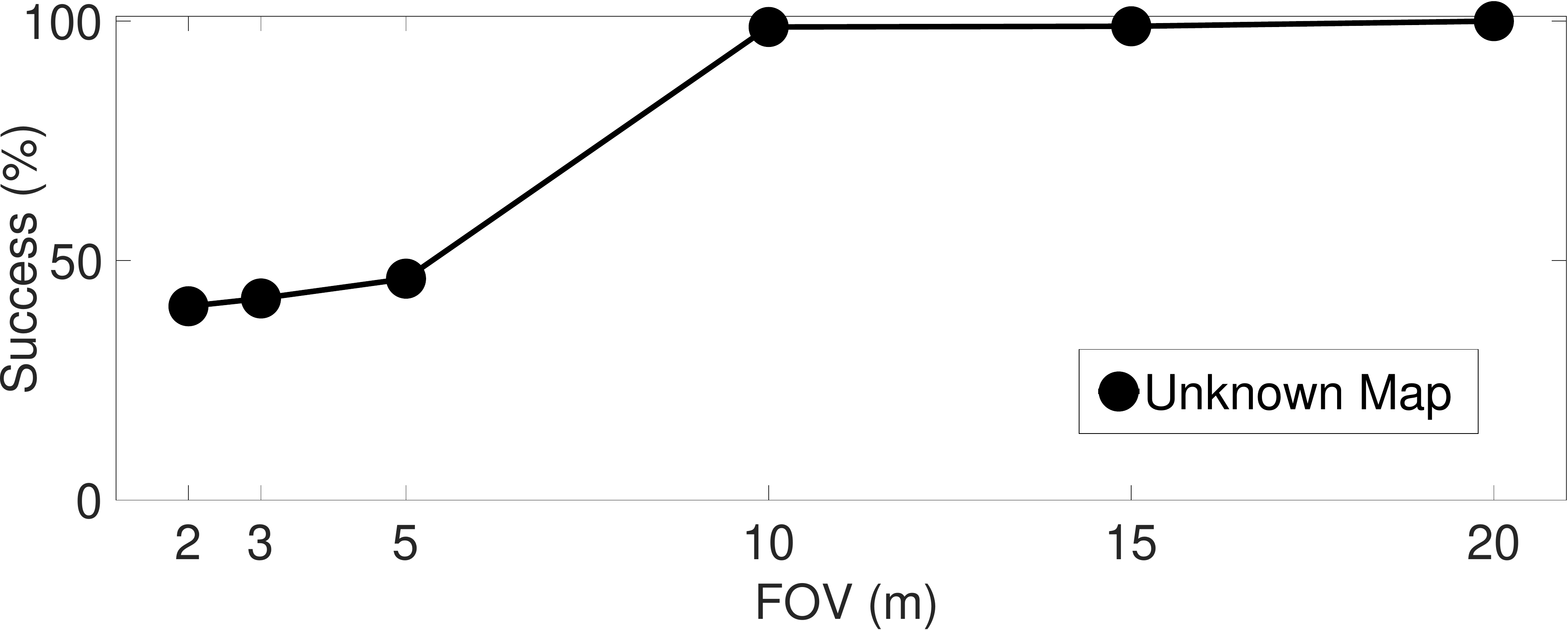}}
    \caption{Plots of the (top) distance traveled and (bottom) success rate as a function of the field-of-view for the commands (left) ``go
      to the hydrant behind the cone'' and (right) ``go to the hydrant
      nearest to the cone'' in simulation.} \label{fig:sensingRange}
\end{figure}
Table~\ref{tab:husky-results-sim} presents the success rate and distance
traveled by the robot for these 100 simulation configurations. We
considered a run to be successful if the planner stops within 1.5\,m of the
intended goal. Comparing against commands such as ``go to the hydrant'' that do not provide an explicit spatial relation, the results demonstrate that our algorithm achieves greater success and yields more efficient paths by taking
advantage of relations in the command (i.e., ``go to the hydrant behind the
cone''). This is apparent in environments that consist of a single hydrant as well as more ambiguous environments that consist of two
hydrants. Particularly telling is the variation in performance as a result
of different fields-of-view. Figure~\ref{fig:sensingRange} shows how the
success rate increases and the distance traveled decreases as the robot's
sensing range increases, quickly approaching the performance of the system
when it begins with a completely known map of the environment.

One interesting failure case is when the robot is instructed to ``go to the
hydrant nearest to the cone'' in an environment with two hydrants. In
instances where the robot sees a hydrant first, it hypothesizes the
location of the cone, and then identifies the observed hydrants and
hypothesized cones as being consistent with the command. Since the robot
never actually confirms the existence of the cone in the real world, this
results in the incorrect hydrant being labeled as the goal.

\newcommand{\includeValueGraphic}[2][]{
  \includegraphics[
    angle=90,
    trim = 20mm 10mm 0mm 10mm, clip,
    width = 0.315\textwidth,
    #1
  ]{#2}
}

\begin{figure*}[!t]
  \centering
  \subfigure[$t=0$\,sec]{%
    \includeValueGraphic{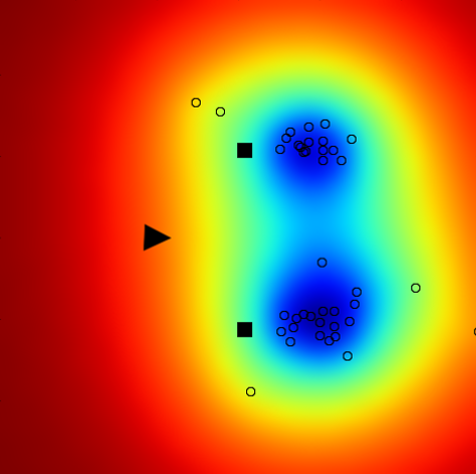}%
  \label{fig:value1}}\hfill%
  \subfigure[$t=4$\,sec]{%
    \includeValueGraphic{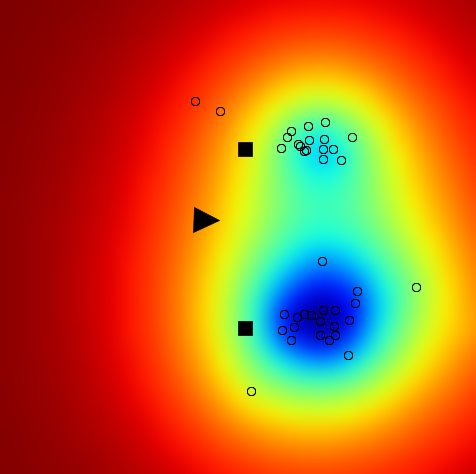}%
  \label{fig:value2}}\hfill%
  \subfigure[$t=8$\,sec]{%
    \includeValueGraphic{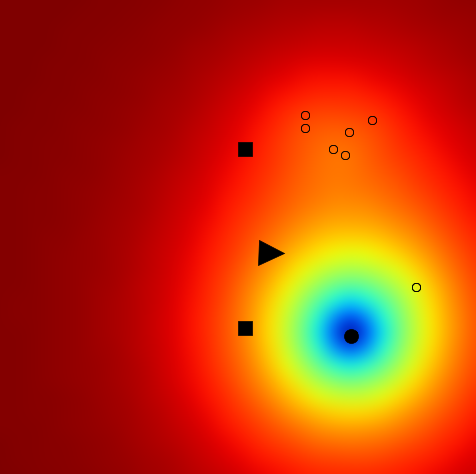}%
  \label{fig:value3}}
  \caption{A visualization of the evolution of the semantic map distribution over time for the command ``go to the hydrant behind the cone,'' where the triangle denotes the robot's location, squares represent observed cones, and circles denote hydrants that are either hypothesized (open) or observed (filled). \subref{fig:value1} The robot starts off observing both cones, and hypothesizes possible hydrants that are consistent with the command. \subref{fig:value2} The robot moves towards the left cluster, but having not observed the hydrant, the map distribution shifts the mass to the right. \subref{fig:value3} The robot then moves right and observes the actual hydrant.}
  \label{fig:valueFunction}
\end{figure*}

\paragraph{Multi-room navigation} The second set of Monte Carlo experiments goes beyond object-based navigation and considers natural language direction-following in a larger environment modeled after a multi-room office building (the MIT Stata Center) that consists of a laboratory, multiple hallways, and a kitchen. We compare our framework against two baselines. The first baseline (``Known Map'') considers the ideal case in which the world model is known a priori, employing our HDCG statistical language grounding model to infer the actions consistent with the route instruction.  The second assumes no prior knowledge of the environment
(as with our method), but
does not use language to modify the map. Instead, the baseline uses our HDCG framework to opportunistically ground the command in the current semantic map generated from observations $z^t$ (but not language). Note that, as with our framework, this baseline performs the grounding anew as the map evolves. We refer to this method as the ``Without Language'' baseline. Note that this and the ``Known Map'' baselines use an anytime RRT$^*$ planner~\citep{karaman11} to plan paths to the goal identified by language grounding.

\begin{table}[!ht]
    \newcolumntype{S}{c@{\hskip 0.22in}}
    \centering
    \caption{Simulation-based evaluation of natural language route direction-following.}
    \begin{tabularx}{0.825\linewidth}{Scccc}
        \toprule
        & \multicolumn{2}{c}{Distance (m)} & \multicolumn{2}{c}{Time (sec)}\\
        \cmidrule{2-3} \cmidrule{4-5}
        Algorithm & Mean & Standard Deviation & Mean & Standard Deviation\\
        \midrule
        Known Map & 12.88 & \hphantom{0}0.06 & 18.32 & \hphantom{0}3.54\\
        With Language (Ours) & 16.64 & \hphantom{0}6.84 & 82.78 & 10.56\\
        Without Language & 25.28 & 12.99 & 85.57 & 17.80\\
        \bottomrule
    \end{tabularx} \label{tab:wheelchair-results-sim}
\end{table}
Table~\ref{tab:wheelchair-results-sim} compares the total distance traveled along with the execution time for the three methods, averaged over ten Monte Carlo simulations, along with the standard deviation. As expected, with access to the world model, the ``Known Map'' baseline has the robot navigate directly to the desired goal, achieving the shortest path and smallest execution time. In contrast, the ``Without Language'' baseline continues to ground the instructions in incomplete maps, often choosing to initially explore the second (incorrect) hallway, before opportunistically discovering the kitchen. This results in longer average paths (and higher standard deviation) as well as significantly higher execution times (also with higher standard deviation). By taking advantage of the environment knowledge implicit in the command, our method enables the robot to act more deliberately, reaching the intended goal along paths that are only slightly longer than those of the ``Known Map'' baseline. However, our framework does
require significantly more time to follow the directions than the known
map scenario. In part, this increase results from the robot stopping each time it reaches an intermediate goal selected by the policy, at which time the algorithm updates the semantic map distribution, grounds the instruction to a set of behaviors, and then evaluates the policy to identify the next action. In contrast, the robot navigates without stopping until it reaches the goal with the known map baseline. The additional computational requirements of our framework will inherently result in larger runtimes, however we note that a non-negligible fraction of the additional time is due to our implementation, which explicitly required the robot to pause for several seconds before moving on to the next waypoint.

\subsubsection{Physical Experiments}
\label{sec:results-following-natural-language-route-instructions}

\begin{figure}[!t]
  \centering
  \subfigure[]{\includegraphics[width=0.325\linewidth]{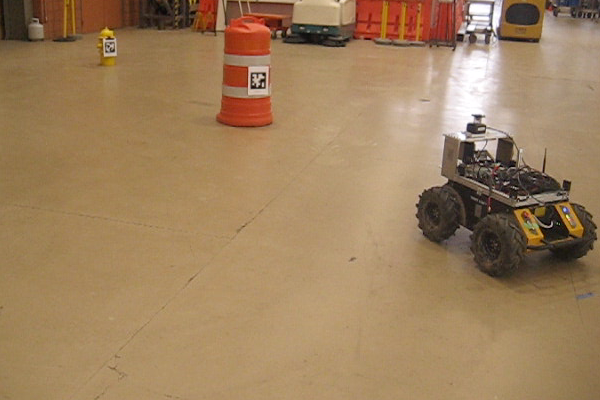}\label{fig:husky-indoor-experiment-tp-1}}\hfil
  \subfigure[]{\includegraphics[width=0.325\linewidth]{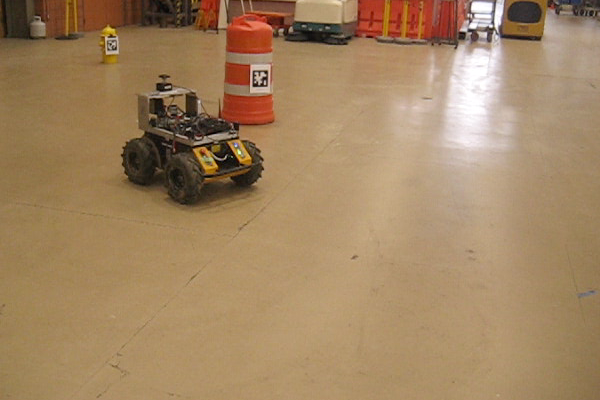}\label{fig:husky-indoor-experiment-tp-2}}\hfil
  \subfigure[]{\includegraphics[width=0.325\linewidth]{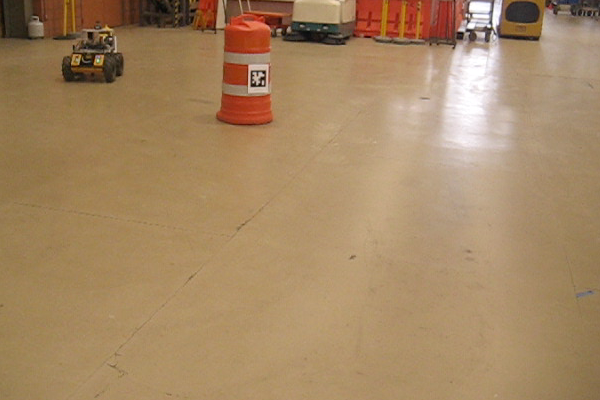}\label{fig:husky-indoor-experiment-tp-3}}\\
  \subfigure[]{\includegraphics[width=0.325\linewidth]{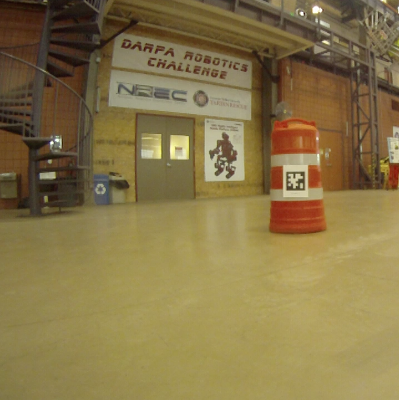}\label{fig:husky-indoor-experiment-fp-1}}\hfil
  \subfigure[]{\includegraphics[width=0.325\linewidth]{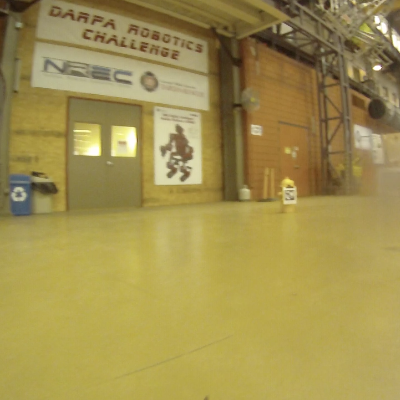}\label{fig:husky-indoor-experiment-fp-2}}\hfil
  \subfigure[]{\includegraphics[width=0.325\linewidth]{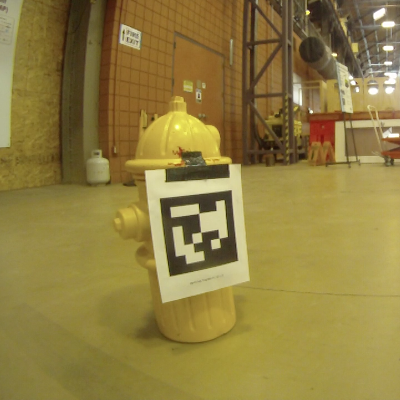}\label{fig:husky-indoor-experiment-fp-3}}
  \caption{Third- (top) and first-person (bottom) perspectives of a Clearpath Robotics Husky A200 Unmanned Ground Vehicle responding to the command ``navigate to the hydrant behind the barrel'' in an a priori unknown environment.  Subfigures \subref{fig:husky-indoor-experiment-tp-1} and \subref{fig:husky-indoor-experiment-fp-1} show the third- and first-person perspective when the robot receives the initial instruction.  Note that the hydrant is not in view, so that object is not available for grounding in the baseline approach.  Subfigures~\subref{fig:husky-indoor-experiment-tp-2} and \subref{fig:husky-indoor-experiment-fp-2} followed by \subref{fig:husky-indoor-experiment-tp-3} and \subref{fig:husky-indoor-experiment-fp-3} show the robot as it navigates to a hypothesized hydrant behind the observed barrel.  Once the hydrant becomes visible and is placed in the environment model, the distribution converges to the visually observed state.}
  \label{fig:husky-indoor-experiment}
\end{figure}
We further evaluate our method through a series of experiments in which different robots were tasked with following natural language navigation instructions in a priori unknown environments. The first set of experiments emulates the aforementioned Monte Carlo experiments in which a Clearpath Husky A200 Unmanned Ground Vehicle (Fig.~\ref{fig:husky-nrec}) and a voice-commandable wheelchair (Fig.~\ref{fig:wheelchair-stata-outdoors})\footnote{The wheelchair employs a cloud-based speech recognizer to convert spoken instructions to text, which is then provided as input to our architecture. The platform also supports limited onboard recognition~\citep{hetherington07} in the event that the cloud-based recognizer is unavailable.} were instructed to navigate to an unknown object. As with the simulation-based experiments, these commands did not match those used to train our language grounding models.

Each experiment involves a variation in the number and position of various objects in the environment (namely, cones and hydrants), the command, as well as changes in the robot's field-of-view. Figure~\ref{fig:husky-indoor-experiment} shows one such experiment in which the Clearpath Husky robot is instructed to navigate to the hydrant behind the cone. Initially, only the cone is visible to the robot (Figs.~\ref{fig:husky-indoor-experiment-tp-1} and \ref{fig:husky-indoor-experiment-fp-1}), at which point the algorithm hypothesizes the location of the hydrant. As the robot navigates according to the world model distribution, it detects the presence of a hydrant (Figs.~\ref{fig:husky-indoor-experiment-tp-2} and \ref{fig:husky-indoor-experiment-fp-2}), and then drives straight to the goal (Figs.~\ref{fig:husky-indoor-experiment-tp-3} and \ref{fig:husky-indoor-experiment-fp-3}). For each configuration of the environment, command, and field-of-view, we perform ten trials with our algorithm with the wheelchair and six with the Husky. As a baseline, we perform an additional run with a completely known world model. We consider a run to be a success when the robot’s final destination is within 1.5\,m of the intended goal.

\begin{table}[!h]
    \newcolumntype{S}{c@{\hskip 0.15in}}
    \newcolumntype{B}{c@{\hskip 0.17in}}
    \newcommand{\hp}[1]{\leavevmode\hphantom{#1}}
    \centering
    \caption{Wheelchair object-relative navigation experimental results with $1\sigma$ confidence intervals.}
    \begin{tabularx}{0.9\textwidth}{llSSSSBBB}
        \toprule
        &&&& \multicolumn{2}{c}{Success Rate (\%)} && \multicolumn{2}{c}{Distance
          (m)}\\
        \cmidrule{5-6} \cmidrule{8-9}
        \multicolumn{2}{c}{Environment} & FOV (m) & Relation & Known & Ours && Known & Ours\\
        \midrule
        1 hydrant & 1 cone & 2.5 & null & 100.0 & 100.0 && 4.69 & 16.56 (7.20)\\
        1 hydrant & 1 cone & 2.5 & ``behind'' & 100.0 & 100.0 && 4.69 & \hp{0}9.91 (3.41)\\
        1 hydrant & 2 cones & 3.0 & ``behind'' & 100.0 & 100.0 && 4.58 & \hp{0}7.64 (2.08)\\
        2 hydrants & 1 cone & 2.5 & ``behind'' & 100.0 & \hp{0}80.0 &&      5.29 & \hp{0}6.00 (1.38)\\
        2 hydrants & 1 cone & 4.0 & ``nearest''  & 100.0 & 100.0 && 4.09 & \hp{0}4.95 (0.39)\\

        2 hydrants & 1 cone & 3.0 & ``nearest'' & 100.0 & \hp{0}50.0 && 6.30 & \hp{0}7.05 (0.58)\\
        \bottomrule
    \end{tabularx}
    \label{tab:wheelchair-results}
\end{table}
Table~\ref{tab:wheelchair-results} presents the performance of our algorithm averaged over ten runs per scenario with the wheelchair, and compares against the baseline that has full knowledge of the environment. Experiments with the Husky demonstrate similar results with an average distance traveled of 8.1\,m ($\sigma =$ 1.3\,m) with our method compared to 8.4\,m ($\sigma=$ 0.6\,m) with the known map baseline and a success rate of 83.3\% v.s.\ 100\% with the baseline, based on six runs each. Together, the results demonstrate that our algorithm is able to take advantage of spatial relations that may be implicit in the instructions to identify more informed, deliberate paths to the goal. The ability to leverage information about spatial relations is important when there are multiple objects in the environment that match the figure in the instruction. For example, when there were two hydrants and the user commands the robot to ``go to the hydrant behind the cone'', the robot successfully identifies the correct hydrant as the goal in eight of the ten experiments. Similar to the failure discussed above for the simulation-based experiments, the two failures occur when the robot initially sees only the incorrect hydrant, upon which the semantic map  hypothesizes the existence of cones in front of the hydrant. This results in a behavior distribution that is peaked around this goal. In the eight successful trials, the robot observes all three objects and infers the correct behavior. Similarly, if we consider the command ``go to the hydrant nearest to the cone'' we find that the robot reaches the goal in all ten experiments with a 4\,m field-of-view. However, reducing the field-of-view to 3\,m results in the robot reaching the goal in only half of the trials.

\begin{figure}[!t]
	\centering
	\subfigure[Voice-commandable wheelchair]{\includegraphics[height=2.4in]{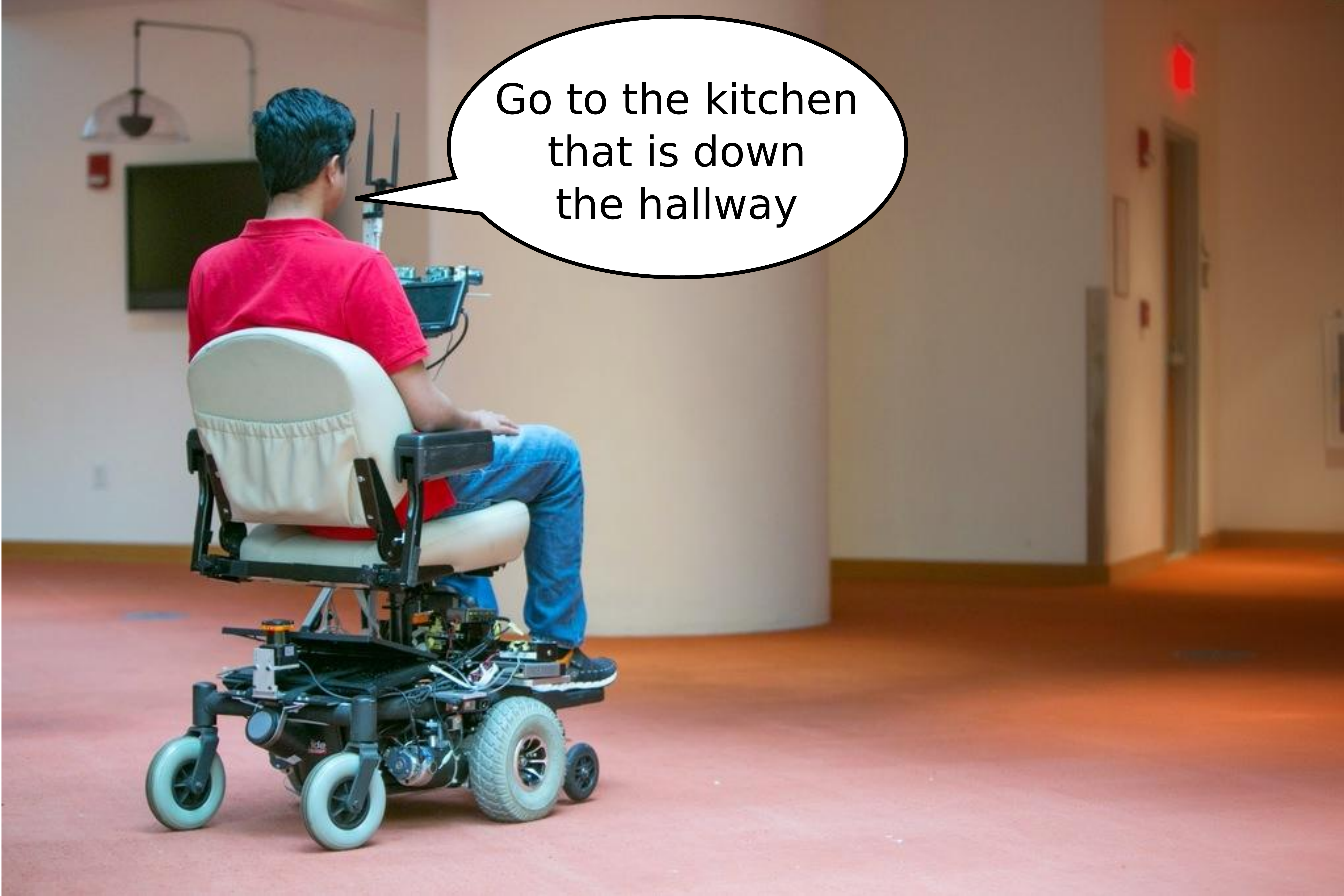}\label{fig:wheelchair}}\hfil%
	\subfigure[Occupancy grid map of the environment]{%
      \includegraphics[height=2.4in]{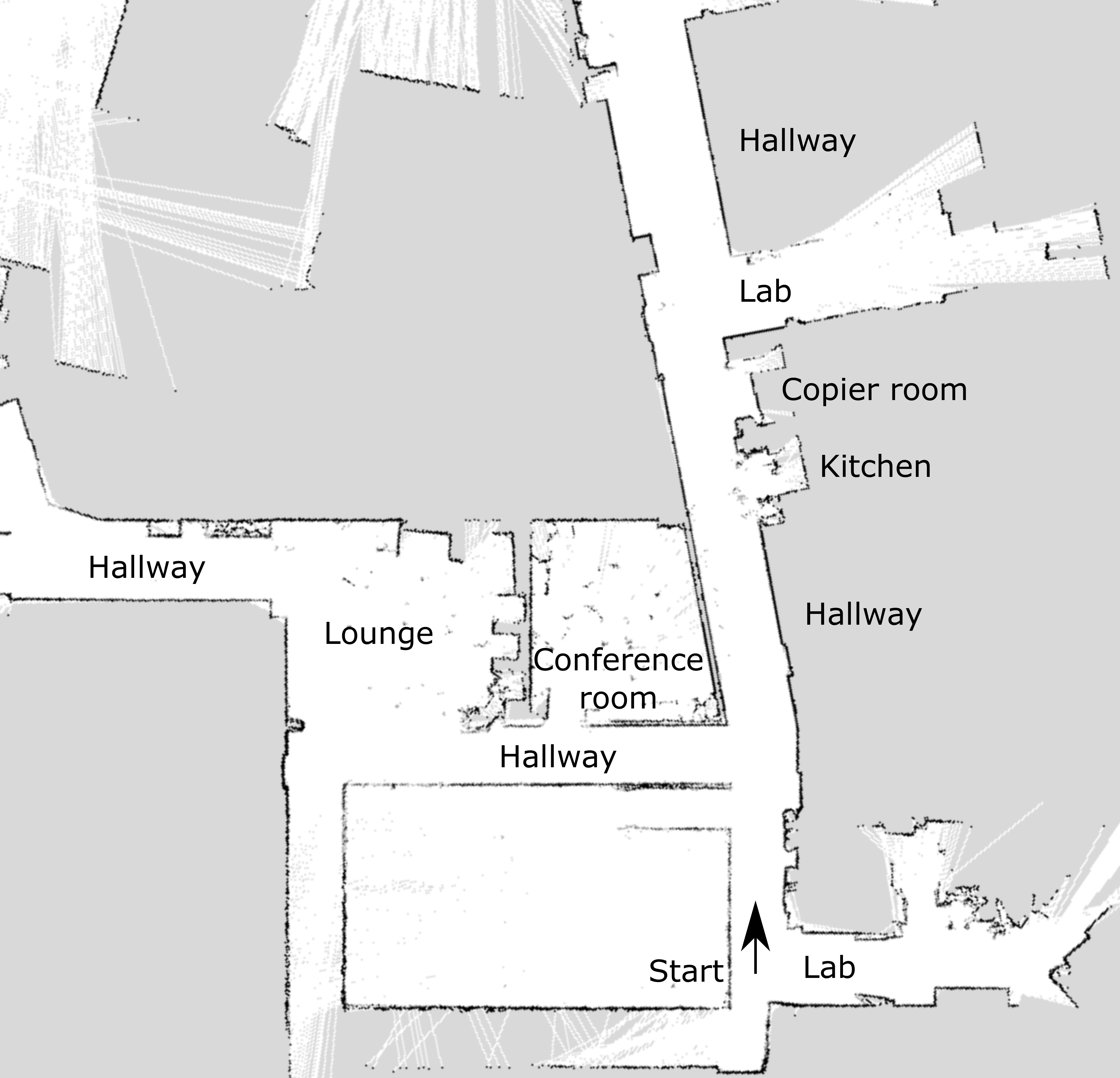}
      \label{fig:stata-map}%
	  }
	\caption{A \subref{fig:wheelchair} voice-commandable wheelchair was tasked with following natural language route instructions in \subref{fig:stata-map} a $\textrm{30\,m} \times \textrm{30\,m}$ (approx.) office-like environment consisting of multiple hallways and rooms. The wheelchair employs a cloud-based speech recognizer to convert spoken instructions to text, which is then provided as input to our architecture. The platform also supports limited onboard recognition~\citep{hetherington07} in the event that the cloud-based recognizer is unavailable.} \label{fig:wheelchair-experiments}
\end{figure}
Next, we evaluate our framework on a mobile robot tasked with following natural language route instructions in an unknown environment. We implemented our architecture on a voice-commandable wheelchair
(Fig.~\ref{fig:wheelchair}) that is equipped with three
forward-facing monocular cameras with a collective field-of-view of 120 degrees, and
forward- and rearward-facing Hokuyo UTM LIDARs. The wheelchair was placed in a lobby within MIT's Stata Center, with several
hallways, offices, and lab spaces, as well as a kitchen on the same
floor. In an effort to facilitate perception, these experiments employed AprilTag fiducials~\citep{olson2011apriltag} to identify the existence and semantic
type of regions
in the environment. We trained the HDCG models on a parallel corpus of
54 fully labeled examples.  We then directed the wheelchair to execute the
instruction ``go to the kitchen that is down the hallway'' that was not seen during training.

\begin{figure}[!p]
    \centering
    {\setlength{\fboxsep}{0pt}%
    \setlength{\fboxrule}{0.5pt}
    \subfigure[$t = 0$\, sec]{\fbox{\includegraphics[width=0.235\linewidth]{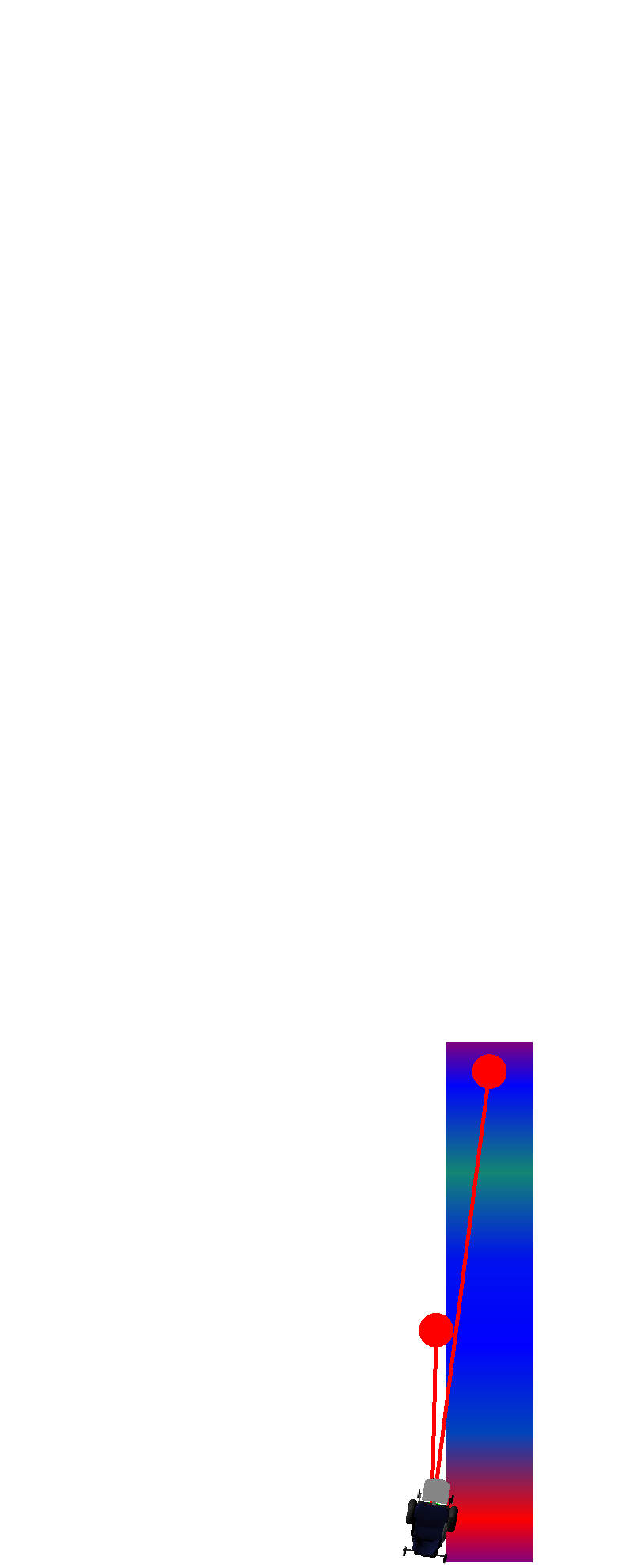}}}\hfil
    \subfigure[$t = 15$\, sec]{\fbox{\includegraphics[width=0.235\linewidth]{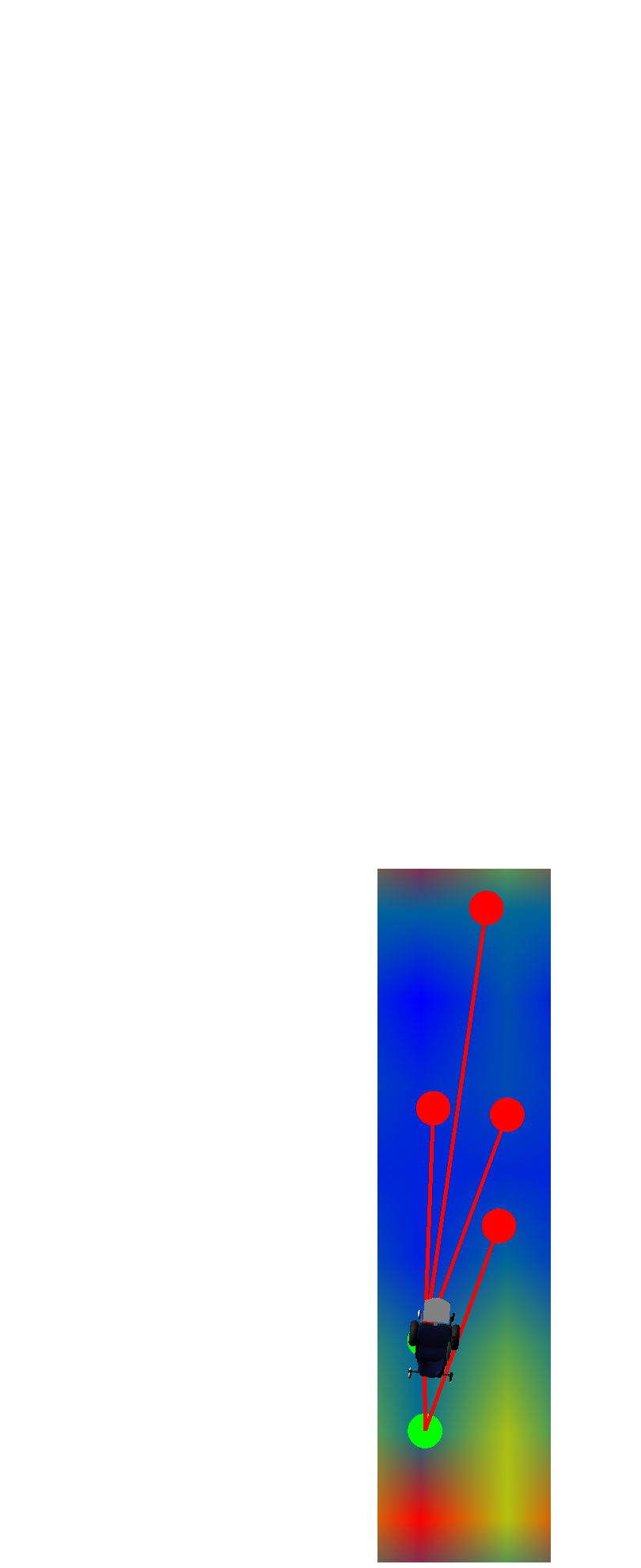}}}\hfil
    \subfigure[$t = 38$\, sec]{\fbox{\includegraphics[width=0.235\linewidth]{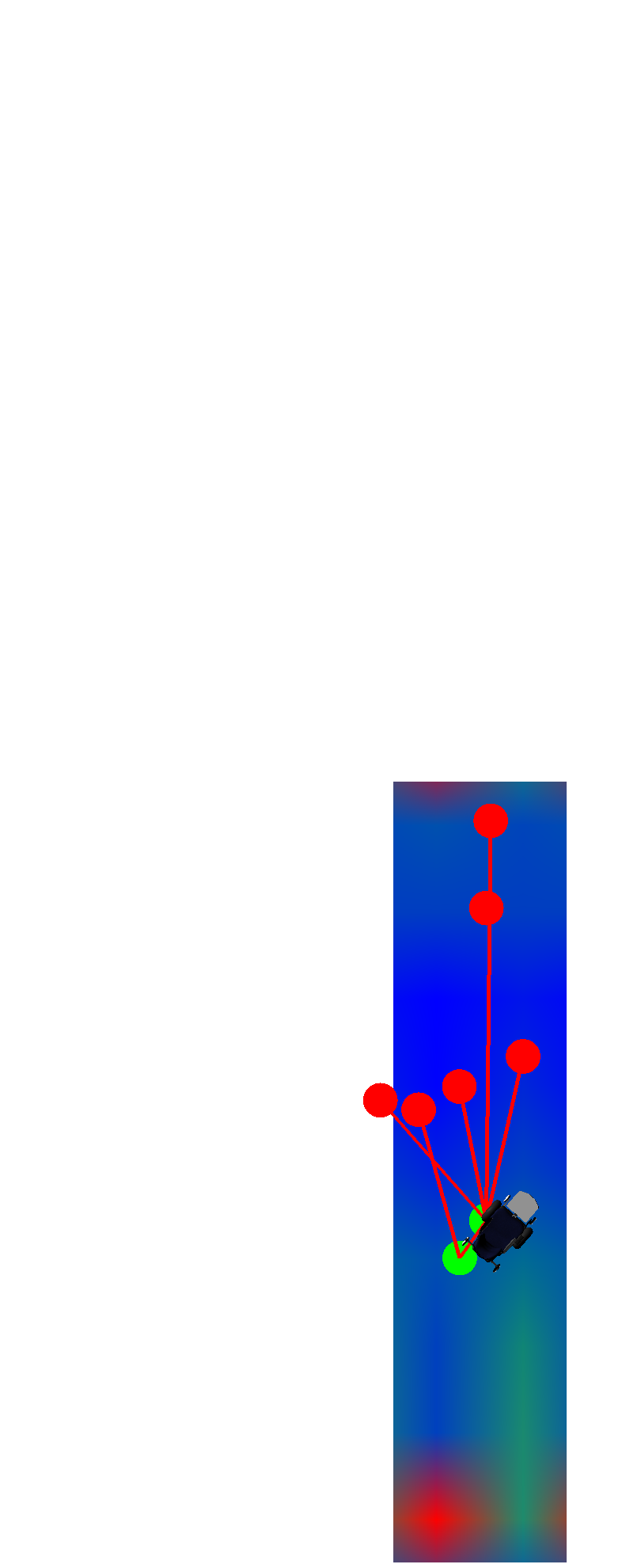}}}\hfil
    \subfigure[$t = 65$\, sec]{\fbox{\includegraphics[width=0.235\linewidth]{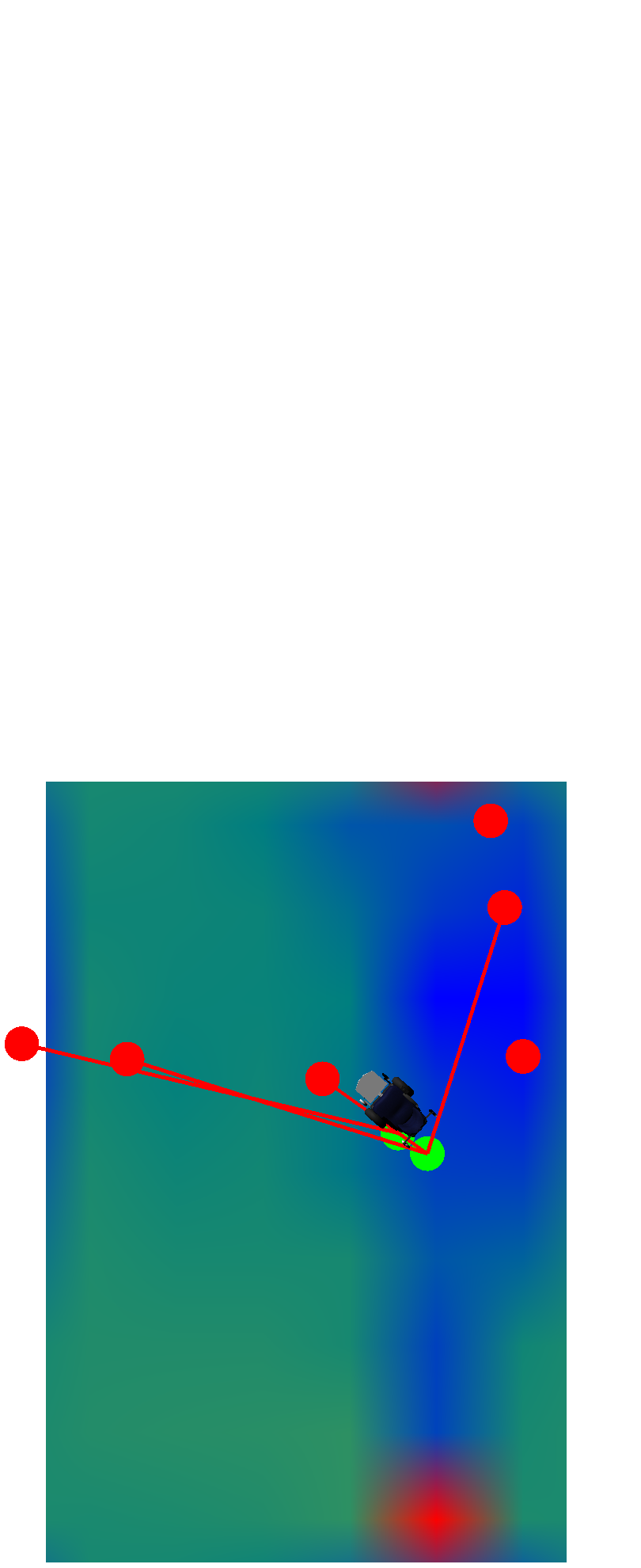}\label{fig:wheelchair-hallway-cost-function-wrong-hallway}}}\\
    \subfigure[$t = 85$\, sec]{\fbox{\includegraphics[width=0.235\linewidth]{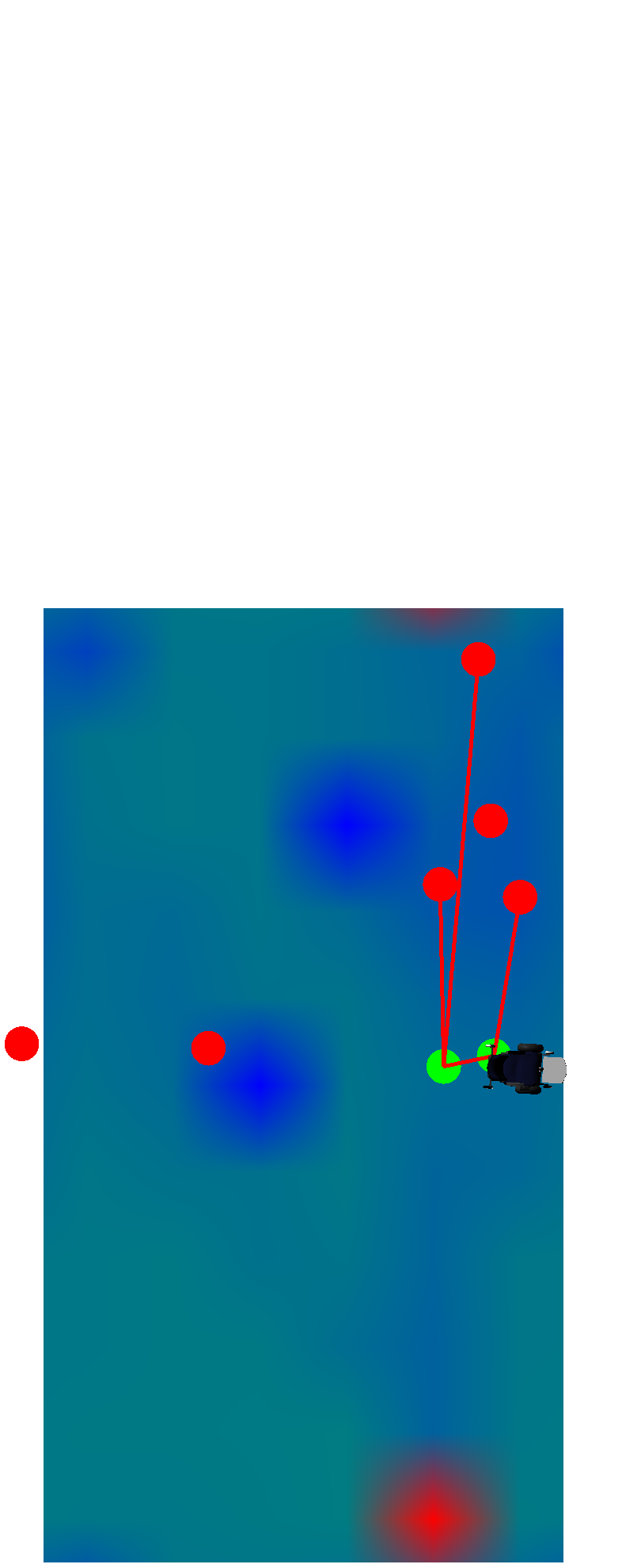}}}\hfil
    \subfigure[$t = 102$\, sec]{\fbox{\includegraphics[width=0.235\linewidth]{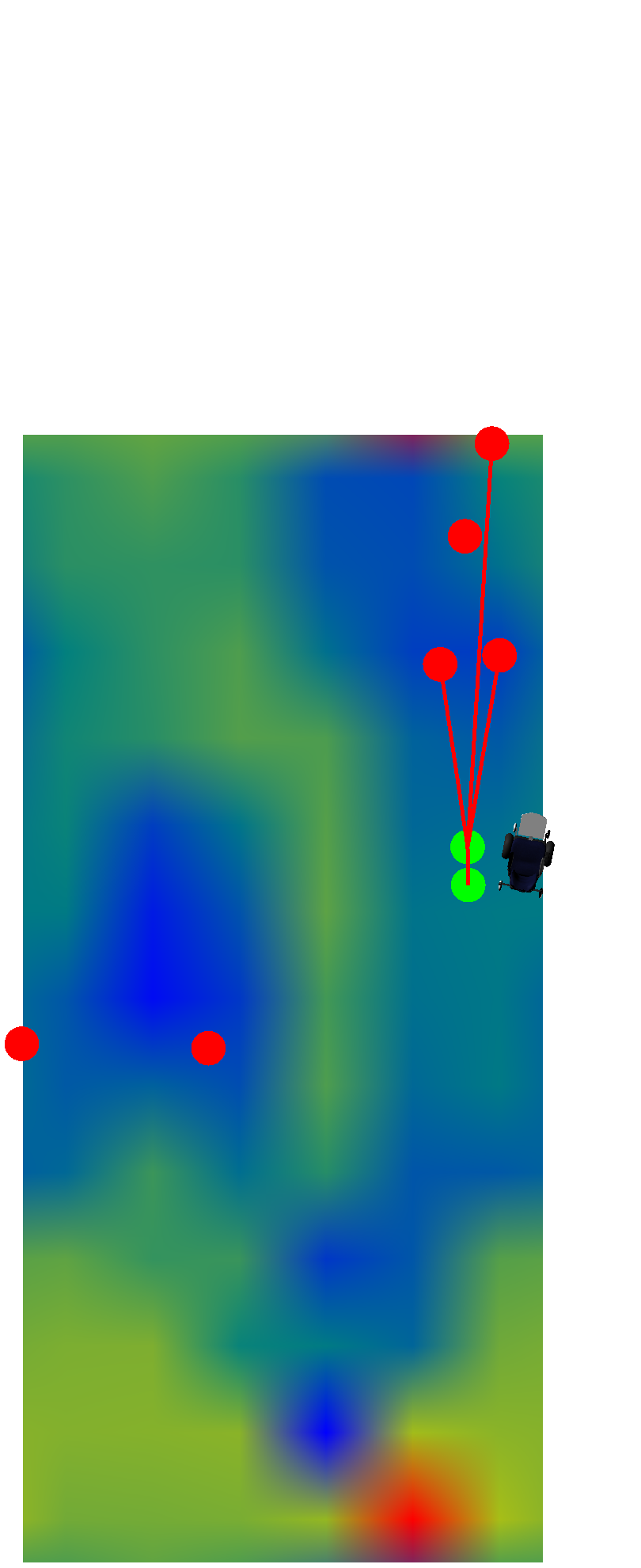}\label{fig:wheelchair-hallway-cost-function-correct-hallway}}}\hfil
    \subfigure[$t = 120$\, sec]{\fbox{\includegraphics[width=0.235\linewidth]{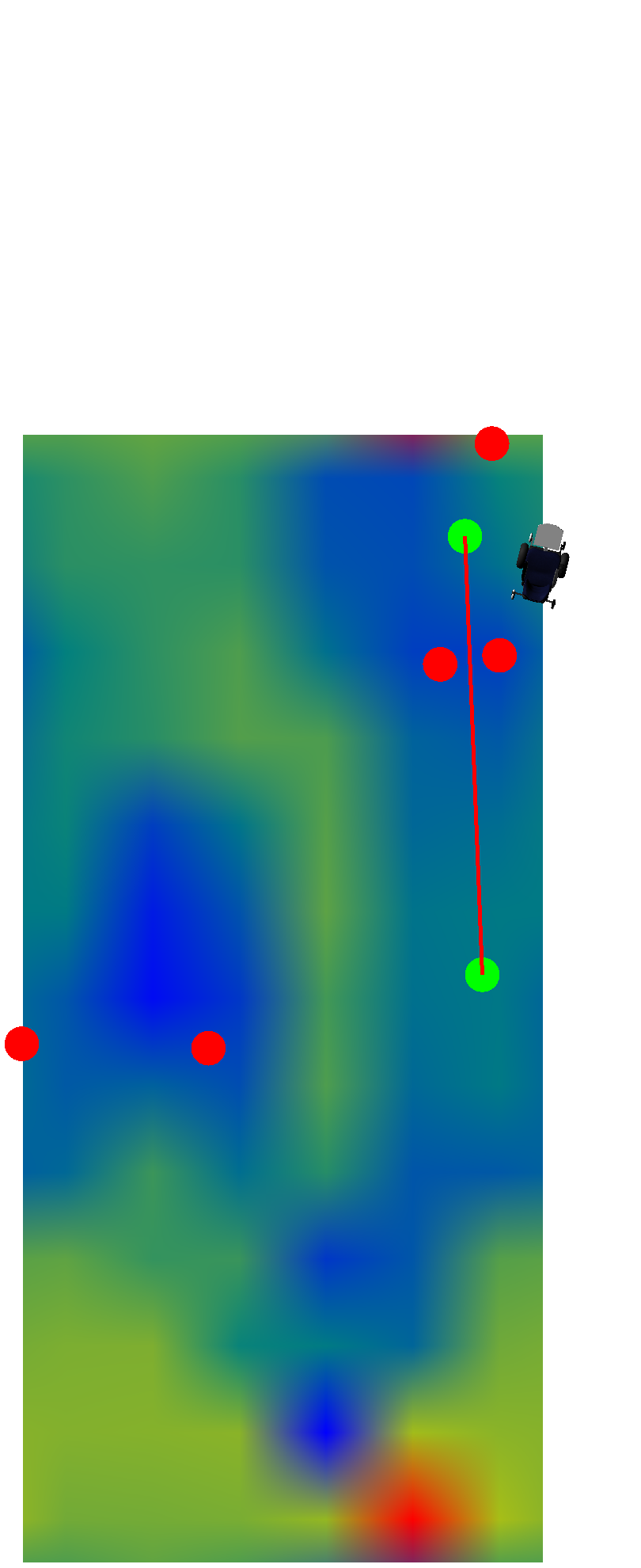}}}\hfil
    \subfigure[$t = 140$\, sec]{\fbox{\includegraphics[width=0.235\linewidth]{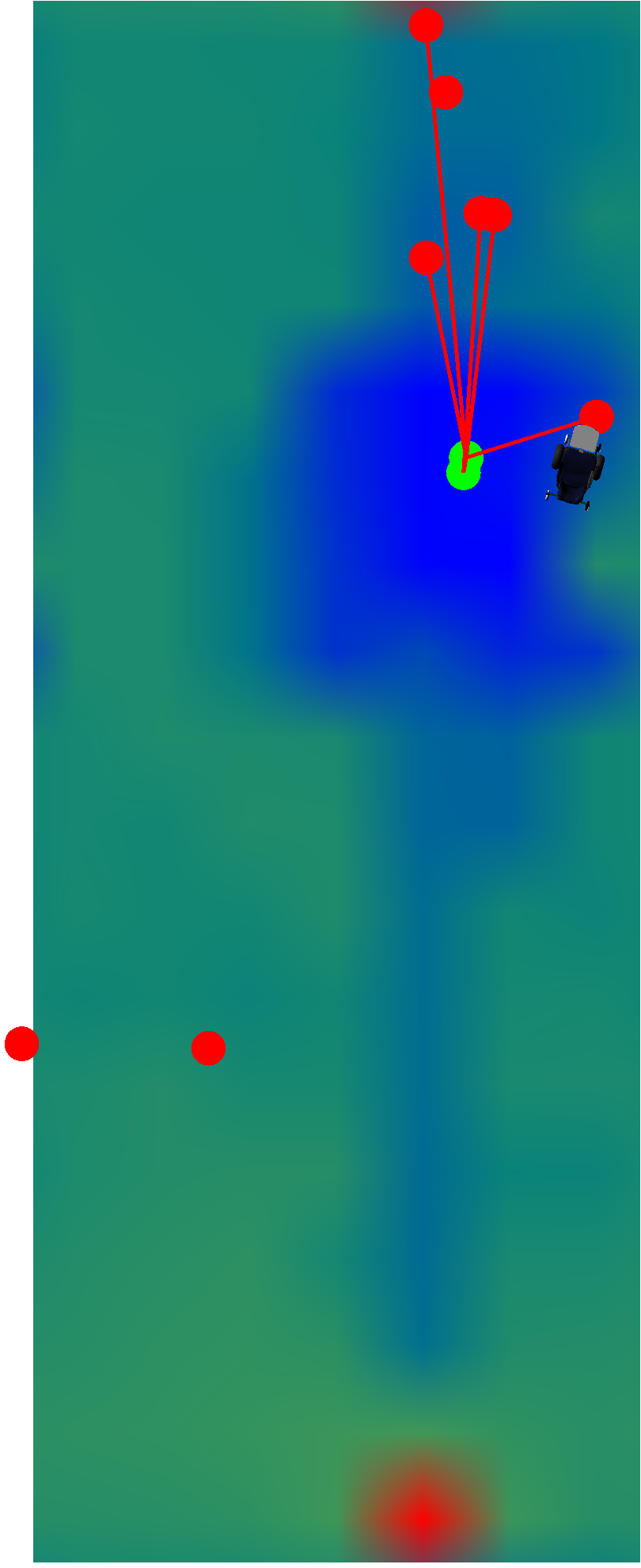}\label{fig:wheelchair-hallway-cost-function-done}}}\\
    }
    \caption{The evolution of the planner cost function for the command ``go to the kitchen that is down the hallway''. Red nodes indicate candidate destinations and green nodes indicate previously visited locations. The normalized cost function is rendered using the colormap: \protect\includegraphics[width=1in]{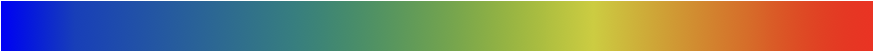}, where blue and red denote costs of $0$ and $1$, respectively. The planner \subref{fig:wheelchair-hallway-cost-function-wrong-hallway} initially directs the robot down the wrong hallway, but after not seeing the hypothesized kitchen, \subref{fig:wheelchair-hallway-cost-function-correct-hallway} the robot navigates down the correct hallway \subref{fig:wheelchair-hallway-cost-function-done} to the goal.}
    \label{fig:wheelchair-hallway-cost-function}
\end{figure}
\begin{figure}[!t]
    \centering
    \includegraphics[width=0.325\linewidth]{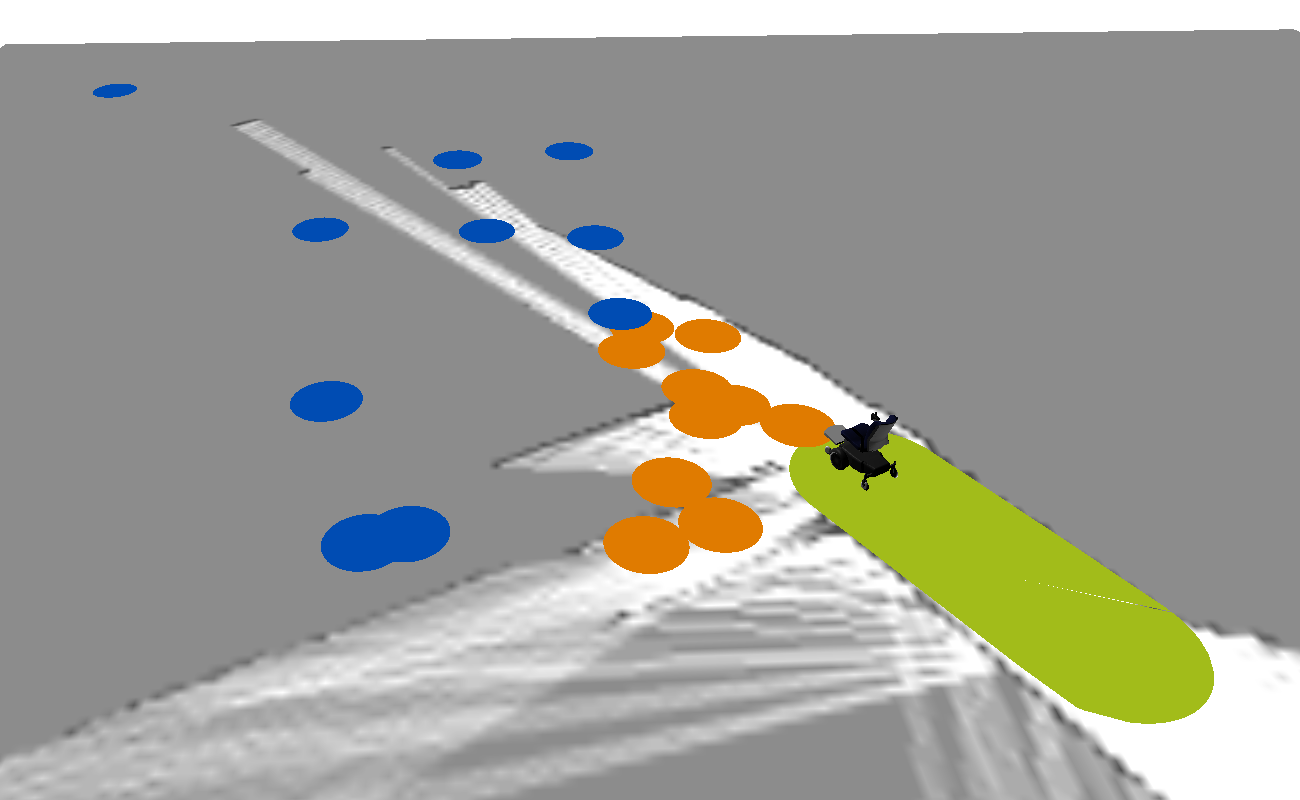}\hfil%
    \includegraphics[width=0.325\linewidth]{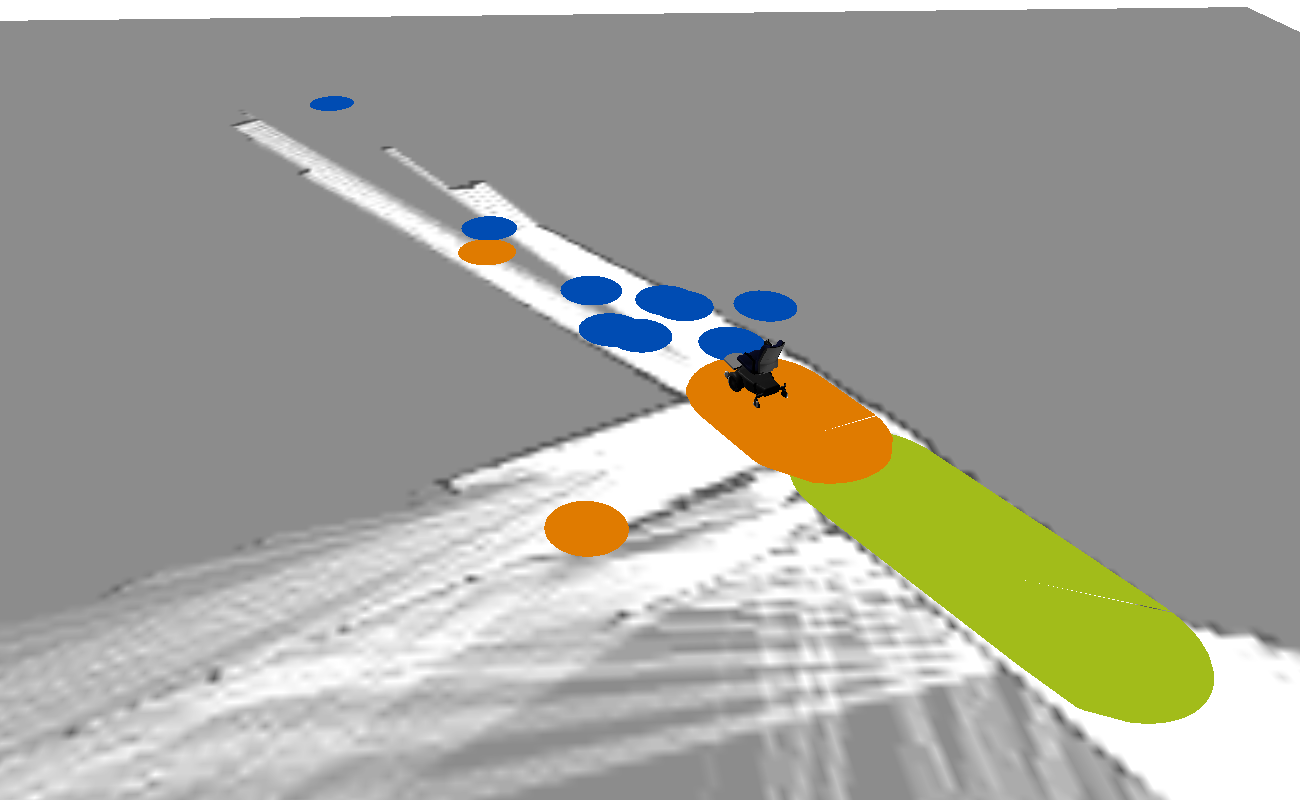}\hfil%
    \includegraphics[width=0.325\linewidth]{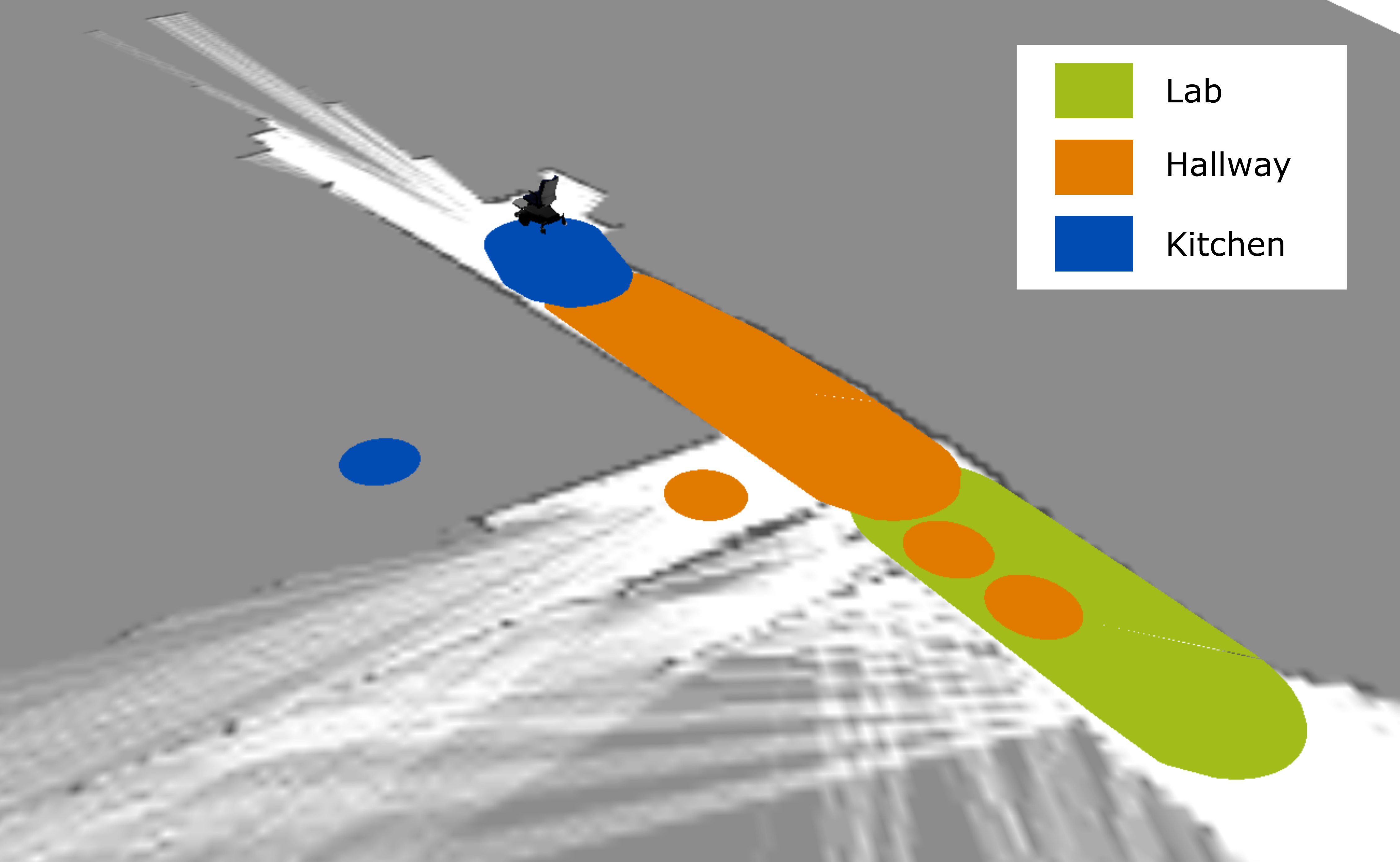}
    \caption{A visualization that shows how the semantic map evolves time as the robot follows the command ``go to the kitchen that is down the hallway.'' Small circles and large filled-in areas denote sampled and visited regions, respectively, each colored according to its class type. The robot (left) first samples possible locations of the kitchen and moves towards them, (middle) then observes the hallway and refines its estimate using the ``down'' relation provided by the user. Finally, the robot (right) reaches the actual kitchen and declares it has finished following the direction.} \label{fig:wheelchair-hallway-map-evolution}
\end{figure}
As with the previous experiments, we compare our framework with two baselines. The ``Known Map'' baseline emulates the previous
state-of-the-art and uses a known map of the environment in order to
infer the actions consistent with the route
direction. The second ``Without Language'' baseline assumes no prior knowledge of the environment
(as with ours) and opportunistically grounds the command in the map, but
does not use language to modify the map. We performed six experiments with our algorithm, three with the
known map method, and five with the method that does not use
language for environment inference, all of which were successful (the robot reached the kitchen). Figure~\ref{fig:wheelchair-hallway-cost-function} visualizes the evolution of the planner's cost function for one of the experiments. The cost, which is a function of the semantic map distribution and the inferred behavior(s), initially suggests that the wheelchair navigate down the wrong hallway (Fig.~\ref{fig:wheelchair-hallway-cost-function-wrong-hallway}), but after not observing the kitchen, the map updates and the planner later leads the robot to the correct goal. Figure~\ref{fig:wheelchair-hallway-map-evolution} shows a visualization of the semantic
maps over several time steps for one successful run on the robot.

\begin{table}[!h]
    \newcolumntype{S}{c@{\hskip 0.22in}}
    \centering
    \caption{Evaluation of natural language route direction-following with the wheelchair.}
    \begin{tabularx}{0.825\linewidth}{Scccc}
        \toprule
        & \multicolumn{2}{c}{Distance (m)} & \multicolumn{2}{c}{Time (sec)}\\
        \cmidrule{2-3} \cmidrule{4-5}
        Algorithm & Mean & Standard Deviation & Mean & Standard Deviation\\
        \midrule
        Known Map & 13.10 & 0.67 & \hphantom{0}62.48 & 16.61\\
        With Language & 12.62 & 0.62 & 122.14 & 32.48\\
        Without Language & 24.91 & 13.55 & 210.35 & 97.73\\
        \bottomrule
    \end{tabularx} \label{tab:wheelchair-results-physical}
\end{table}
Table~\ref{tab:wheelchair-results-physical} compares the total distance traveled and
execution time for the three methods.
Our algorithm resulted in paths with lengths
close to those of the known map, and significantly outperformed the
method that did not use language for mapping. As with the Monte Carlo simulations, our framework required significantly more time to follow the directions than the known
map baseline. Some of this additional time results from the robot stopping each time it reaches an intermediate goal selected by the policy (in contrast to the known map baseline, which does not stop until it reaches the goal) at which time the algorithm updates the semantic map distribution, grounds the instruction to a set of behaviors, evaluates the policy to identify the next action, and then performs motion planning. While this will inherently result in larger runtimes compared to the known map setting, we note that a non-negligible fraction of the additional time (nearly half in some cases) is due to our implementation, which explicitly required the robot to pause for several seconds before moving on to the next waypoint.

\subsection{Instruction Following for Mobile Manipulation}
\label{sec:results-natural-language-understanding-for-mobile-manipulation}

\begin{figure}[!th]
	\centering
	\subfigure[indoor environment]{\includegraphics[width=0.49\linewidth]{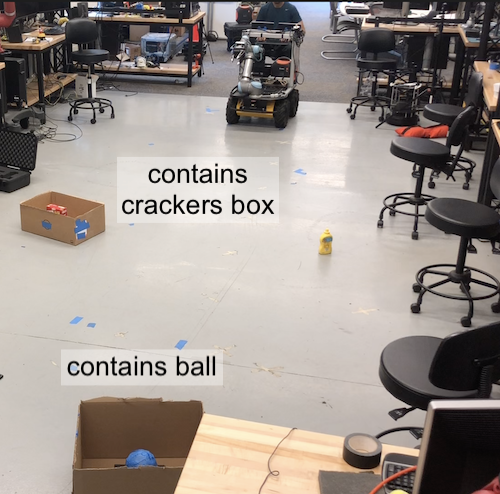}\label{fig:exp-husky-arm-indoor-environment}}\hfil%
	\subfigure[outdoor environment]{\includegraphics[width=0.49\linewidth]{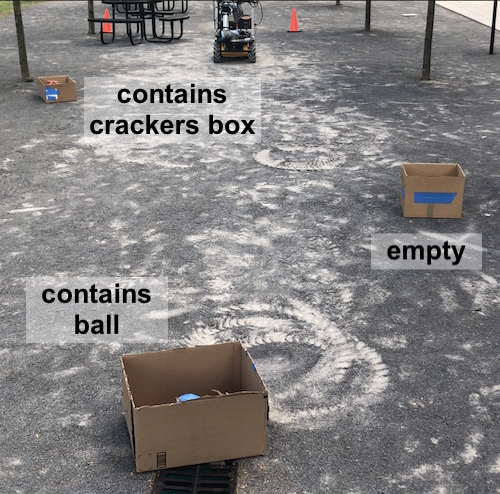}\label{fig:exp-husky-arm-outdoor-environment}}\\
	\subfigure[indoor map at $t = 57$\,sec]{\includegraphics[width=0.49\linewidth]{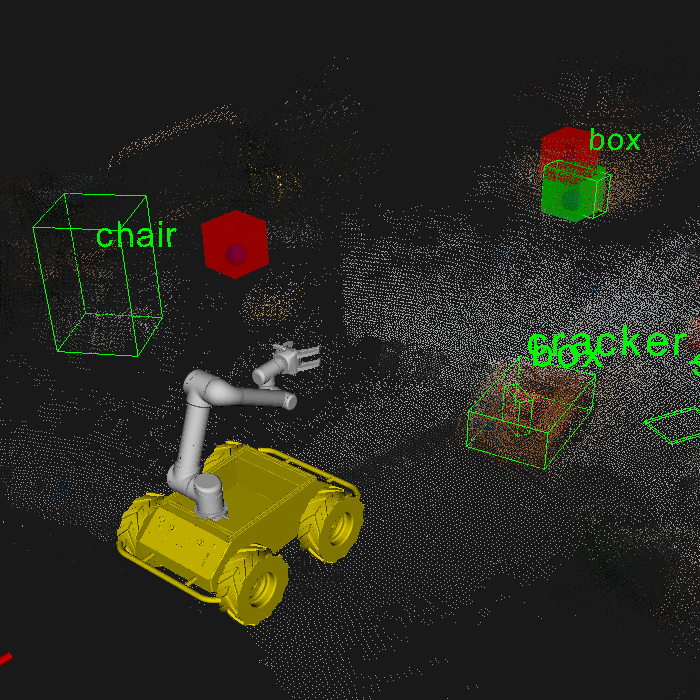}\label{fig:exp-husky-arm-indoor-map}}\hfil%
	\subfigure[outdoor map at $t = 177$\,sec]{\includegraphics[width=0.49\linewidth]{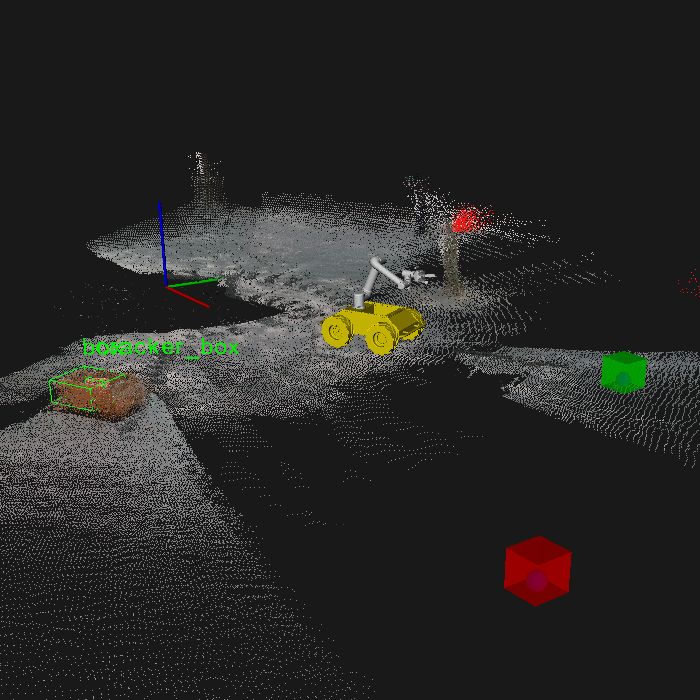}\label{fig:exp-husky-arm-outdoor-map}}
	\caption{The mobile manipulation experiments involved placing the arm-equipped Husky in a priori unknown \subref{fig:exp-husky-arm-indoor-environment} indoor and \subref{fig:exp-husky-arm-outdoor-environment} outdoor environments with several boxes that contained different or no objects. The robot was then given a natural language instruction to retrieve a specific object from a box. Without any prior knowledge of the environment, the robot initially navigates towards a hypothesized box (rendered as solid red cubes in the bottom figures). Upon detecting a box, the robot updates the world model distribution and then explores the nearest observed box (a green wireframe denotes objects that the robot has observed), which does not contain the object of interest. At this point, the robot either \subref{fig:exp-husky-arm-indoor-map} explores the next box that comes within the robot's field-of-view (for the indoor experiments), or \subref{fig:exp-husky-arm-outdoor-map} continues to explore the environment (for the outdoor experiments) as guided by the maintained world distribution.}
	\label{fig:husky-arm-environments}
\end{figure}
Having evaluated the performance of the proposed model through simulated experiments and in the real world by using fiducial based perception, we now expand upon our analysis by involving non-fiducial based perception, which is computationally more expensive and often not as accurate as fiducial-based perception. To analyze the runtime performance of the proposed framework for mobile manipulation tasks, we implemented the architecture on a Clearpath Husky A200 Unmanned Ground Vehicle fitted with a Universal Robotics UR5 arm and Robotiq 3-Finger Adaptive Robot Gripper~\ref{fig:husky-rochester}. Image data for visual perception was captured using an Intel RealSense RGB-D sensor mounted on the wrist of the UR5 arm. The perception pipeline consisted of multiple custom-trained YOLO-V3~\citep{yolov3} detectors. Specifically, we used a full YOLO-V3 model trained on the COCO~\citep{lin2014microsoft} dataset and 15 tiny YOLO-V3 models trained on individual classes from the YCB~\citep{calli2015benchmarking} and OpenImages-V4~\citep{OpenImages} datasets. We projected the 2D image-space bounding boxes generated by the YOLO-V3 detectors to the aligned 3D point clouds in order to obtain metric information about the objects in the robot's environment. Associated with each detected object was its semantic label generated by YOLO-V3 detector, a six degree-of-freedom pose estimate, and a 3D collision geometry represented by an oriented bounding box. The range of sensing was restricted to 4.5\,m indoors and 7.0\,m outdoors to eliminate noisy point cloud data.

Figure~\ref{fig:husky-arm-environments} illustrates the workspace setup for both indoor and outdoor environments. In both environments, the robot was initially instructed to ``retrieve the ball inside the box''. The indoor experiment followed with a second command to ``pick up the crackers box inside the box'', whereas the outdoor experiment followed with ``go to the crackers box''. In both of the settings, the box containing the ball was past the sensing horizon of the robot at the start, while the box containing the crackers box was observable. The workspace in the indoor environment was set up such that the robot will most likely detect and check the second box after inspecting the closer box and finding it empty. The larger workspace in the outdoor experiment allowed us to set up boxes far away from each other so that the robot would need to explore the environment using the hypothesized distribution to eventually reach the goal location. In order to build accurate models of the environment, a speed of 0.3\,m per perception cycle chosen for these experiments. We used 10 particles to represent the distribution over world models in the indoor experiment, while we used 20 particles for the outdoor experiment to account for the larger environment.

We developed and deployed a motion planner capable of performing manipulation actions that used TRAC-IK~\citep{Beeson-humanoids-15} for inverse kinematics. The architecture guided the robot to perform the most likely action identified through behavior inference. Upon reaching a box, the arm was positioned to look inside it. If this observation conveyed the absence of the object of interest, the robot would back up, pan the camera $\pm 30\deg$, and navigate towards a new goal selected from the updated distribution. If the desired object defined by the selected goal was observed in the box, the robot would execute the remaining activities to complete its inferred action.
\begin{figure}[!t]
	\centering
    \includegraphics[width=0.9\textwidth]{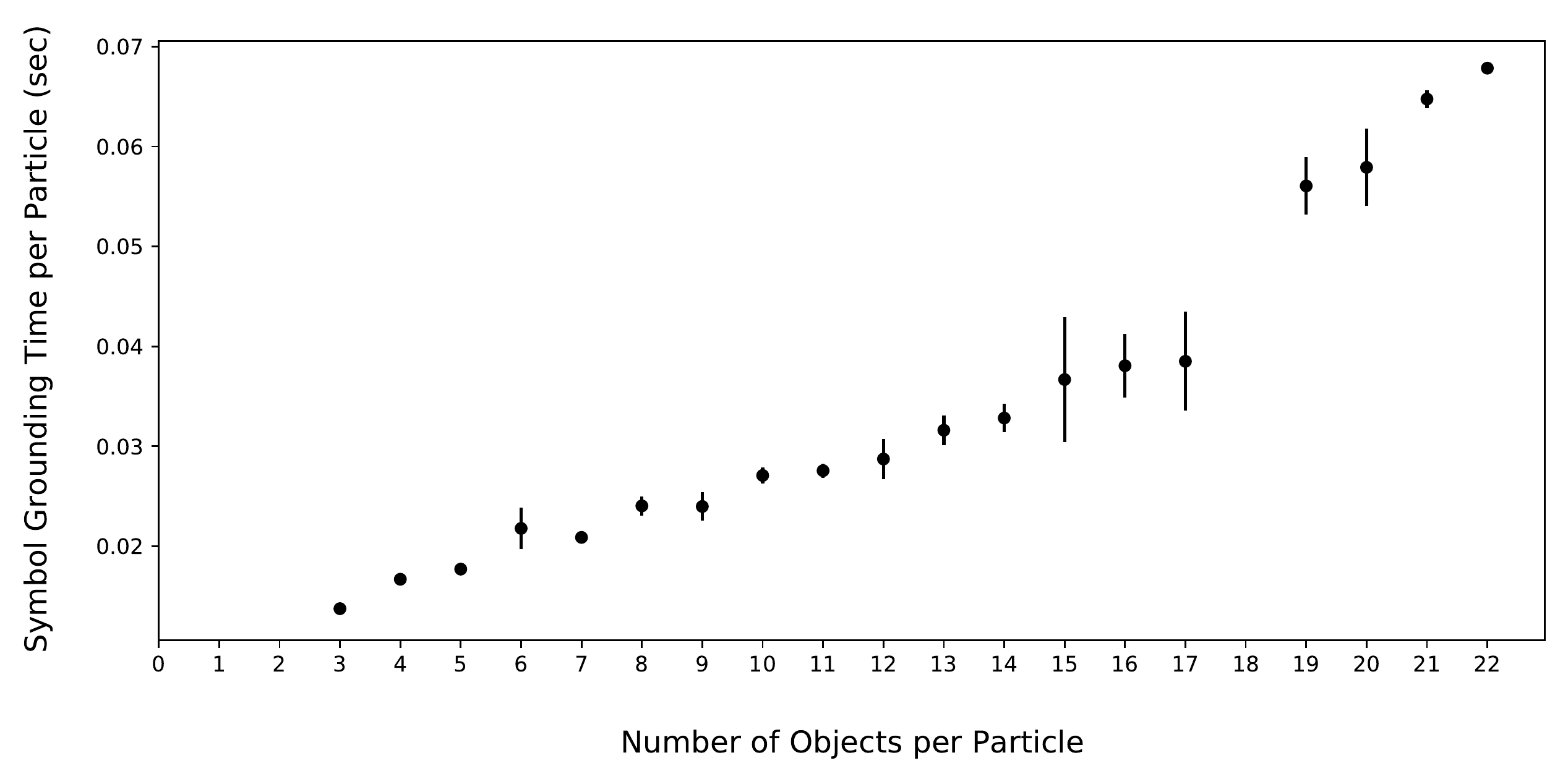}
	\caption{A graph that shows the increase in symbol grounding runtime per particle as a function of an increase in the number of detected objects in the particle.}
	\label{fig:exp-husky-arm-sg-vs-nobj}
\end{figure}
\begin{table}[!htb]
  \centering
  \caption{Perception and symbol grounding runtime analysis for indoor and outdoor trials (three each).}%
  \label{table:exp-husky-arm-results}
  \begin{tabularx}{1.0\linewidth}{Xcc}
      \toprule
      & Indoor & Outdoor \\
      \midrule
	  average behavior inference runtime per particle (seconds) & $0.035 \pm 0.01$  & $0.019\pm 0.002$\\
	  average perception cycle runtime (seconds) & $4.141 \pm 0.11$  & $4.099 \pm 0.060$\\
	  first task runtime (seconds) & $351.2$ & $593.5$\\
	  second task runtime (seconds) & $149.5$ & $\hphantom{0}20.4$\\
	  total number of detected objects & $24$ & $\hphantom{0}11$\\
	  total number of perception cycles during first task & $69$ & $127$\\
	  total number perception cycles during second task & $27$ & $\hphantom{00}3$\\
	  \bottomrule
  \end{tabularx}
\end{table}
\begin{figure}[!p]
  \centering
  \subfigure[Time $t = 0$ sec]{\includegraphics[width=0.495\linewidth]{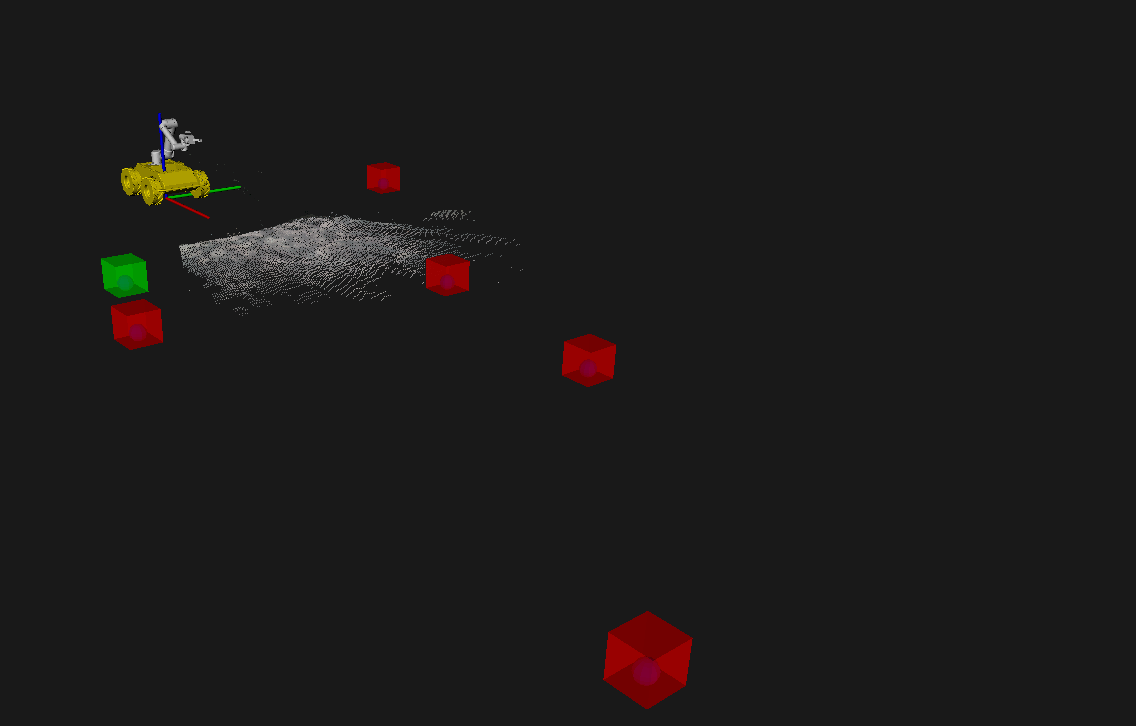}\label{fig:exp-husky-arm-results-outdoor-1}}\hfil%
  \subfigure[Time $t = 90$ sec]{\includegraphics[width=0.495\linewidth]{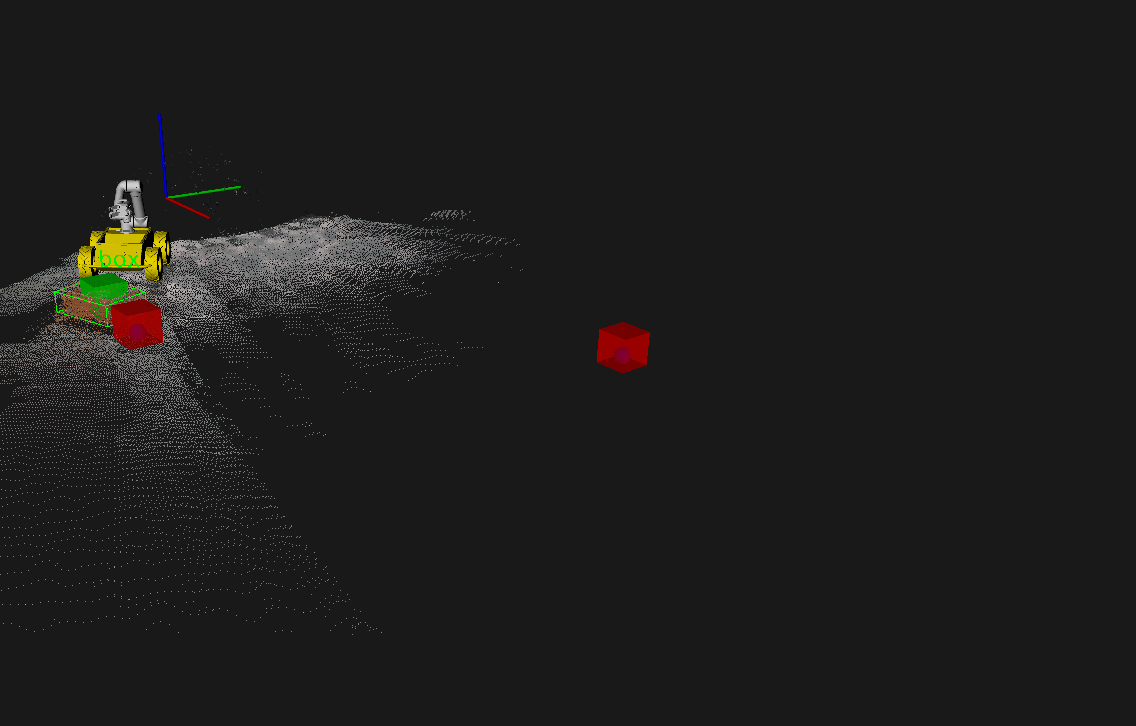}\label{fig:exp-husky-arm-results-outdoor-2}}\\%
  \subfigure[Time $t = 320$ sec]{\includegraphics[width=0.495\linewidth]{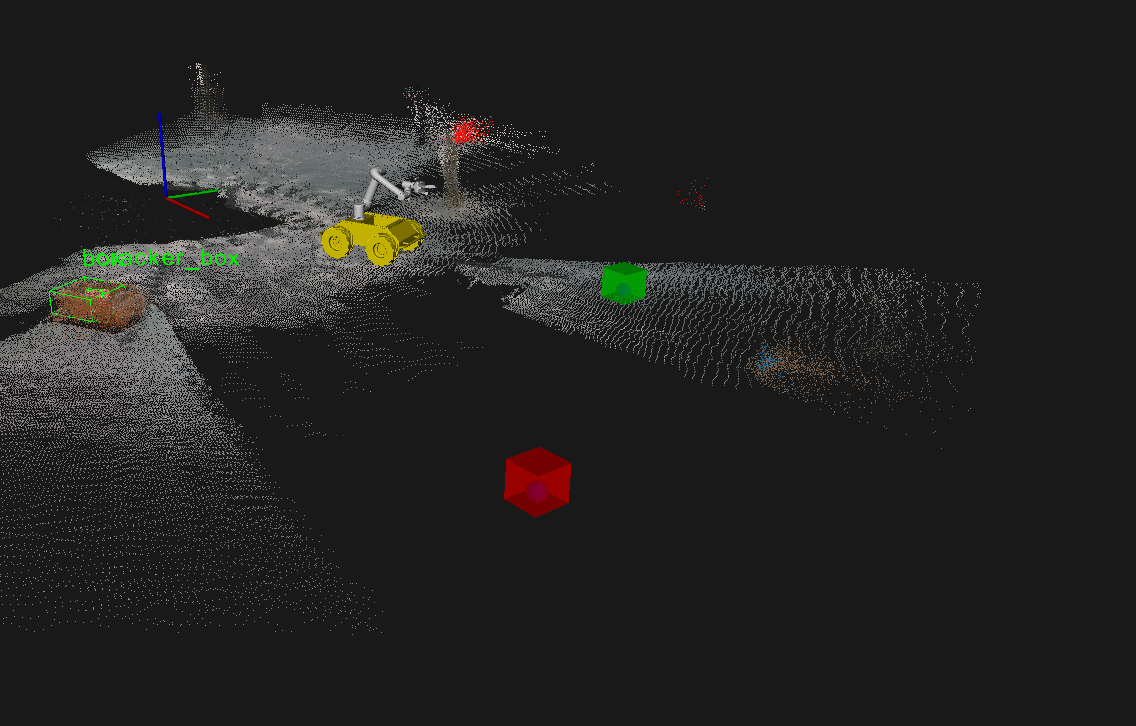}\label{fig:exp-husky-arm-results-outdoor-3}}\hfil%
  \subfigure[Time $t = 370$ sec]{\includegraphics[width=0.495\linewidth]{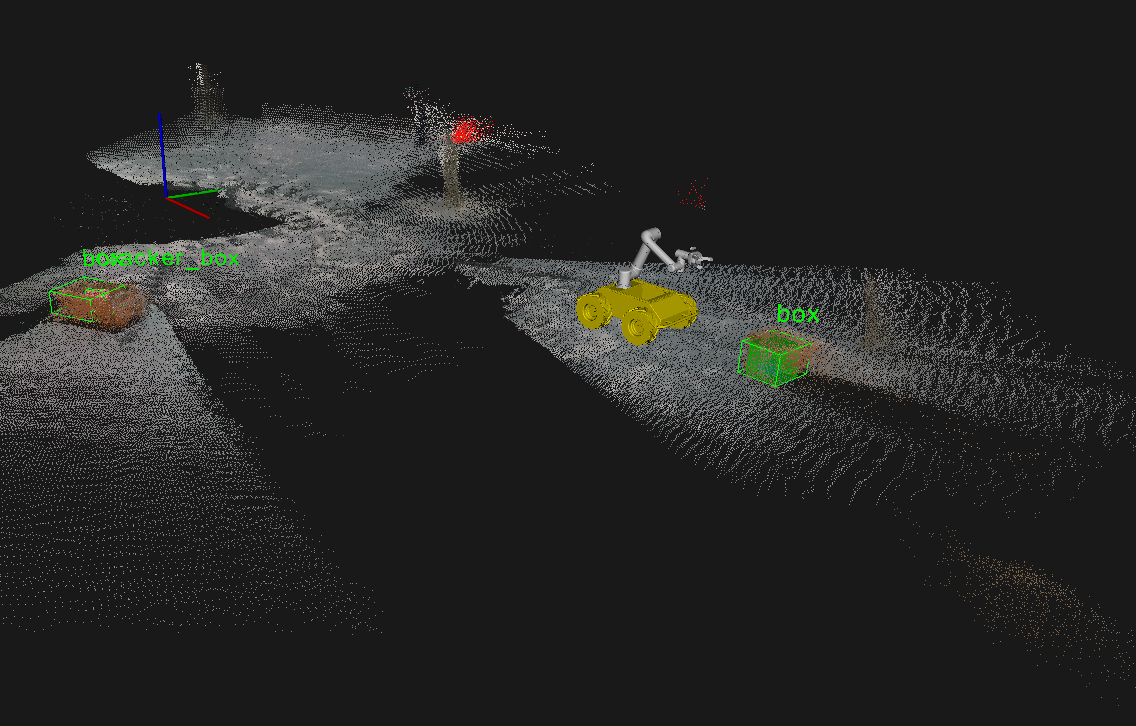}\label{fig:exp-husky-arm-results-outdoor-4}}\hfil\\%
  \subfigure[Time $t = 500$ sec]{\includegraphics[width=0.495\linewidth]{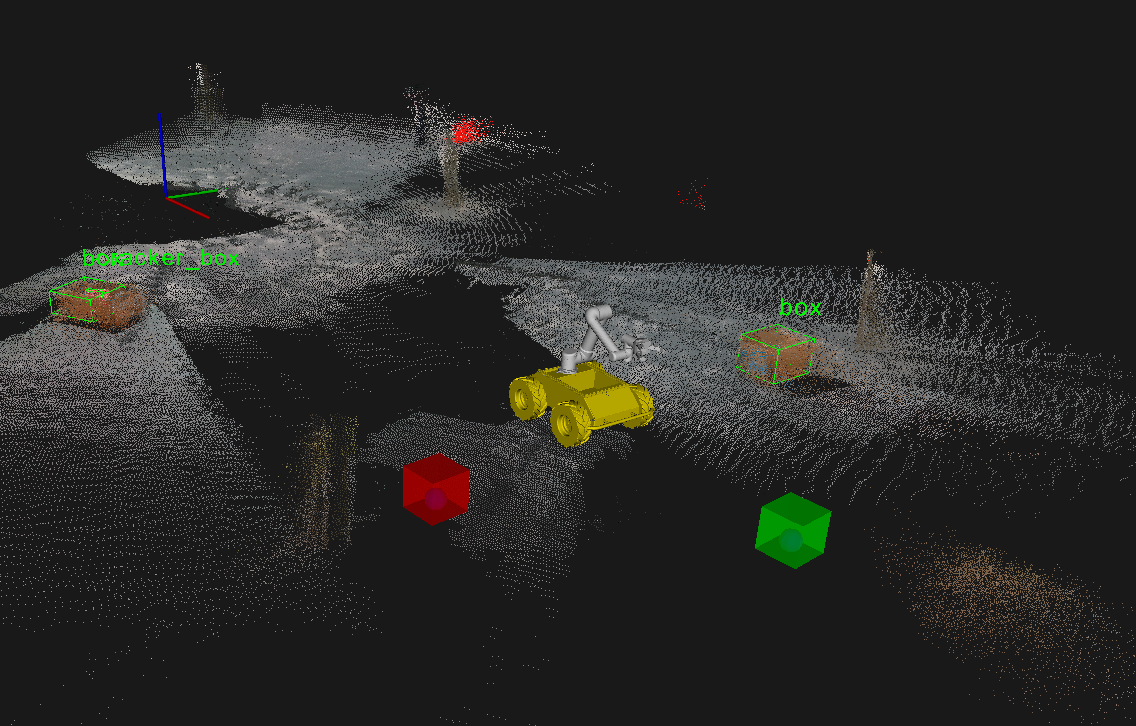}\label{fig:exp-husky-arm-results-outdoor-5}}\hfil%
  \subfigure[Time $t = 563$ sec]{\includegraphics[width=0.495\linewidth]{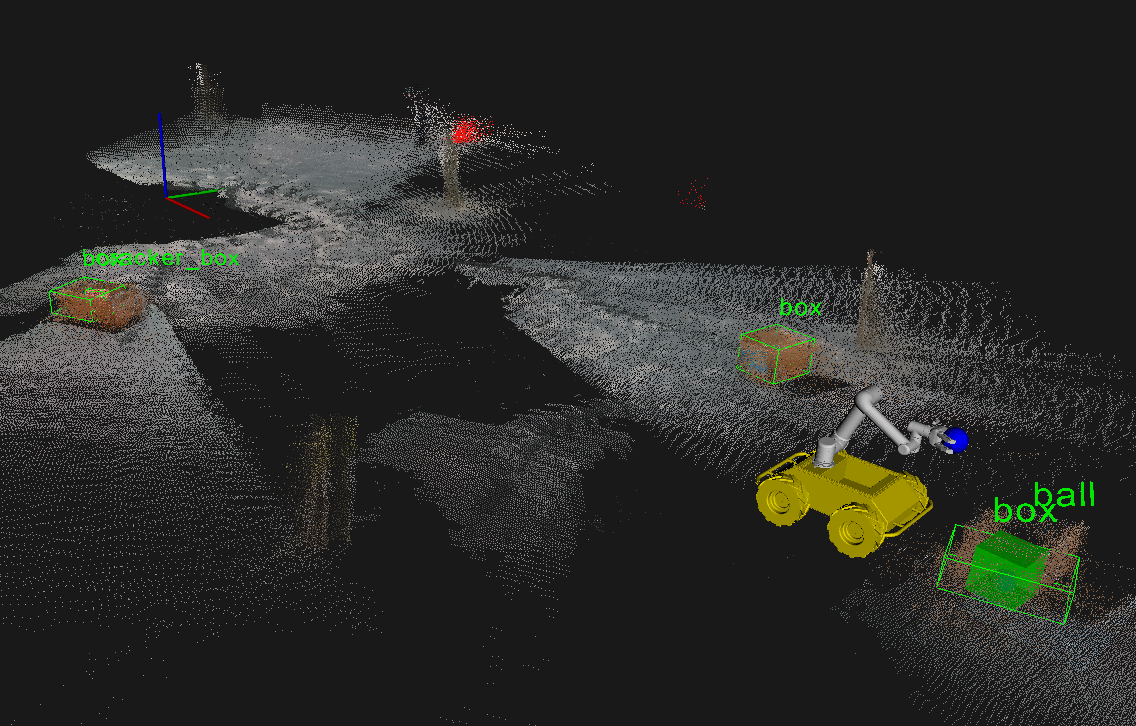}\label{fig:exp-husky-arm-results-outdoor-6}}%
  \caption{Visualization of the state of the robot while executing the instruction ``retrieve the ball inside the box'' in the outdoor environment.  In \subref{fig:exp-husky-arm-results-outdoor-1} we visualize the hypothesized locations of boxes (in red), each containing a ball, sampled from the world model distribution that our algorithm maintains. The solid green cube denotes the hypothesized box that is the current goal of the planner. The robot then \subref{fig:exp-husky-arm-results-outdoor-2} detects an actual box and looks inside it to find that it does not contain a ball. As \subref{fig:exp-husky-arm-results-outdoor-3} the robot navigates to a hypothesized box, it \subref{fig:exp-husky-arm-results-outdoor-4} detects actual boxes that are found to not contain a ball, while also failing to confirm the presence of hypothesized boxes sampled from the distribution. The algorithm \subref{fig:exp-husky-arm-results-outdoor-5} updates the world model distribution accordingly, and the planner updates the goal. This continues until \subref{fig:exp-husky-arm-results-outdoor-6} the robot observes a box containing a ball and subsequently retrieves the ball, satisfying the instruction.}
  \label{fig:exp-husky-arm-results-outdoor}
\end{figure}
\begin{figure}[!p]
  \centering
  \subfigure[Time $t = 0$ sec]{\includegraphics[width=0.495\linewidth]{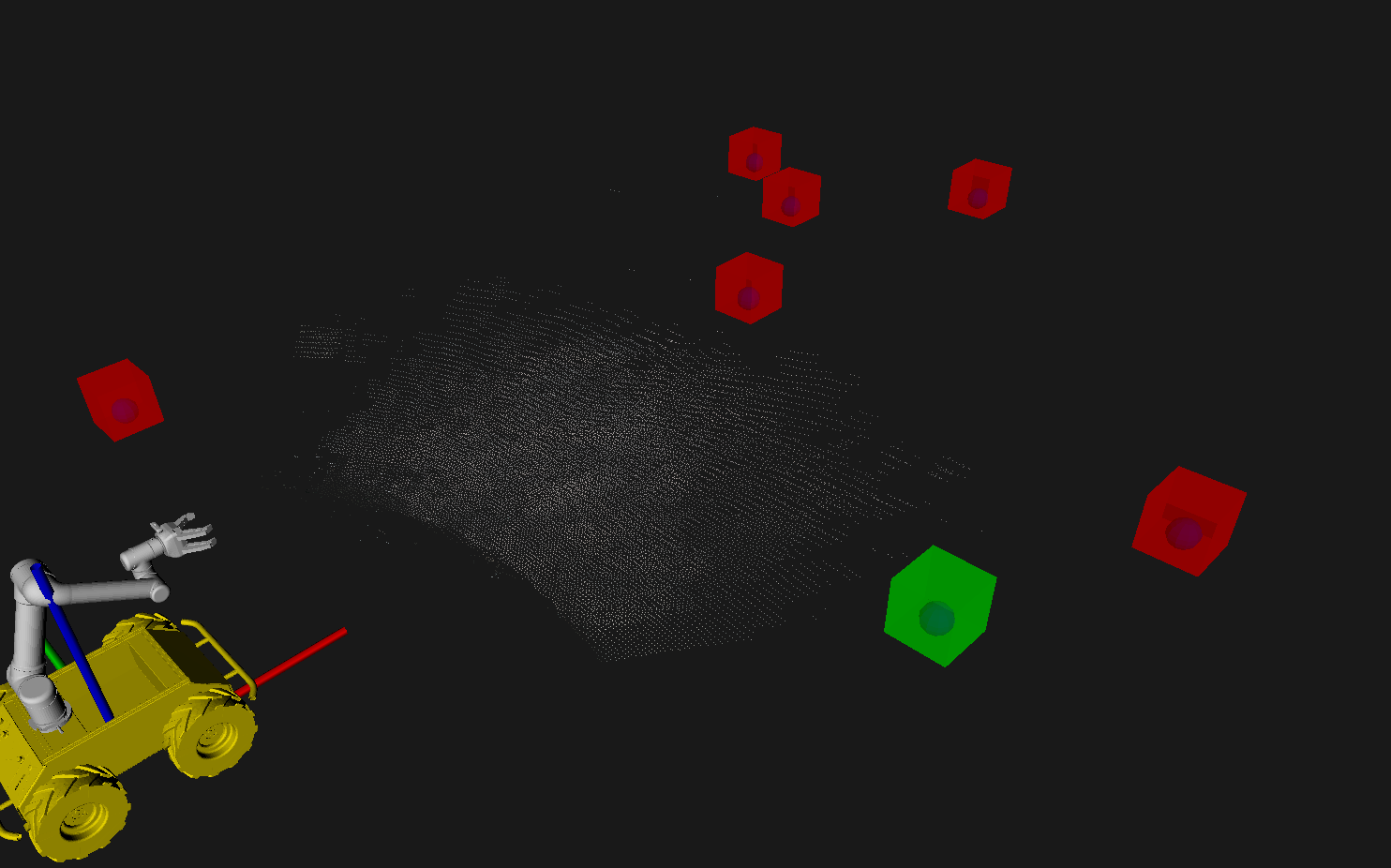}\label{fig:exp-husky-arm-results-indoor-1}}\hfil%
  \subfigure[Time $t = 110$ sec]{\includegraphics[width=0.495\linewidth]{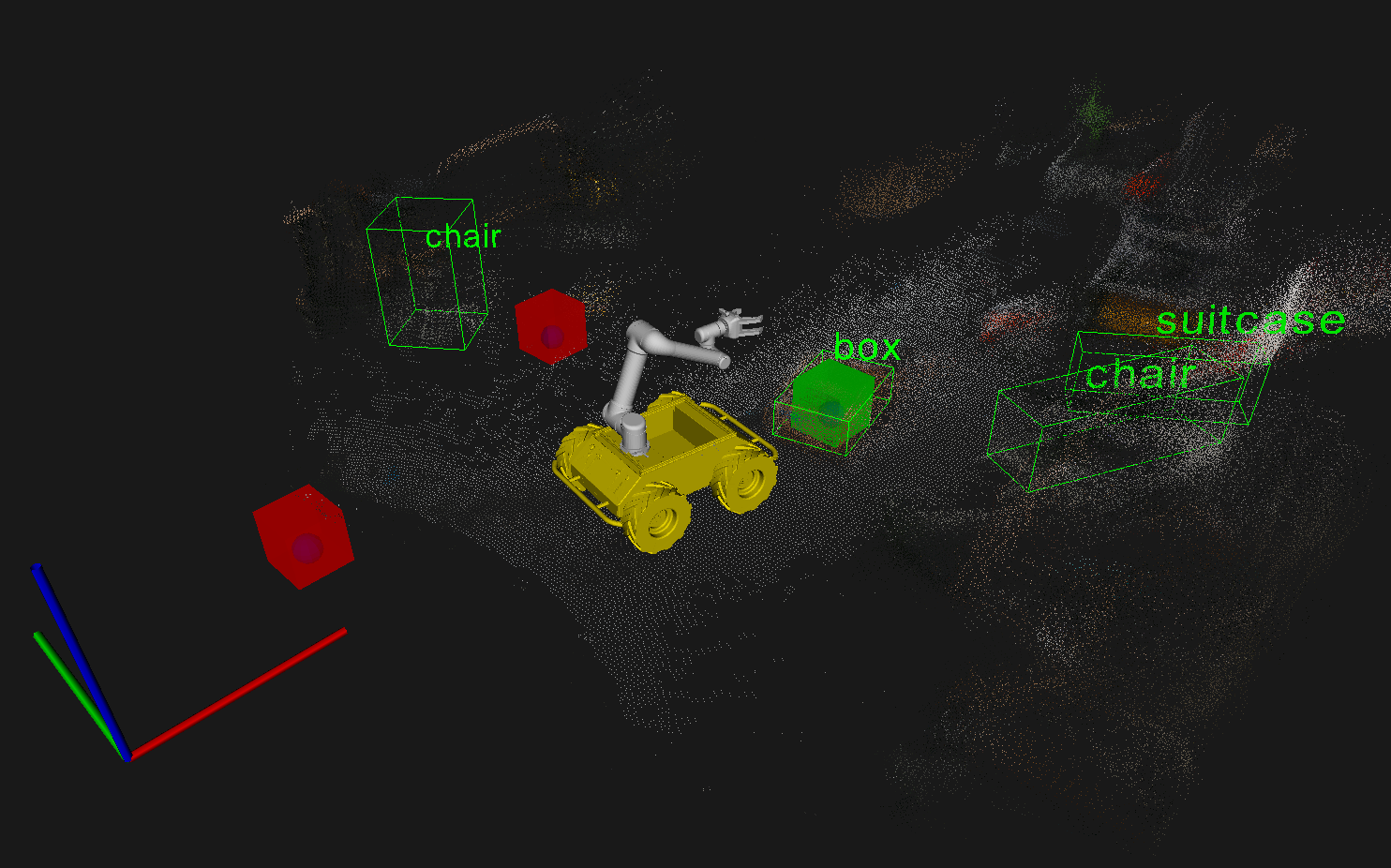}\label{fig:exp-husky-arm-results-indoor-2}}\\%
  \subfigure[Time $t = 141$ sec]{\includegraphics[width=0.495\linewidth]{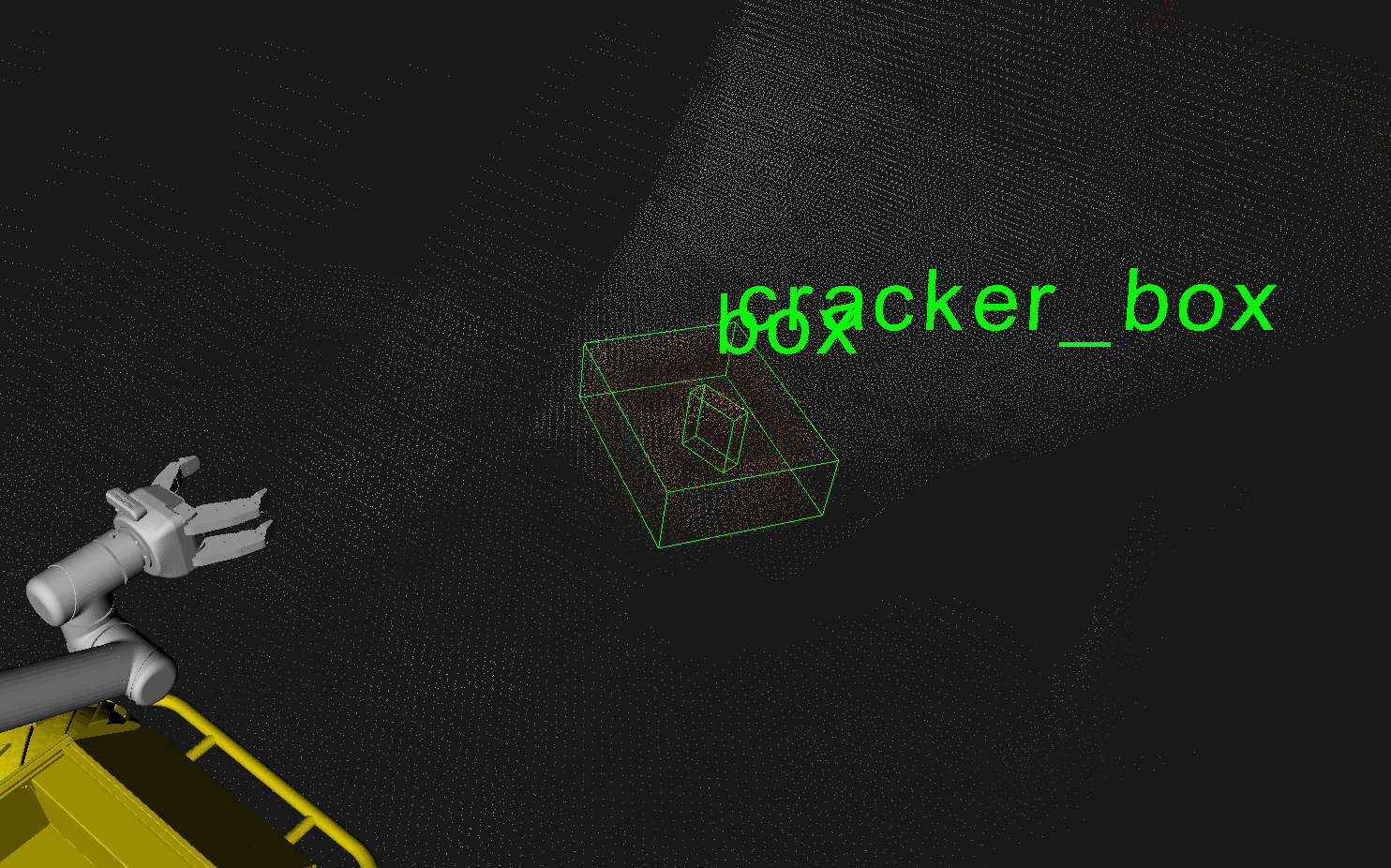}\label{fig:exp-husky-arm-results-indoor-3}}\hfil%
  \subfigure[Time $t = 193$ sec]{\includegraphics[width=0.495\linewidth]{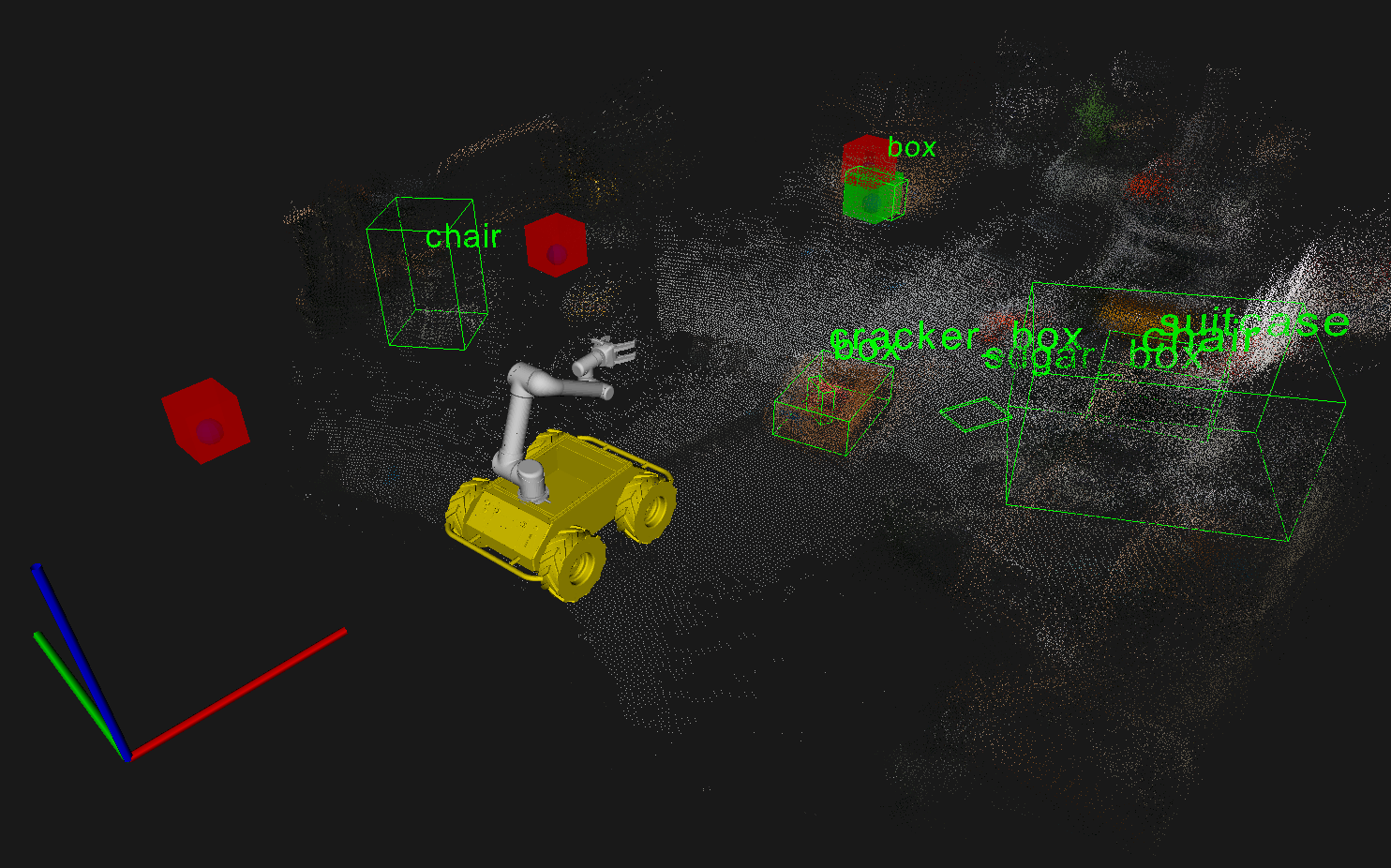}\label{fig:exp-husky-arm-results-indoor-4}}\hfil\\%
  \subfigure[Time $t = 310$ sec]{\includegraphics[width=0.495\linewidth]{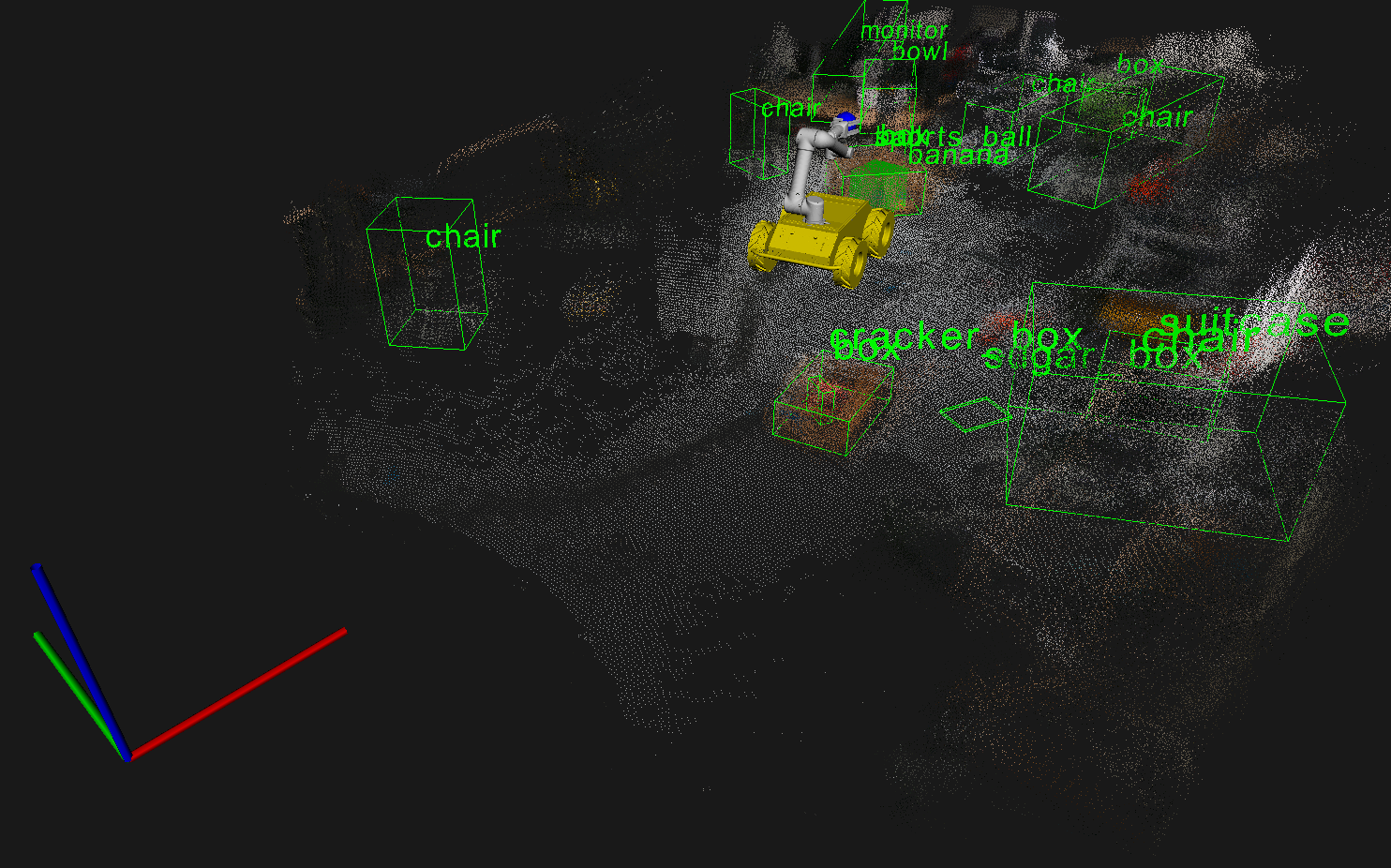}\label{fig:exp-husky-arm-results-indoor-5}}\hfil%
  \subfigure[Time $t = 346$ sec]{\includegraphics[width=0.495\linewidth]{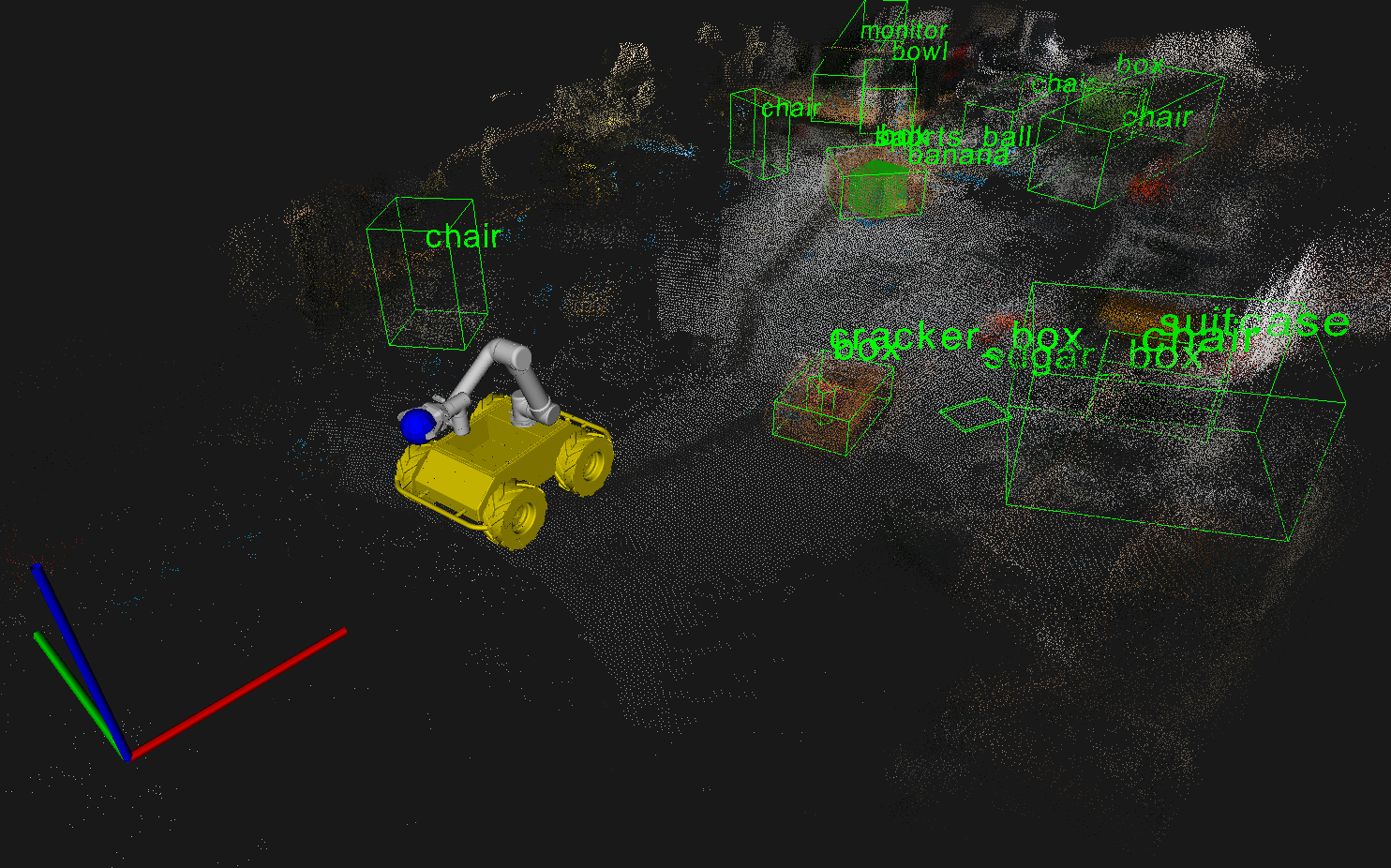}\label{fig:exp-husky-arm-results-indoor-6}}%
  \caption{ Visualization of the state of the robot while executing the instruction ``retrieve the ball inside the box'' in the indoor environment.  In \subref{fig:exp-husky-arm-results-indoor-1} we visualize the hypothesized locations of boxes (in red), each containing a ball, sampled from the world model distribution that our algorithm maintains. The solid green cube denotes the hypothesized box that is the current goal of the planner. The robot then detects an actual box \subref{fig:exp-husky-arm-results-indoor-2} and looks inside it to find \subref{fig:exp-husky-arm-results-indoor-3} that it does not contain a ball, but instead contains a crackers box. The robot \subref{fig:exp-husky-arm-results-indoor-4} eventually detects the second box. The algorithm updates the world model distribution accordingly, and the planner updates the goal. The robot \subref{fig:exp-husky-arm-results-indoor-5} observes a box containing a ball and subsequently \subref{fig:exp-husky-arm-results-indoor-6} retrieves the ball, satisfying the instruction.}
  \label{fig:exp-husky-arm-results-indoor}
\end{figure}

We analyzed the runtime of task execution by measuring the individual runtimes of perception and behavior inference as outlined in Table~\ref{table:exp-husky-arm-results}. Figures~\ref{fig:exp-husky-arm-results-outdoor} and~\ref{fig:exp-husky-arm-results-indoor} shows the state of the robot while executing the instruction.
The robot took approximately six~minutes to execute the first task in the indoor environment, while requiring approximately ten~minutes for the outdoor experiments. Such long task execution runtimes are undesirable for fluent human-robot collaboration.  As observed in Table~\ref{table:exp-husky-arm-results}, a large fraction of the task execution runtime was spent on perception, building a rich model of the robot's environment such as the one in Figure~\ref{fig:exp-husky-arm-results-indoor-6}. Such high fidelity world models are computationally expensive to build and are also unnecessarily detailed for grounding, planning, and executing the instructed task. This raises open questions regarding how to optimally perceive the robot's world for collaborative robots which are required to understand and execute a diverse set of instructions. Furthermore, Figure~\ref{fig:exp-husky-arm-sg-vs-nobj} indicates a positive correlation between the symbol grounding runtime per particle and the number of objects present in the particle (i.e., the fidelity of the world model). This means that maintaining a distribution over highly detailed world models and reasoning in its context is computationally expensive in terms of both perception and symbol grounding. We elaborate on this issue and possible ways to improve scalability below.

\section{Discussion} \label{sec:discussion}

As established through the experimental evaluation in simulation and on actual robots, the proposed model enables natural-language instruction-following in previously unseen or partially observed environments. It provides a guided and efficient exploration mechanism that allows faster runtimes for task execution in previously unseen environments as compared to the baseline of opportunistic exploration. However, we note that the spatial extent of the environments considered in the experimental evaluation is small compared to those of typical field and service robotics settings. While we have previously demonstrated the ability to learn maps of larger, multi-building environments from human-provided descriptions~\citep{walter14}, evaluating the proposed framework at this scale remains as future work.

As the experiments demonstrate, maintaining a distribution of highly-detailed world models and reasoning in its context is computationally expensive in terms of both perception and symbol grounding. Addressing this scalability challenge, a recent line of work~\citep{patki18a, patki2019a, patki2020a} proposes learning to adapt the robot's perception pipeline by exploiting implied utterance information to construct task-relevant world models. These more compact, task-relevant world models afford faster perception and symbol-grounding runtimes, as compared to a baseline configuration that uses a non-adaptive and flat perception pipeline. Recent work~\citep{patki2020a} demonstrates approximately a 50\% reduction in task-execution runtimes in both indoor and outdoor experiments, and illustrates the performance gains that can be achieved by incorporating our proposed adaptive-perception framework. Learning to constrain the robot's perception pipeline adaptively shortens the perception runtime by obviating irrelevant detectors. This reduction in the perception runtime enables our framework to operate more efficiently, while processing the same number of observations, and so reduces overall task-execution time. Furthermore, as behavior inference is performed on each hypothesized world model in the distribution, the efficiency gains provided by adaptive perception enables reasoning over a larger number of environment hypotheses in the same amount of time. This important ability allows maintaining more particles and thus to more efficiently explore previously unseen environments. As part of ongoing work, we are investigating the ability to extend our framework such that it is scalable to larger environments than those considered here.

Another limitation of the current approach is the reliance on fully-annotated data.  Densely labeling examples for both annotation and behavior inference with different symbolic representations is a nontrivial task that involves inferring the presence of relationships needed to perform grounding in a space of hypothesized worlds.  Approaches that are amenable to partially-annotated data, e.g., by using the available annotations to learn how to automatically label the rest of the dataset, would significantly facilitate integrating annotation- and behavior-inference and enable robots to learn models for both processes in situ.  A human operator attempting to demonstrate the concept ``pick up the cup on the table'' would have to group physical objects ``cup'' and ``table,'' along with the spatial relation represented by the phrase ``on'' for behavior inference, and also associate the physical relationship between two hypothesized objects with their associated semantic labels to properly capture this information for annotation inference.   Algorithms capable of this remain an open area of investigation.

Another deficiency of this approach is that the procedures for both annotation and behavior inference does not express bounds on the space of worlds or language wherein the expressed symbols would remain valid.  Uncertainty in automatic speech-recognition and parsing could influence the quality of annotation inference, and behavior inference could further be impacted by noise in perception.  More efficient and effective methods for evaluating confidence in an expressed set of symbols may influence whether the robot should exploit these annotations or engage in dialogue with a human to confirm that such information should be added to the environment model.

\section{Conclusions} \label{sec:conclusions}
Significant progress in grounded natural-language understanding has enabled robots to interpret a diverse array of free-form navigation, manipulation, and mobile-manipulation commands. However, most contemporary approaches require a pre-existing, detailed spatial-semantic map of the robot's environment that represents the objects or regions that the utterance may reference. Consequently, these methods fail when robots are deployed in previously unseen or partially-observed environments, particularly when mental models of the environment differ between the human user and robot. This paper describes a learning framework that allows field and service robots to execute natural-language instructions in previously unseen or partially-observed environments. The experimental results in simulation and on three different robotic platforms indicate that the proposed model facilitates faster task execution in previously unseen environments as compared to a contemporary language-grounding baseline that does not take advantage of environment information available in the instruction. The results also show that the method is amenable to tasks that include navigation and mobile manipulation. Importantly, experimental data also reveal limitations of the approach, including the challenge of maintaining a distribution over highly-detailed world models and reasoning in its context, which is computationally expensive in terms of both perception and symbol grounding. This observation raises interesting, open-ended questions that are the focus on ongoing research, such as how to represent complex environments efficiently while also supporting a diverse array of tasks in large-scale environments.

\section{Acknowledgements} \label{sec:acknowledgements}

This work was supported in part by the National Science Foundation under grants IIS-1638072 and IIS-1637813, by the Robotics Consortium of the U.S. Army Research Laboratory under the Collaborative Technology Alliance Program Cooperative Agreement W911NF-10-2-0016, and by ARO grants W911NF-15-1-0402 and W911NF-17-1-0188.

\bibliographystyle{apalike}
\bibliography{references}

\begin{thebibliography}{}

\bibitem[Abbeel and Ng, 2004]{abbeel04}
Abbeel, P. and Ng, A.~Y. (2004).
\newblock Apprenticeship learning via inverse reinforcement learning.
\newblock In {\em Proceedings of the International Conference on Machine
  Learning (ICML)}.

\bibitem[Anderson et~al., 2018]{anderson18}
Anderson, P., Wu, Q., Teney, D., Bruce, J., Johnson, M., S{\"u}nderhauf, N.,
  Reid, I., Gould, S., and van~den Hengel, A. (2018).
\newblock Vision-and-language navigation: {I}nterpreting visually-grounded
  navigation instructions in real environments.
\newblock In {\em Proceedings of the IEEE Conference on Computer Vision and
  Pattern Recognition (CVPR)}, pages 3674--3683.

\bibitem[Arkin et~al., 2020]{arkin20a}
Arkin, J., Park, D., Roy, S., Walter, M.~R., Roy, N., Howard, T.~M., and Paul,
  R. (2020).
\newblock Multimodal estimation and communication of latent semantic knowledge
  for robust execution of robot instructions.
\newblock {\em International Journal of Robotics Research},
  39(10--11):1279--1304.

\bibitem[Arkin et~al., 2018]{arkin2018iser}
Arkin, J., Paul, R., Park, D., Roy, S., Roy, N., and Howard, T.~M. (2018).
\newblock Real-time human-robot communication for manipulation tasks in
  partially observed environments.
\newblock In {\em Proceedings of the International Symposium on Experimental
  Robotics (ISER)}, pages 448--460.

\bibitem[Arvidson et~al., 2010]{arvidson2010spirit}
Arvidson, R.~E., Bell, J., Bellutta, P., Cabrol, N.~A., Catalano, J., Cohen,
  J., Crumpler, L.~S., Des~Marais, D., Estlin, T., Farrand, W., et~al. (2010).
\newblock Spirit {M}ars {R}over {M}ission: {O}verview and selected results from
  the northern {H}ome {P}late {W}inter {H}aven to the side of {S}camander
  crater.
\newblock {\em Journal of Geophysical Research: Planets}, 115(E7).

\bibitem[Aydemir et~al., 2013]{aydemir13}
Aydemir, A., Pronobis, A., G{\"o}belbecker, M., and Jensfelt, P. (2013).
\newblock Active visual object search in unknown environments using uncertain
  semantics.
\newblock {\em IEEE Transactions on Robotics}, 29(4):986--1002.

\bibitem[Aydemir et~al., 2011]{aydemir11}
Aydemir, A., Sj{\"o}{\"o}, K., Folkesson, J., Pronobis, A., and Jensfelt, P.
  (2011).
\newblock Search in the real world: {A}ctive visual object search based on
  spatial relations.
\newblock In {\em Proceedings of the IEEE International Conference on Robotics
  and Automation (ICRA)}, pages 2818--2824.

\bibitem[Bacha et~al., 2008]{bacha08a}
Bacha, A., Bauman, C., Faruque, R., Fleming, M., Terwelp, C., Reinholtz, C.,
  Hong, D., Wicks, A., Alberi, T., Anderson, D., Cacciola, S., Currier, P.,
  Dalton, A., Farmer, J., Hurdus, J., Kimmel, S., King, P., Taylor, A., Covern,
  D.~V., and Webster, M. (2008).
\newblock Odin: {T}eam {VictorTango}'s entry in the {DARPA} {U}rban
  {C}hallenge.
\newblock {\em Journal of Field Robotics}, 25(8):467--492.

\bibitem[Bansal et~al., 2019]{bansal19}
Bansal, S., Tolani, V., Gupta, S., Malik, J., and Tomlin, C. (2019).
\newblock Combining optimal control and learning for visual navigation in novel
  environments.
\newblock In {\em Proceedings of the Conference on Robot Learning (CoRL)},
  pages 420--429.

\bibitem[Barber et~al., 2016]{barber16a}
Barber, D., Howard, T.~M., and Walter, M. (2016).
\newblock A multimodal interface for real-time soldier-robot teaming.
\newblock In {\em Unmanned Systems Technology XVIII}, volume 9837.
  International Society for Optics and Photonics.

\bibitem[Beeson and Ames, 2015]{Beeson-humanoids-15}
Beeson, P. and Ames, B. (2015).
\newblock {TRAC-IK}: {A}n open-source library for improved solving of generic
  inverse kinematics.
\newblock In {\em Proceedings of the IEEE-RAS International Conference on
  Humanoid Robots (Humanoids)}, pages 928--935.

\bibitem[Blanco et~al., 2006]{blanco06}
Blanco, J.-L., Gonzalez, J., and Fernandez-Madrigal, J. (2006).
\newblock Consistent observation grouping for generating metric-topological
  maps that improves robot localization.
\newblock In {\em Proceedings of the IEEE International Conference on Robotics
  and Automation (ICRA)}, pages 818--823.

\bibitem[Bohren et~al., 2008]{bohren08a}
Bohren, J., Foote, T., Keller, J., Kushleyev, A., Lee, D., Stewart, A.,
  Vernaza, P., Derenick, J., Spletzer, J., and Satterfield, B. (2008).
\newblock Little {B}en: {T}he {B}en {F}ranklin {R}acing {T}eam's entry in the
  2007 {DARPA} {U}rban {C}hallenge.
\newblock {\em Journal of Field Robotics}, 25(9):598--614.

\bibitem[Bollini et~al., 2010]{bollini10}
Bollini, M., Tellex, S., Thompson, T., Roy, N., and Rus, D. (2010).
\newblock Interpreting and executing recipes with a cooking robot.
\newblock In {\em Proceedings of the International Symposium on Experimental
  Robotics (ISER)}, pages 481--495.

\bibitem[Boteanu et~al., 2016]{boteanu16a}
Boteanu, A., Arkin, J., Howard, T.~M., and Kress-Gazit, H. (2016).
\newblock A model for verifiable grounding and execution of complex language
  instructions.
\newblock In {\em 2016 IEEE/RSJ International Conference on Intelligent Robots
  and Systems}, pages 2649--2654.

\bibitem[Bowen et~al., 2008]{bowen2008nereus}
Bowen, A.~D., Yoerger, D.~R., Taylor, C., McCabe, R., Howland, J.,
  Gomez-Ibanez, D., Kinsey, J.~C., Heintz, M., McDonald, G., Peters, D.~B.,
  et~al. (2008).
\newblock The {N}ereus hybrid underwater robotic vehicle for global ocean
  science operations to 11,000 m depth.
\newblock In {\em Proceedings of the IEEE/MTS OCEANS Conference and
  Exhibition}.

\bibitem[Broad et~al., 2017]{broad17a}
Broad, A., Arkin, J., Ratliff, N., Howard, T.~M., and Argall, B. (2017).
\newblock Real-time natural language corrections for assistive robotic
  manipulators.
\newblock {\em International Journal of Robotics Research}, 36(5-7):684--698.

\bibitem[Calli et~al., 2015]{calli2015benchmarking}
Calli, B., Walsman, A., Singh, A., Srinivasa, S., Abbeel, P., and Dollar, A.~M.
  (2015).
\newblock Benchmarking in manipulation research: {T}he {YCB} object and model
  set and benchmarking protocols.
\newblock {\em IEEE Robotics and Automation Magazine}, pages 36--52.

\bibitem[Camilli et~al., 2010]{camilli2010tracking}
Camilli, R., Reddy, C.~M., Yoerger, D.~R., Van~Mooy, B.~A., Jakuba, M.~V.,
  Kinsey, J.~C., McIntyre, C.~P., Sylva, S.~P., and Maloney, J.~V. (2010).
\newblock Tracking hydrocarbon plume transport and biodegradation at
  {D}eepwater {H}orizon.
\newblock {\em Science}, 330(6001):201--204.

\bibitem[Chen and Mooney, 2011]{chen11}
Chen, D.~L. and Mooney, R.~J. (2011).
\newblock Learning to interpret natural language navigation instructions from
  observations.
\newblock In {\em Proceedings of the National Conference on Artificial
  Intelligence (AAAI)}, pages 859--865.

\bibitem[Chiang et~al., 2019]{chiang19}
Chiang, H.-T.~L., Faust, A., Fiser, M., and Francis, A. (2019).
\newblock Learning navigation behaviors end-to-end with {AutoRL}.
\newblock {\em IEEE Robotics and Automation Letters}, 4(2):2007--2014.

\bibitem[Chung et~al., 2015]{chung15}
Chung, I., Propp, O., Walter, M., and Howard, T. (2015).
\newblock On the performance of hierarchical distributed correspondence graphs
  for efficient symbol grounding of robot instructions.
\newblock In {\em Proceedings of the IEEE/RSJ International Conference on
  Intelligent Robots and Systems (IROS)}, pages 5247--5252.

\bibitem[Crammer and Singer, 2002]{crammer02}
Crammer, K. and Singer, Y. (2002).
\newblock On the algorithmic implementation of multiclass kernel-based vector
  machines.
\newblock {\em Journal of Machine Learning Research}, 2:265--292.

\bibitem[Daniele et~al., 2017]{daniele17}
Daniele, A.~F., Bansal, M., and Walter, M.~R. (2017).
\newblock Navigational instruction generation as inverse reinforcement learning
  with neural machine translation.
\newblock In {\em Proceedings of the ACM/IEEE International Conference on
  Human-Robot Interaction (HRI)}, pages 109--118.

\bibitem[Doucet, 1998]{doucet98}
Doucet, A. (1998).
\newblock On sequential simulation-based methods for bayesian filtering.
\newblock Technical Report CUED/F-INFENG/TR 310, Department of Engineering,
  University of Cambridge.

\bibitem[Doucet et~al., 2000]{doucet00}
Doucet, A., de~Freitas, N., Murphy, K., and Russell, S. (2000).
\newblock Rao-{B}lackwellised particle filtering for dynamic {Bayesian}
  networks.
\newblock In {\em Proceedings of the Conference on Uncertainty in Artificial
  Intelligence (UAI)}, pages 176--183.

\bibitem[Duff et~al., 2003]{duff2003automation}
Duff, E.~S., Roberts, J.~M., and Corke, P.~I. (2003).
\newblock Automation of an underground mining vehicle using reactive navigation
  and opportunistic localization.
\newblock In {\em Proceedings of the IEEE/RSJ International Conference on
  Intelligent Robots and Systems (IROS)}, pages 3775--3780.

\bibitem[Durrant-Whyte et~al., 2007]{durrant-whyte07}
Durrant-Whyte, H., Pagac, D., Rogers, B., Stevens, M., and Nelmes, G. (2007).
\newblock Field and service applications---an autonomous straddle carrier for
  movement of shipping containers---from research to operational autonomous
  systems.
\newblock {\em IEEE Robotics \& Automation Magazine}, 14(3):14--23.

\bibitem[Durrant-Whyte, 1996]{durrant1996autonomous}
Durrant-Whyte, H.~F. (1996).
\newblock An autonomous guided vehicle for cargo handling applications.
\newblock {\em International Journal of Robotics Research}, 15(5):407--440.

\bibitem[Duvallet et~al., 2013]{duvallet13}
Duvallet, F., Kollar, T., and Stentz, A. (2013).
\newblock Imitation learning for natural language direction following through
  unknown environments.
\newblock In {\em Proceedings of the IEEE International Conference on Robotics
  and Automation (ICRA)}, pages 1047--1053.

\bibitem[Duvallet et~al., 2014]{duvallet14}
Duvallet, F., Walter, M.~R., Howard, T., Hemachandra, S., Oh, J., Teller, S.,
  Roy, N., and Stentz, A. (2014).
\newblock Inferring maps and behaviors from natural language instructions.
\newblock In {\em Proceedings of the International Symposium on Experimental
  Robotics (ISER)}.

\bibitem[Eustice et~al., 2005]{eustice05}
Eustice, R., Singh, H., and Leonard, J. (2005).
\newblock Exactly sparse delayed-state filters.
\newblock In {\em Proceedings of the IEEE International Conference on Robotics
  and Automation (ICRA)}, pages 2417--2424.

\bibitem[Furgale and Barfoot, 2010]{furgale2010visual}
Furgale, P. and Barfoot, T.~D. (2010).
\newblock Visual teach and repeat for long-range rover autonomy.
\newblock {\em Journal of Field Robotics}, 27(5):534--560.

\bibitem[Galindo et~al., 2005]{galindo05}
Galindo, C., Saffiotti, A., Coradeschi, S., Buschka, P., Fernandez-Madrigal,
  J., and Gonzalez, J. (2005).
\newblock Multi-hierarchical semantic maps for mobile robotics.
\newblock In {\em Proceedings of the IEEE/RSJ International Conference on
  Intelligent Robots and Systems (IROS)}, pages 2278--2283.

\bibitem[German et~al., 2008]{german2008hydrothermal}
German, C.~R., Yoerger, D.~R., Jakuba, M., Shank, T.~M., Langmuir, C.~H., and
  Nakamura, K.-i. (2008).
\newblock Hydrothermal exploration with the {A}utonomous {B}enthic {E}xplorer.
\newblock {\em Deep Sea Research Part I: Oceanographic Research Papers},
  55(2):203--219.

\bibitem[Gong and Zhang, 2018]{gong18}
Gong, Z. and Zhang, Y. (2018).
\newblock Temporal spatial inverse semantics for robots communicating with
  humans.
\newblock In {\em Proc.\ IEEE Int'l Conf.\ on Robotics and Automation (ICRA)}.

\bibitem[Gretton et~al., 2007]{gretton07}
Gretton, A., Borgwardt, K., Rasch, M., Sch{\"o}lkopf, B., and Smola, A.~J.
  (2007).
\newblock A kernel method for the two-sample-problem.
\newblock In {\em Advances in Neural Information Processing Systems (NeurIPS)},
  pages 513--520.

\bibitem[Gupta et~al., 2017]{gupta17}
Gupta, S., Davidson, J., Levine, S., Sukthankar, R., and Malik, J. (2017).
\newblock Cognitive mapping and planning for visual navigation.
\newblock In {\em Proceedings of the IEEE Conference on Computer Vision and
  Pattern Recognition (CVPR)}, pages 2616--2625.

\bibitem[Harnad, 1990]{harnad90}
Harnad, S. (1990).
\newblock The symbol grounding problem.
\newblock {\em Physica D}, 42:335--346.

\bibitem[Hemachandra et~al., 2015]{hemachandra15}
Hemachandra, S., Duvallet, F., Howard, T.~M., Roy, N., Stentz, A., and Walter,
  M.~R. (2015).
\newblock Learning models for following natural language directions in unknown
  environments.
\newblock In {\em Proceedings of the IEEE International Conference on Robotics
  and Automation (ICRA)}, pages 5608--5615.

\bibitem[Hemachandra et~al., 2011]{hemachandra11}
Hemachandra, S., Kollar, T., Roy, N., and Teller, S. (2011).
\newblock Following and interpreting narrated guided tours.
\newblock In {\em Proceedings of the IEEE International Conference on Robotics
  and Automation (ICRA)}, pages 2574--2579.

\bibitem[Hemachandra et~al., 2014]{hemachandra14}
Hemachandra, S., Walter, M.~R., Tellex, S., and Teller, S. (2014).
\newblock Learning spatial-semantic representations from natural language
  descriptions and scene classifications.
\newblock In {\em Proceedings of the IEEE International Conference on Robotics
  and Automation (ICRA)}, pages 2623--2630.

\bibitem[Hetherington, 2007]{hetherington07}
Hetherington, I.~L. (2007).
\newblock {PocketSUMMIT}: Small-footprint continuous speech recognition.
\newblock In {\em Proceedings of Annual Conference of the International Speech
  Communication Association (INTERSPEECH)}, pages 1465--1468.

\bibitem[Howard et~al., 2014a]{howard14b}
Howard, T., Chung, I., Propp, O., Walter, M., and Roy, N. (2014a).
\newblock Efficient natural language interfaces for assistive robots.
\newblock In {\em Proceedings of the IEEE/RSJ International Conference on
  Intelligent Robots and Systems (IROS) Workshop on Rehabilitation and
  Assistive Robotics}.

\bibitem[Howard et~al., 2014b]{howard14}
Howard, T.~M., Tellex, S., and Roy, N. (2014b).
\newblock A natural language planner interface for mobile manipulators.
\newblock In {\em Proceedings of the IEEE International Conference on Robotics
  and Automation (ICRA)}, pages 6652--6659.

\bibitem[Johnson-Roberson et~al., 2010]{johnson2010generation}
Johnson-Roberson, M., Pizarro, O., Williams, S.~B., and Mahon, I. (2010).
\newblock Generation and visualization of large-scale three-dimensional
  reconstructions from underwater robotic surveys.
\newblock {\em Journal of Field Robotics}, 27(1):21--51.

\bibitem[Kaess et~al., 2008]{kaess08}
Kaess, M., Ranganathan, A., and Dellaert, F. (2008).
\newblock {iSAM}: {I}ncremental smoothing and mapping.
\newblock {\em Transactions on Robotics}, 24(6):1365--1378.

\bibitem[Kaess et~al., 2009]{kaess08-software}
Kaess, M., Ranganathan, A., and Dellaert, F. (2009).
\newblock {iSAM}: {I}ncremental smoothing and mapping.
\newblock \url{http://people.csail.mit.edu/kaess/isam/}.

\bibitem[Kang et~al., 2003]{kang2003robhaz}
Kang, S., Cho, C., Lee, J., Ryu, D., Park, C., Shin, K.-C., and Kim, M. (2003).
\newblock {ROBHAZ-DT2}: {D}esign and integration of passive double tracked
  mobile manipulator system for explosive ordnance disposal.
\newblock In {\em Proceedings of the IEEE/RSJ International Conference on
  Intelligent Robots and Systems (IROS)}, pages 2624--2629.

\bibitem[Karaman et~al., 2011]{karaman11}
Karaman, S., Walter, M.~R., Perez, A., Frazzoli, E., and Teller, S. (2011).
\newblock Anytime motion planning using the {RRT}$^*$.
\newblock In {\em Proceedings of the IEEE International Conference on Robotics
  and Automation (ICRA)}, pages 1478--1483.

\bibitem[Keiji et~al., 2011]{keiji2011redesign}
Keiji, N., Swiga, K., Yoshito, O., et~al. (2011).
\newblock Redesign of rescue mobile robot {Q}uince---{T}oward emergency
  response to the nuclear accident at {F}ukushima {D}aiichi {N}uclear {P}ower
  {S}tation on {M}arch 2011.
\newblock In {\em Proceedings of the IEEE International Symposium on Safety,
  Security, and Rescue Robotics (SSRR)}, pages 13--18.

\bibitem[Kim and Chen, 2015]{kim15}
Kim, D.~K. and Chen, T. (2015).
\newblock Deep neural network for real-time autonomous indoor navigation.
\newblock {\em arXiv preprint arXiv:1511.04668}.

\bibitem[Kollar and Roy, 2009]{kollar09}
Kollar, T. and Roy, N. (2009).
\newblock Utilizing object-object and object-scene context when planning to
  find things.
\newblock In {\em Proceedings of the IEEE International Conference on Robotics
  and Automation (ICRA)}, pages 4116--4121.

\bibitem[Kollar et~al., 2010]{kollar10}
Kollar, T., Tellex, S., Roy, D., and Roy, N. (2010).
\newblock Toward understanding natural language directions.
\newblock In {\em Proceedings of the ACM/IEEE International Conference on
  Human-Robot Interaction (HRI)}, pages 259--266.

\bibitem[Krieg-Br{\"u}ckher et~al., 2005]{krieg05}
Krieg-Br{\"u}ckher, B., Frese, U., L{\"u}ttich, K., Mandel, C., Massakowski,
  T., and Ross, R.~J. (2005).
\newblock Specification of an ontology for route graphs.
\newblock {\em Spatial Cognition IV: Reasoning, Action, Interaction},
  3343:390--412.

\bibitem[Kuznetsova et~al., 2018]{OpenImages}
Kuznetsova, A., Rom, H., Alldrin, N., Uijlings, J., Krasin, I., Pont-Tuset, J.,
  Kamali, S., Popov, S., Malloci, M., Duerig, T., and Ferrari, V. (2018).
\newblock The {O}pen {I}mages {D}ataset {V}4: {U}nified image classification,
  object detection, and visual relationship detection at scale.
\newblock {\em arXiv:1811.00982}.

\bibitem[Landsiedel et~al., 2017]{landsiedel17}
Landsiedel, C., Rieser, V., Walter, M.~R., and Wollherr, D. (2017).
\newblock A review of spatial reasoning and interaction for real-world
  robotics.
\newblock {\em Advanced Robotics}, 31(5):222--242.

\bibitem[Leonard et~al., 2008]{leonard08a}
Leonard, J., How, J., Teller, S., Berger, M., Campbell, S., Fiore, G.,
  Fletcher, L., Frazzoli, E., Huang, A., Karaman, S., Koch, O., Kuwata, Y.,
  Moore, D., Olson, E., Peters, S., Teo, J., Truax, R., Walter, M., Barrett,
  D., Epstein, A., Maheloni, K., Moyer, K., Jones, T., Buckley, R., Antone, M.,
  Galejs, R., Krishnamurthy, S., and Williams, J. (2008).
\newblock A perception-driven autonomous urban vehicle.
\newblock {\em Journal of Field Robotics}, 25(10):727--774.

\bibitem[Lin et~al., 2014]{lin2014microsoft}
Lin, T.-Y., Maire, M., Belongie, S., Hays, J., Perona, P., Ramanan, D.,
  Doll{\'a}r, P., and Zitnick, C.~L. (2014).
\newblock Microsoft {COCO}: Common objects in context.
\newblock In {\em Proceedings of the European Conference on Computer Vision
  (ECCV)}, pages 740--755.

\bibitem[MacMahon et~al., 2006]{macmahon06}
MacMahon, M., Stankiewicz, B., and Kuipers, B. (2006).
\newblock Walk the talk: {C}onnecting language, knowledge, and action in route
  instructions.
\newblock In {\em Proceedings of the National Conference on Artificial
  Intelligence (AAAI)}, pages 1475--1482.

\bibitem[Maimone et~al., 2007]{maimone2007two}
Maimone, M., Cheng, Y., and Matthies, L. (2007).
\newblock Two years of visual odometry on the {M}ars exploration rovers.
\newblock {\em Journal of Field Robotics}, 24(3):169--186.

\bibitem[Marshall et~al., 2008]{marshall2008autonomous}
Marshall, J., Barfoot, T., and Larsson, J. (2008).
\newblock Autonomous underground tramming for center-articulated vehicles.
\newblock {\em Journal of Field Robotics}, 25(6-7):400--421.

\bibitem[Mart{\'i}nez~Mozos et~al., 2007]{mozos07}
Mart{\'i}nez~Mozos, O., Triebel, R., Jensfelt, P., Rottmann, A., and Burgard,
  W. (2007).
\newblock Supervised semantic labeling of places using information extracted
  from sensor data.
\newblock {\em Robotics and Autonomous Systems}, 55(5):391--402.

\bibitem[Matuszek et~al., 2010]{matuszek10}
Matuszek, C., Fox, D., and Koscher, K. (2010).
\newblock Following directions using statistical machine translation.
\newblock In {\em Proceedings of the ACM/IEEE International Conference on
  Human-Robot Interaction (HRI)}, pages 251--258.

\bibitem[Matuszek et~al., 2012]{matuszek12a}
Matuszek, C., Herbst, E., Zettlemoyer, L., and Fox, D. (2012).
\newblock Learning to parse natural language commands to a robot control
  system.
\newblock In {\em Proceedings of the International Symposium on Experimental
  Robotics (ISER)}, pages 403--415.

\bibitem[Meger et~al., 2008]{meger08}
Meger, D., Forss{\'e}n, P.-E., Lai, K., Helmer, S., McCann, S., Southey, T.,
  Baumann, M., Little, J.~J., and Lowe, D.~G. (2008).
\newblock Curious {G}eorge: {A}n attentive semantic robot.
\newblock {\em Robotics and Autonomous Systems}, 56(6):503--511.

\bibitem[Mei et~al., 2016]{mei16}
Mei, H., Bansal, M., and Walter, M.~R. (2016).
\newblock Listen, attend, and walk: Neural mapping of navigational instructions
  to action sequences.
\newblock In {\em Proceedings of the National Conference on Artificial
  Intelligence (AAAI)}.

\bibitem[Miller et~al., 2008]{miller08a}
Miller, I., Campbell, M., Huttenlocher, D., Kline, F.-R., Nathan, A., Lupashin,
  S., Catlin, J., Schimpf, B., Moran, P., Zych, N., Garcia, E., Kurdziel, M.,
  and Fujishima, H. (2008).
\newblock Team {C}ornell's {S}kynet: {R}obust perception and planning in an
  urban environment.
\newblock {\em Journal of Field Robotics}, 25(8):493--527.

\bibitem[Mirowski et~al., 2018]{mirowski18}
Mirowski, P., Grimes, M., Malinowski, M., Hermann, K.~M., Anderson, K.,
  Teplyashin, D., Simonyan, K., Zisserman, A., and Hadsell, R. (2018).
\newblock Learning to navigate in cities without a map.
\newblock In {\em Advances in Neural Information Processing Systems (NeurIPS)},
  pages 2419--2430.

\bibitem[Misra et~al., 2016]{misra16}
Misra, D.~K., Sung, J., Lee, K., and Saxena, A. (2016).
\newblock Tell me {D}ave: Context-sensitive grounding of natural language to
  manipulation instructions.
\newblock {\em International Journal of Robotics Research}, 35(1-3):281--300.

\bibitem[Montemerlo et~al., 2008]{montemerlo08a}
Montemerlo, M., Becker, J., Bhat, S., Dahlkamp, H., Dolgov, D., Ettinger, S.,
  Haehnel, D., Hilden, T., Hoffmann, G., Huhnke, B., Johnston, D., Klumpp, S.,
  Langer, D., Levandowski, A., Levinson, J., Marcil, J., Orenstein, D.,
  Paefgen, J., Penny, I., Petrovskaya, A., Pflueger, M., Stanek, G., Stavens,
  D., Vogt, A., and Thrun, S. (2008).
\newblock Junior: {T}he {S}tanford entry in the {U}rban {C}hallenge.
\newblock {\em Journal of Field Robotics}, 25(9):569--597.

\bibitem[Nagatani et~al., 2013]{nagatani2013emergency}
Nagatani, K., Kiribayashi, S., Okada, Y., Otake, K., Yoshida, K., Tadokoro, S.,
  Nishimura, T., Yoshida, T., Koyanagi, E., Fukushima, M., et~al. (2013).
\newblock Emergency response to the nuclear accident at the {F}ukushima
  {D}aiichi {N}uclear {P}ower {P}lants using mobile rescue robots.
\newblock {\em Journal of Field Robotics}, 30(1):44--63.

\bibitem[Nagatani et~al., 2008]{nagatani2008development}
Nagatani, K., Yoshida, K., Kiyokawa, K., Yagi, Y., Adachi, T., Saitoh, H.,
  Suzuki, T., and Takizawa, O. (2008).
\newblock Development of a networked robotic system for disaster mitigation.
\newblock In {\em Proceedings of the International Conference on Field and
  Service Robotics (FSR)}, pages 453--462.

\bibitem[N{\"u}chter et~al., 2003]{nuchter03}
N{\"u}chter, A., Surmann, H., Lingemann, K., and Hertzberg, J. (2003).
\newblock Semantic scene analysis fo scanned {3D} indoor environments.
\newblock In {\em Proceedings of the International Workshop on Vision, Modeling
  and Visualization (VMV)}, pages 215--221.

\bibitem[Oh et~al., 2017]{oh17a}
Oh, J., Howard, T.~M., Walter, M., Barber, D., Zhu, M., Park, S., Suppe, A.,
  Navarro-Serment, L., Duvallet, F., Boularias, A., Romero, O., Vinokrov, J.,
  Keegan, T., Dean, R., Lennon, C., Bodt, B., Childers, M., Shi, J.,
  Daniilidis, K., Roy, N., Lebiere, C., Hebert, M., and Stentz, A. (2017).
\newblock Integrated intelligence for human-robot teams.
\newblock In {\em Proceedings of the International Symposium on Experimental
  Robotics (ISER)}, pages 309--322.

\bibitem[Olson, 2011]{olson2011apriltag}
Olson, E. (2011).
\newblock {AprilTag}: {A} robust and flexible visual fiducial system.
\newblock In {\em Proceedings of the IEEE International Conference on Robotics
  and Automation (ICRA)}, pages 3400--3407.

\bibitem[Olson et~al., 2006]{olson06}
Olson, E., Leonard, J., and Teller, S. (2006).
\newblock Fast iterative optimization of pose graphs with poor initial
  estimates.
\newblock In {\em Proceedings of the IEEE International Conference on Robotics
  and Automation (ICRA)}, pages 2262--2269.

\bibitem[Paskin, 2003]{paskin03a}
Paskin, M.~A. (2003).
\newblock Thin junction tree filters for simultaneous localization and mapping.
\newblock In {\em Proceedings of the International Joint Conference on
  Artificial Intelligence (IJCAI)}, pages 1157--1164.

\bibitem[Patki et~al., 2019]{patki2019a}
Patki, S., Daniele, A., Walter, M., and Howard, T. (2019).
\newblock Inferring compact representations for efficient natural language
  understanding of robot instructions.
\newblock In {\em Proceedings of the IEEE International Conference on Robotics
  and Automation (ICRA)}, pages 6926--6933.

\bibitem[Patki et~al., 2020]{patki2020a}
Patki, S., Fahnestock, E., Howard, T.~M., and Walter, M.~R. (2020).
\newblock Language-guided semantic mapping and mobile manipulation in partially
  observable environments.
\newblock In {\em Proceedings of the Conference on Robot Learning (CoRL)},
  pages 1201--1210.

\bibitem[Patki and Howard, 2018]{patki18a}
Patki, S. and Howard, T.~M. (2018).
\newblock Language-guided adaptive perception for efficient grounded
  communication with robotic manipulators in cluttered environments.
\newblock In {\em Proceedings of the Annual Meeting of the Special Interest
  Group on Discourse and Dialogue (SIGDIAL)}, pages 151--160.

\bibitem[Paul et~al., 2018]{paul18}
Paul, R., Arkin, J., Aksaray, D., Roy, N., and Howard, T.~M. (2018).
\newblock Efficient grounding of abstract spatial concepts for natural language
  interaction with robot platforms.
\newblock {\em International Journal of Robotics Research}, 37(10):1269--1299.

\bibitem[Paul et~al., 2016]{paul16a}
Paul, R., Arkin, J., Roy, N., and Howard, T.~M. (2016).
\newblock Efficient grounding of abstract spatial concepts for natural language
  interaction with robot manipulators.
\newblock In {\em Proceedings of Robotics: Science and Systems (RSS)}.

\bibitem[Pronobis and Jensfelt, 2012]{pronobis12}
Pronobis, A. and Jensfelt, P. (2012).
\newblock Large-scale semantic mapping and reasoning with heterogeneous
  modalities.
\newblock In {\em Proceedings of the IEEE International Conference on Robotics
  and Automation (ICRA)}, pages 3515--3522.

\bibitem[Pronobis et~al., 2010]{pronobis10}
Pronobis, A., Mart{\'i}nez~Mozos, O., Caputo, B., and Jensfelt, P. (2010).
\newblock Multi-modal semantic place classification.
\newblock {\em International Journal of Robotics Research}, 29(2--3):298--320.

\bibitem[Ranganathan and Dellaert, 2011]{ranganathan11}
Ranganathan, A. and Dellaert, F. (2011).
\newblock Online probabilistic topological mapping.
\newblock {\em International Journal of Robotics Research}, 30(6):755--771.

\bibitem[Rasouli et~al., 2020]{rasouli20}
Rasouli, A., Lanillos, P., Cheng, G., and Tsotsos, J.~K. (2020).
\newblock Attention-based active visual search for mobile robots.
\newblock {\em Autonomous Robots}, 44(2):131--146.

\bibitem[Ratliff et~al., 2006]{ratliff06}
Ratliff, N.~D., Bagnell, J.~A., and Zinkevich, M.~A. (2006).
\newblock Maximum margin planning.
\newblock In {\em Proceedings of the International Conference on Machine
  Learning (ICML)}, pages 729--736.

\bibitem[Redmon and Farhadi, 2018]{yolov3}
Redmon, J. and Farhadi, A. (2018).
\newblock {YOLOv3}: {A}n incremental improvement.
\newblock {\em arXiv:1804.02767}.

\bibitem[Roberts et~al., 2000]{roberts2000autonomous}
Roberts, J.~M., Duff, E.~S., Corke, P.~I., Sikka, P., Winstanley, G.~J., and
  Cunningham, J. (2000).
\newblock Autonomous control of underground mining vehicles using reactive
  navigation.
\newblock In {\em Proceedings of the IEEE International Conference on Robotics
  and Automation (ICRA)}, pages 3790--3795.

\bibitem[Ross et~al., 2011]{ross11}
Ross, S., Gordon, G.~J., and Bagnell, J.~A. (2011).
\newblock A reduction of imitation learning and structured prediction to
  no-regret online learning.
\newblock In {\em Proceedings of the International Conference on Artificial
  Intelligence and Statistics (AISTATS)}, pages 627--635.

\bibitem[Roy et~al., 2003]{roy03}
Roy, D., Hsiao, K.-Y., and Mavridis, N. (2003).
\newblock Conversational robots: {B}uilding blocks for grounding word meaning.
\newblock In {\em Proc.\ HLT-NAACL Workshop on Learning Word Meaning from
  Non-Linguistic Data}, pages 70--77.

\bibitem[Ryu et~al., 2004]{ryu2004multi}
Ryu, D., Kang, S., Kim, M., and Song, J.-B. (2004).
\newblock Multi-modal user interface for teleoperation of {ROBHAZ-DT2} field
  robot system.
\newblock In {\em Proceedings of the IEEE/RSJ International Conference on
  Intelligent Robots and Systems (IROS)}, pages 168--173.

\bibitem[Sadeghi and Levine, 2017]{sadeghi17}
Sadeghi, F. and Levine, S. (2017).
\newblock {CAD2RL}: {R}eal single-image flight without a single real image.
\newblock In {\em Proceedings of Robotics: Science and Systems (RSS)}.

\bibitem[Scheding et~al., 1999]{scheding1999experiment}
Scheding, S., Dissanayake, G., Nebot, E.~M., and Durrant-Whyte, H. (1999).
\newblock An experiment in autonomous navigation of an underground mining
  vehicle.
\newblock {\em IEEE Transactions on Robotics and Automation}, 15(1):85--95.

\bibitem[Scheding et~al., 1997]{scheding1997experiments}
Scheding, S., Nebot, E.~M., Stevens, M., Durrant-Whyte, H., Roberts, J., Corke,
  P., Cunningham, J., and Cook, B. (1997).
\newblock Experiments in autonomous underground guidance.
\newblock In {\em Proceedings of the IEEE International Conference on Robotics
  and Automation (ICRA)}, pages 1898--1903.

\bibitem[She and Chai, 2017]{she17}
She, L. and Chai, J. (2017).
\newblock Interactive learning of grounded verb semantics towards human-robot
  communication.
\newblock In {\em Proceedings of the Association for Computational Linguistics
  (ACL)}, pages 1634--1644.

\bibitem[Shridhar and Hsu, 2018]{shridhar18}
Shridhar, M. and Hsu, D. (2018).
\newblock Interactive visual grounding of referring expressions for human-robot
  interaction.
\newblock In {\em Proceedings of Robotics: Science and Systems (RSS)}.

\bibitem[Singh et~al., 2004]{singh2004seabed}
Singh, H., Can, A., Eustice, R., Lerner, S., McPhee, N., and Roman, C. (2004).
\newblock Seabed {AUV} offers new platform for high-resolution imaging.
\newblock {\em Eos, Transactions American Geophysical Union}, 85(31):289--296.

\bibitem[Smola et~al., 2007]{smola2007}
Smola, A., Gretton, A., Song, L., and Sch{\"o}lkopf, B. (2007).
\newblock A {H}ilbert space embedding for distributions.
\newblock In {\em Proceedings of the International Conference on Algorithmic
  Learning Theory (ALT)}, pages 13--31.

\bibitem[Syed et~al., 2008]{syed08a}
Syed, U., Bowling, M., and Schapire, R.~E. (2008).
\newblock Apprenticeship learning using linear programming.
\newblock In {\em Proceedings of the International Conference on Machine
  Learning (ICML)}, pages 1032--1039.

\bibitem[Tai et~al., 2017]{tai17}
Tai, L., Paolo, G., and Liu, M. (2017).
\newblock Virtual-to-real deep reinforcement learning: {C}ontinuous control of
  mobile robots for mapless navigation.
\newblock In {\em Proceedings of the IEEE/RSJ International Conference on
  Intelligent Robots and Systems (IROS)}, pages 31--36.

\bibitem[Tellex et~al., 2014]{tellex14}
Tellex, S., Knepper, R., Li, A., Rus, D., and Roy, N. (2014).
\newblock Asking for help using inverse semantics.
\newblock In {\em Proc.\ Robotics: Science and Systems (RSS)}.

\bibitem[Tellex et~al., 2011]{tellex11}
Tellex, S., Kollar, T., Dickerson, S., Walter, M.~R., Banerjee, A.~G., Teller,
  S., and Roy, N. (2011).
\newblock Understanding natural language commands for robotic navigation and
  mobile manipulation.
\newblock In {\em Proceedings of the National Conference on Artificial
  Intelligence (AAAI)}, pages 1507--1514.

\bibitem[Tellex et~al., 2012]{tellex12}
Tellex, S., Thaker, P., Deits, R., Kollar, T., and Roy, N. (2012).
\newblock Toward information theoretic human-robot dialog.
\newblock In {\em Proceedings of Robotics: Science and Systems (RSS)}.

\bibitem[Thomason et~al., 2018]{thomason18}
Thomason, J., Sinapov, J., Mooney, R.~J., and Stone, P. (2018).
\newblock Guiding exploratory behaviors for multi-modal grounding of linguistic
  descriptions.
\newblock In {\em Proceedings of the National Conference on Artificial
  Intelligence (AAAI)}, pages 5520--5527.

\bibitem[Thomason et~al., 2016]{thomason16}
Thomason, J., Sinapov, J., Svetlik, M., Stone, P., and Mooney, R.~J. (2016).
\newblock Learning multi-modal grounded linguistic semantics by playing ``{I}
  spy''.
\newblock In {\em Proceedings of the International Joint Conference on
  Artificial Intelligence (IJCAI)}, pages 3477--3483.

\bibitem[Thomason et~al., 2015]{thomason15}
Thomason, J., Zhang, S., Mooney, R.~J., and Stone, P. (2015).
\newblock Learning to interpret natural language commands through human-robot
  dialog.
\newblock In {\em Proceedings of the International Joint Conference on
  Artificial Intelligence (IJCAI)}, pages 1923--1929.

\bibitem[Thrun et~al., 2004]{thrun04}
Thrun, S., Liu, Y., Koller, D., Ng, A., Ghahramani, Z., and Durrant-Whyte, H.
  (2004).
\newblock Simultaneous localization and mapping with sparse extended
  information filters.
\newblock {\em International Journal of Robotics Research}, 23(7--8):693--716.

\bibitem[Thrun et~al., 2006]{thrun06a}
Thrun, S., Montemerlo, M., Dahlkamp, H., Stavens, D., Aron, A., Diebel, J.,
  Fong, P., Gale, J., Halpenny, M., Hoffmann, G., Lau, K., Oakley, C.,
  Palatucci, M., Pratt, V., Stang, P., Strohband, S., Dupont, C., Jendrossek,
  L.-E., Koelen, C., Markey, C., Rummel, C., van Niekerk, J., Jensen, E.,
  Alessandrini, P., Bradski, G., Davies, B., Ettinger, S., Kaehler, A., Nefian,
  A., and Mahoney, P. (2006).
\newblock Stanley: {T}he robot that won the {DARPA} {G}rand {C}hallenge.
\newblock {\em Journal of Field Robotics}, 23(9):661--692.

\bibitem[Torralba et~al., 2003]{torralba03}
Torralba, A., Murphy, K.~P., Freeman, W.~T., and Rubin, M.~A. (2003).
\newblock Context-based vision system for place and object recognition.
\newblock In {\em Proceedings of the International Conference on Computer
  Vision (ICCV)}, pages 273--280.

\bibitem[Tucker et~al., 2017]{tucker17}
Tucker, M., Aksaray, D., Paul, R., Stein, G.~J., and Roy, N. (2017).
\newblock Learning unknown groundings for natural language interaction with
  mobile robots.
\newblock In {\em International Symposium on Robotics Research (ISRR)}, pages
  317--333.

\bibitem[Urmson et~al., 2008]{urmson08a}
Urmson, C., Anhalt, J., Bagnell, D., Baker, C., Bittner, R., Clark, M.~N.,
  Dolan, J., Duggins, D., Galatali, T., Geyer, C., Gittleman, M., Harbaugh, S.,
  Hebert, M., Howard, T.~M., Kolski, S., Kelly, A., Likhachev, M., McNaughton,
  M., Miller, N., Peterson, K., Pilnick, B., Rajkumar, R., Rybski, P., Salesky,
  B., Seo, Y.-W., Singh, S., Snider, J., Stentz, A., Whittaker, W.~R.,
  Wolkowicki, Z., Ziglar, J., Bae, H., Brown, T., Demitrish, D., Litkouhi, B.,
  Nickolaou, J., Sadekar, V., Zhang, W., Struble, J., Taylor, M., Darms, M.,
  and Ferguson, D. (2008).
\newblock Autonomous driving in urban environments: {B}oss and the {U}rban
  {C}hallenge.
\newblock {\em Journal of Field Robotics}, 25(8):425--466.

\bibitem[Urmson et~al., 2006]{urmson06a}
Urmson, C., Ragusa, C., Ray, D., Anhalt, J., Bartz, D., Galatali, T.,
  Gutierrez, A., Johnston, J., Harbaugh, S., {``Yu''}~Kato, H., Messner, W.,
  Miller, N., Peterson, K., Smith, B., Snider, J., Spiker, S., Ziglar, J.,
  {``Red''}~Whittaker, W., Clark, M., Koon, P., Mosher, A., and Struble, J.
  (2006).
\newblock A robust approach to high-speed navigation for unrehearsed desert
  terrain.
\newblock {\em Journal of Field Robotics}, 23(8):467--508.

\bibitem[Vasudevan and Siegwart, 2008]{vasudevan08}
Vasudevan, S. and Siegwart, R. (2008).
\newblock Bayesian space conceptualization and place classification for
  semantic maps in mobile robotics.
\newblock {\em Robotics and Autonomous Systems}, 56(6):522--537.

\bibitem[Walter et~al., 2015]{walter15}
Walter, M.~R., Antone, M., Chuangsuwanich, E., Correa, A., Davis, R., Fletcher,
  L., Frazzoli, E., Friedman, Y., Glass, J., How, J.~P., Jeon, J.~H., Karaman,
  S., Luders, B., Roy, N., Tellex, S., and Teller, S. (2015).
\newblock A situationally-aware voice-commandable robotic forklift working
  alongside people in unstructured outdoor environments.
\newblock {\em Journal of Field Robotics}, 32(4):590--628.

\bibitem[Walter et~al., 2007]{walter07}
Walter, M.~R., Eustice, R.~M., and Leonard, J.~J. (2007).
\newblock Exactly sparse extended information filters for feature-based {SLAM}.
\newblock {\em International Journal of Robotics Research}, 26(4):335--359.

\bibitem[Walter et~al., 2013]{walter13}
Walter, M.~R., Hemachandra, S., Homberg, B., Tellex, S., and Teller, S. (2013).
\newblock Learning semantic maps from natural language descriptions.
\newblock In {\em Proceedings of Robotics: Science and Systems (RSS)}.

\bibitem[Walter et~al., 2014]{walter14}
Walter, M.~R., Hemachandra, S., Homberg, B., Tellex, S., and Teller, S. (2014).
\newblock A framework for learning semantic maps from grounded natural language
  descriptions.
\newblock {\em International Journal of Robotics Research}, 33(9):1167--1190.

\bibitem[Williams et~al., 2012]{williams2012monitoring}
Williams, S.~B., Pizarro, O.~R., Jakuba, M.~V., Johnson, C.~R., Barrett, N.~S.,
  Babcock, R.~C., Kendrick, G.~A., Steinberg, P.~D., Heyward, A.~J., Doherty,
  P.~J., et~al. (2012).
\newblock Monitoring of benthic reference sites: using an autonomous underwater
  vehicle.
\newblock {\em IEEE Robotics \& Automation Magazine}, 19(1):73--84.

\bibitem[Winograd, 1971]{winograd71}
Winograd, T. (1971).
\newblock {\em Procedures As A Representation for Data in a Computer Program
  for Understanding Natural Language}.
\newblock PhD thesis, Massachusetts Institute of Technology.

\bibitem[Yamauchi, 2004]{yamauchi2004packbot}
Yamauchi, B.~M. (2004).
\newblock {PackBot}: {A} versatile platform for military robotics.
\newblock In {\em Proceedings of the International Society for Optics and
  Photonics (SPIE), Unmanned Ground Vehicle Technology VI}, volume 5422, pages
  228--237.

\bibitem[Yoerger et~al., 2007]{yoerger2007techniques}
Yoerger, D.~R., Jakuba, M., Bradley, A.~M., and Bingham, B. (2007).
\newblock Techniques for deep sea near bottom survey using an autonomous
  underwater vehicle.
\newblock {\em The International Journal of Robotics Research}, 26(1):41--54.

\bibitem[Zender et~al., 2008]{zender08}
Zender, H., Mart{\'\i}nez~Mozos, O., Jensfelt, P., Kruijff, G., and Burgard, W.
  (2008).
\newblock Conceptual spatial representations for indoor mobile robots.
\newblock {\em Robotics and Autonomous Systems}, 56(6):493--502.

\bibitem[Zeng et~al., 2019]{zeng19}
Zeng, W., Luo, W., Suo, S., Sadat, A., Yang, B., Casas, S., and Urtasun, R.
  (2019).
\newblock End-to-end interpretable neural motion planner.
\newblock In {\em Proceedings of the IEEE Conference on Computer Vision and
  Pattern Recognition (CVPR)}, pages 8660--8669.

\bibitem[Zhu et~al., 2017]{zhu17}
Zhu, Y., Mottaghi, R., Kolve, E., Lim, J.~J., Gupta, A., Fei-Fei, L., and
  Farhadi, A. (2017).
\newblock Target-driven visual navigation in indoor scenes using deep
  reinforcement learning.
\newblock In {\em Proceedings of the IEEE International Conference on Robotics
  and Automation (ICRA)}, pages 3357--3364.

\end{thebibliography}

\end{document}